\documentclass[final]{elsarticle}
\usepackage[a4paper]{geometry}
\usepackage{booktabs}
\usepackage{graphicx}
\usepackage{amsmath}
\usepackage{multirow}
\usepackage{amsfonts}
\usepackage{makecell}
\usepackage{algorithm}
\usepackage{enumitem}
\usepackage{float}
\usepackage{algpseudocode}
\usepackage[toc,page]{appendix}
\usepackage{array}
\usepackage{subcaption}
\usepackage{array,arydshln}
\usepackage{amssymb}
\usepackage{hhline}
\usepackage{times,rotating}
\usepackage{tabularx,ragged2e}
\newcolumntype{T}[1]{>{\raggedright\arraybackslash}p{#1}}
\newcolumntype{M}[1]{>{\centering\arraybackslash}m{#1}}

\newcolumntype{L}[1]{>{\raggedright\let\newline\\\arraybackslash\hspace{0pt}}m{#1}}
\newcolumntype{C}[1]{>{\centering\let\newline\\\arraybackslash\hspace{0pt}}m{#1}}
\newcolumntype{R}[1]{>{\raggedleft\let\newline\\\arraybackslash\hspace{0pt}}m{#1}} 

\newtheorem{problem}{Problem}

\usepackage{multicol} 

\usepackage{bm} 

\usepackage{rotating}

\usepackage{xcolor}
\definecolor{orange}{HTML}{FFC17D}
\definecolor{blue}{HTML}{7BABFF}
\definecolor{green}{HTML}{A1D68B}

\journal{Information Fusion}

\begin{document}

\begin{frontmatter}

\title{Lights and Shadows in Evolutionary Deep Learning: Taxonomy, Critical Methodological Analysis, Cases of Study, Learned Lessons, Recommendations and Challenges}

\author[a]{Aritz D. Martinez}
\author[a,c]{Javier Del Ser\corref{cor1}}
\author[a]{Esther Villar-Rodriguez}
\author[a]{Eneko Osaba}
\author[f]{Javier Poyatos}
\author[f]{\\Siham Tabik}
\author[f]{Daniel Molina}
\author[f]{Francisco Herrera}

\address[a]{TECNALIA, Basque Research \& Technology Alliance (BRTA), 48160 Derio, Spain}
\address[c]{University of the Basque Country (UPV/EHU), 48013 Bilbao, Spain}
\address[f]{DaSCI Andalusian Institute of Data Science and Computational Intelligence, University of Granada, 18071 Granada, Spain}
\cortext[cor1]{Corresponding author. TECNALIA, Basque Research \& Technology Alliance (BRTA), P. Tecnologico, Ed. 700. 48170 Derio (Bizkaia), Spain. E-mail: \texttt{javier.delser@tecnalia.com} (Javier Del Ser).}

\begin{abstract}
Much has been said about the fusion of bio-inspired optimization algorithms and Deep Learning models for several purposes: from the discovery of network topologies and hyperparametric configurations with improved performance for a given task, to the optimization of the model’s parameters as a replacement for gradient-based solvers. Indeed, the literature is rich in proposals showcasing the application of assorted nature-inspired approaches for these tasks. In this work we comprehensively review and critically examine contributions made so far based on three axes, each addressing a fundamental question in this research avenue: a) optimization and taxonomy (\emph{Why?}), including a historical perspective, definitions of optimization problems in Deep Learning, and a taxonomy associated with an in-depth analysis of the literature, b) critical methodological analysis (\emph{How?}), which together with two case studies, allows us to address learned lessons and recommendations for good practices following the analysis of the literature, and c) challenges and new directions of research (\emph{What can be done, and what for?}). In summary, three axes -- optimization and taxonomy, critical analysis, and challenges -- which outline a complete vision of a merger of two technologies drawing up an exciting future for this area of fusion research.
\end{abstract}


\end{frontmatter}

\section{Introduction} \label{sec:intro}

Nowadays there is overall consensus on the capital importance gained by Deep Learning in the Artificial Intelligence field \cite{lecun15}. Initial results of Deep Learning date back to the late 80's, stepping on a history of preceding achievements in neural computation \cite{lecun1989backpropagation}. However, it was not until years later when advances in high-performance computing, new achievements in neural network training \cite{hinton2006fast}, and the availability of massive datasets paved the way for the renowned success of this family of learning models. Nowadays, plenty of application areas have harnessed the superior modeling capabilities of Deep Learning models, from natural language processing \cite{young2018recent}, speech and audio processing \cite{kolbk2017speech,zhang2017very}, social network analysis \cite{pal2016deep} or autonomous driving \cite{grigorescu2019survey}, to mention a few. As a result, Deep Learning models such as Convolutional Neural Networks (CNN, \cite{krizhevsky2012imagenet}), Recurrent Neural Networks (RNN, \cite{cho2014learning}) or Generative Adversarial Networks (GAN, \cite{goodfellow2014generative}) prevail in many tasks, including image classification, time series forecasting or visual object generation.

There are several properties of Deep Learning models that make them outperform traditional \textit{shallow learning} methods. Among them, Deep Learning models can automatically learn hierarchical features from raw data, so that features organized in higher levels of the hierarchy are composed by a combination of simpler lower-level features. As a result of this capability, features with minimal human effort and domain knowledge can be learned and fused together for a manifold of tasks, such as classification, regression or representation learning \cite{najafabadi2015deep}. Furthermore, Deep Learning models comprise a large number of parameters to represent such hierarchical features, which are adjusted (\emph{trained}) as per the task under consideration. In addition, Deep Learning approaches can model highly non-linear mappings between their inputs and outputs \cite{yu2013deep,fong2018meta}. Finally, decisions issued by these black-box models can be explained to non-expert users, making these black-box models of practical use in domains where explainability is a must \cite{arrieta2020explainable}.

In the Artificial Intelligence field we can find many evidences of the potential of the fusion of different technologies to tackle complex tasks. Deep Learning is not an exception to this statement: the fact that the architectural design, hyper-parameter tuning and training of Deep Learning can be formulated as optimization problems has motivated a long story between these models and the field of bio-inspired optimization, particularly Evolutionary Computation and Swarm Intelligence methods. The number of layers, their dimension and type of neurons, intermediate processing elements and other structural parameters can span a large solution space demanding search heuristics for their efficient exploration. Similarly, hyper-parameter tuning in Deep Learning models can be approached via heuristic wrappers, whereas their training process is essentially the minimization of a task-dependent loss function with respect to all the trainable parameters. 

The purpose of this manuscript is to perform a thorough assessment of the potential of meta-heuristic algorithms for Deep Learning, with a proper understanding of the current state-of-the-art of this research area. It is supported by an exhaustive critical examination of the recent literature falling in this intersection, and a profound reflection, informed with empirical results, on the lights and shadows of this research area. The contributions of this study can be summarized as follows:
\begin{itemize}[leftmargin=*]
\item We perform a brief hindsight on the historical confluence of both research areas so as to frame the importance of our study.

\item We mathematically define concepts and notions on bio-inspired optimization and Deep Learning that help the reader follow the rest of the overview. 

\item We present a taxonomy that allows categorizing every proposal reported so far according to three criteria: a) the Deep Learning model under consideration; b) the optimization goal for which the bio-inspired algorithm is devised, distinguishing among architectural design, hyper-parameter configuration and model training; and c) the search mechanism followed by the bio-inspired solver(s) in use, either Evolutionary Computation, Swarm Intelligence or hybrid methods.

\item Based on this taxonomy, we perform a detailed examination of the literature belonging to each category, pausing at milestones that supposed a genuine advance in the field, as well as identifying poor practices and misconceptions that should be avoided for the benefit of the community. 

\item We design and discuss on two experiments focused on two cases of study aimed at eliciting empirical evidence on the performance of bio-inspired optimization algorithms when applied to the topological design, hyper-parameter tuning and training of Deep Learning models. 

\item We provide a series of lessons and methodological recommendations learned from the literature analysis and the experiments, which should establish the minimum requirements to be met by future studies on the fusion of bio-inspired optimization and Deep Learning.

\item A prospect is made towards the future of this research area by identifying several challenges that remain insufficiently addressed to date, and by suggesting research directions that could bring effective solutions to such challenges.
\end{itemize}

\begin{figure}[ht!]
\centering
\includegraphics[width=0.45\columnwidth]{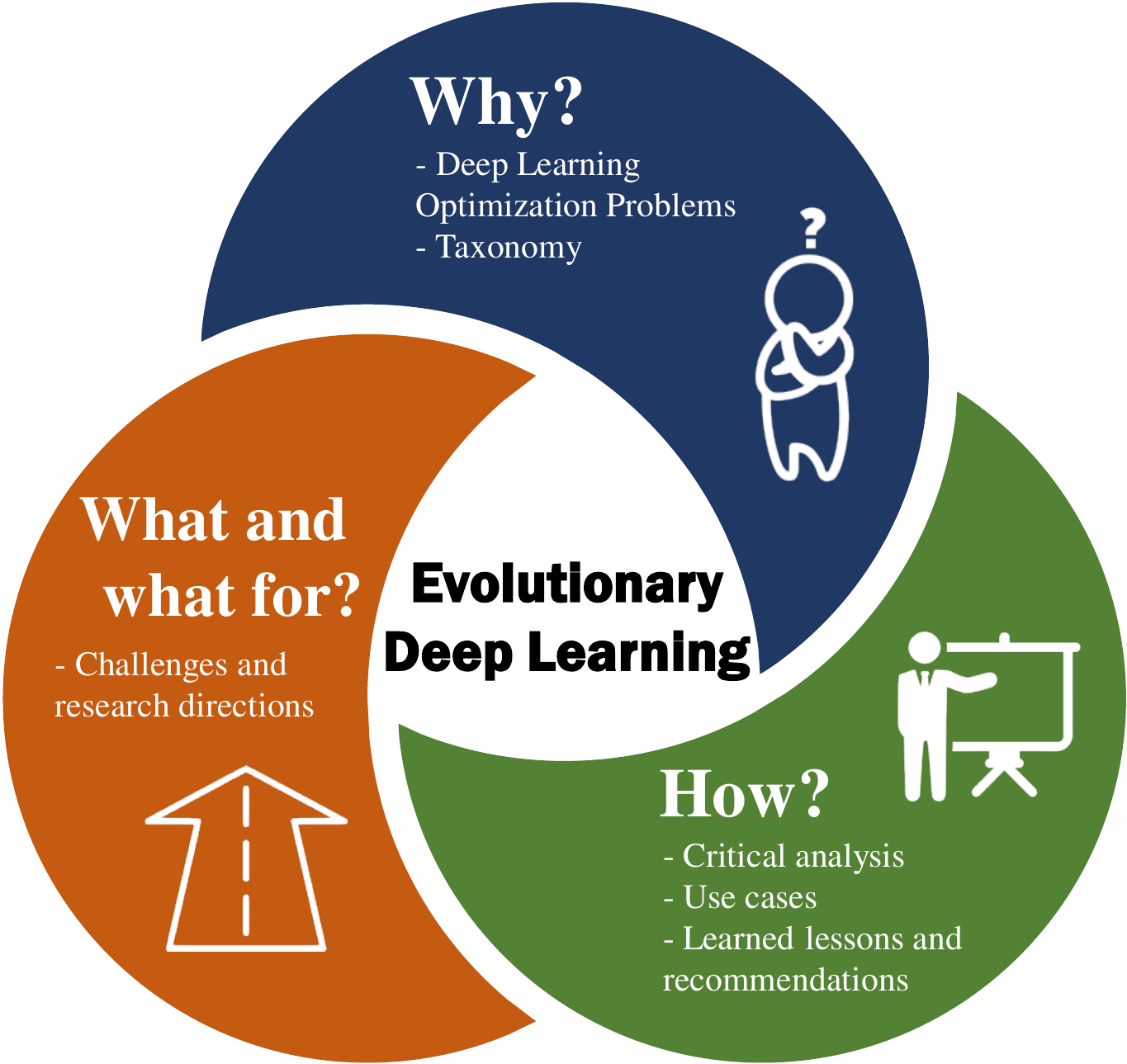}
\caption{Diagram depicting the three axes and four fundamental questions on Evolutionary Deep Learning tackled in the overview, along with the specific aspects that contributes to each question.}
\label{fig:axes}
\end{figure}

In summary, we comprehensively review and critically examine contributions made in this research avenue based on three axes: a) \emph{optimization and taxonomy}, which comprises a historical perspective on this fusion of technologies, a clear definition of the optimization problems in Deep Learning, and a taxonomy associated to an in-depth analysis of the literature; b) \emph{critical analysis}, informed by two case studies, which altogether elicit a number of lessons learned and recommendations for good practices, and c) \emph{challenges} that motivate new directions of research for the near future. These three axes aim to provide a clear response to four important questions related to Evolutionary Deep Learning\footnote{Throughout the survey we embrace the term \emph{Evolutionary Deep Learning} to refer to the use of bio-inspired algorithms for solving optimization problems related to Deep Learning, no matter if they belong to Evolutionary Computation or to Swarm Intelligence.}, which are represented in Figure \ref{fig:axes}:
\begin{itemize}[leftmargin=*]
\item Why are bio-inspired algorithms of interest for the optimization of Deep Learning models?
\item How should research studies falling in the intersection between bio-inspired optimization and Deep Learning be made?
\item What can be done in future investigations on this topic?
\item What should future research efforts be conducted for?
\end{itemize}

The structure of the manuscript conforms to the above three axes: first, Section \ref{sec:history} briefly overviews the historical connection between Deep Learning and bio-inspired optimization. Next, Section \ref{sec:preliminaries} defines mathematically the optimization problems associated to Deep Learning models. Section \ref{sec:taxonomy_literature} presents our taxonomy and an analysis of the literature falling on each of its categories. Section \ref{sec:critic} exposes methodological caveats resulting from our critical literature study. Sections \ref{ssec:exp1} and \ref{ssec:exp2} present the two designed cases of study, and discuss the results obtained therefrom. Section \ref{sec:lessons} enumerates learned lessons and prescribes good practices to be followed by prospective studies. Section \ref{sec:challenges} outlines several challenges and research directions that should drive future research efforts of the interested audience. Finally, Section \ref{sec:conclusions} point out the final outline and conclusions of the overview. Additionally, \ref{ssec:DLmodels} revisits several Deep Learning models, whereas \ref{ssec:metah} introduces the reader to the field of bio-inspired optimization, placing emphasis on Evolutionary Computation and Swarm Intelligence.

\section{Historical Tour on Evolutionary Neural Networks and Evolutionary Deep Learning} \label{sec:history}

\begin{figure}[ht!]
\centering
\includegraphics[width=0.95\columnwidth]{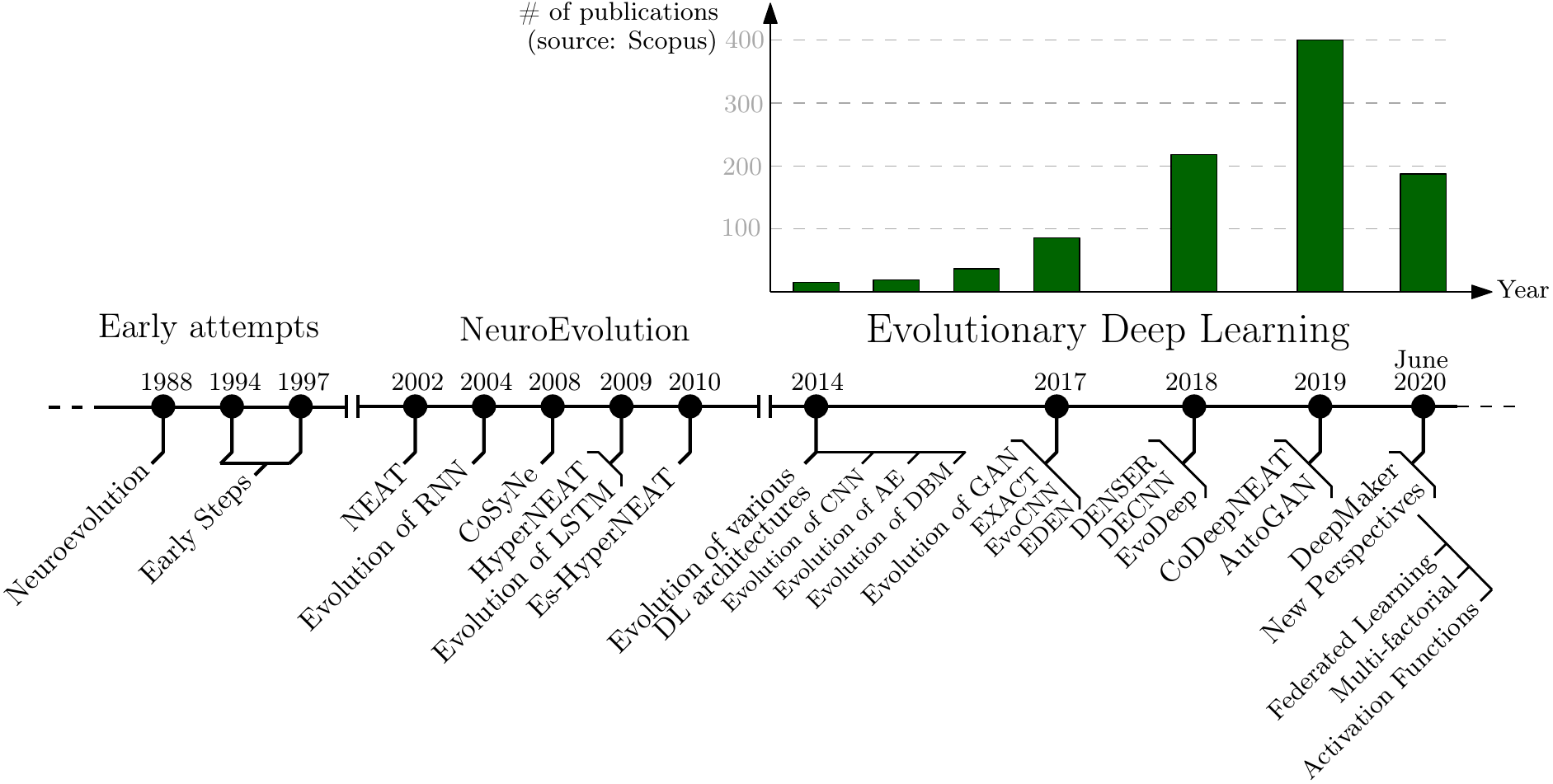}
\caption{Timeline with main milestones in the history of Evolutionary Machine Learning, and a bar diagram showing the number of publications reported in this research area during the period 2013-June 2020. Data retrieved from Scopus by submitting the query (\texttt{EVOLUTIONARY} OR \texttt{SWARM} \texttt{INTELLIGENCE}) AND \texttt{DEEP} \texttt{LEARNING}.}
\label{fig:timeline}
\end{figure}

Despite its relative youth, the current momentum of the synergy between Deep Learning and bio-inspired optimization is founded on a set of historical milestones that suggested the scientific community to combine these two branches of Artificial Intelligence. We herein revisit briefly this profitable background that led into the literature mainstream that motivates the current study. Figure \ref{fig:timeline} summarizes graphically such milestones, arranging them in a timeline along with the number of related publications reported in the last few years (data retrieved from the Scopus database, with the search terms indicated in the caption of the figure).

Although timid attempts at evolving neural networks with bio-inspired solvers had been reported in the late 90s \cite{yao1997new}, it was not until 2002 when Stanley and Miikkulainen settled a major breakthrough in the research community with their seminal work ``Evolving neural networks through augmenting topologies'' \cite{stanley2002evolving}. The NEAT approach proposed in this work allowed connection and layer types of an artificial neural network architecture (ANN) to be optimized by means of a meta-heuristic algorithm towards a progressively better precision of the evolved ANN model for a given task. NEAT embraces the main workflow of population-based meta-heuristics, particularly genetic algorithms: a population of encoded candidates is generated representing several network architectures, from which new candidate architectures are produced and evaluated on a given task (in the original work, a Reinforcement Learning task). After all candidates in the population have been evaluated, mutation and crossover operators are applied, generating a new population by means of combining network architectures, generating new layers or varying their hyper-parameters. This iterative search process is stopped when a stopping criterion set beforehand is met. As just stated, this algorithm was mainly proposed to solve reinforcement learning tasks, keeping in mind the profit of applying meta-heuristics to such environments, such as getting interesting behaviours and not falling in local optima in expense of precision.

Shortly after its first publication, NEAT's unprecedented results spurred a flurry of new extensions and variants, not only in terms of new tasks and applications, but also in what refers to its core algorithmic components. Regarding the latter, the acknowledged importance of using good network encoding strategies soon became a major research goal in this literature strand, given the huge search space spawned by the evolution of architecture and weights of ANNs. In its original version, NEAT encoded candidates using a Direct encoding strategy, i.e. networks' hidden units, connections and parameters (phenotype) were \textit{directly} represented as an array representing each point of the network (genotype). However, in 2009 Stanley et al. proposed the so-called Hypercube-based NEAT, or HyperNEAT, \cite{Stanley2009}, which relies on a generative encoding strategy to evolve large-scale neural networks using geometric regularities of the task domain and compositional pattern producing networks (namely, an ANN variant comprising multiple potentially heterogeneous activation functions that can be evolved via genetic algorithms). Besides the optimization of the activation function of each neuron in the network, HyperNEAT also proposed to utilize an indirect encoding to represent the networks to be evolved, inheriting other concepts from preceding NEAT versions such as speciation and historical marking. A new NEAT approach was developed years later for CNNs, which was coined as CoDeepNeat \cite{Miikkulainen2017EvolvingDN}. Following the NEAT design principles, CoDeepNeat was proved to excel at evolving layers, parameters, topology and hyper-parameters of CNNs. To this end, it uses an indirect encoding approach so that the network information can be encoded as rules or processes for creating individuals, ultimately yielding a reduced representation of the search space that can be explored more efficiently by the bio-inspired algorithm in use.
\begin{table}[h!]
    \centering
    \caption{Software frameworks to evolve Deep Learning models using meta-heuristics and other search strategies.}
    \resizebox{\linewidth}{!}{\begin{tabular}{|C{1.5cm}|C{2.2cm}|C{1cm}|C{3cm}|C{6cm}|C{6cm}|L{2cm}|}
        \toprule
        \textbf{Reference} & \textbf{Name} & \textbf{Year} & \textbf{Deep Learning models} & \textbf{Optimization domains} & \textbf{Optimization algorithm} & \textbf{Programming Language} \\
        \midrule
        \cite{Real2017} & EvoCNN & 2017 & CNN & Topology, structural hyper-parameters, weight initialization & Genetic algorithm & Python \\
        \midrule
        \cite{munoz2018framework} & DNF & 2018 & CNN & Trainable parameters (weights and biases) & Particle Swarm Optimization, Global-best Harmony Search and Differential Evolution & Python, TensorFlow \\
        \midrule
        \cite{Comput2018} & EvoDeep & 2018 & CNN & Topology, structural hyper-parameters, training hyper-parameters & Evolutionary Algorithm with specific mutation and crossover operators & Python, Keras, Tensorflow \\
        \midrule
        \cite{assunccao2019denser} & DENSER & 2019 & CNN & Topology, structural hyper-parameters, training hyper-parameters & Genetic Algorithm and Dynamic Structured Grammatical Evolution & Python \\
        \midrule
        \cite{googlecloudautoml} & Google Cloud AutoML & 2019 & CNN, RNN & Topology, structural hyper-parameters, training hyper-parameters & Transfer learning + neural architecture search & Proprietary \\
        \midrule
        \cite{jin2019auto} & AutoKeras & 2019 & CNN, RNN & Topology & Hyperband, Random, Greedy & Python, Keras \\
        \midrule
        \cite{pham2018efficient} & ENAS & 2018 & CNN, RNN & Topology & Policy gradient-based subgraph selection &  Python, Tensorflow \\
        \midrule
        \cite{liang2019evolutionary} & LEAF & 2019 & CNN, RNN & Topology, structural hyper-parameters, training hyper-parameters & CoDeepNEAT & Python, Pytorch\\
        \midrule
        \cite{charte2020evoaaa} & EvoAAA & 2020 & AE & Topology & Genetic Algorithm, Evolution Strategy, Differential Evolution  & R \\
        \midrule
        \cite{Molino2019} & Ludwig & 2019  & CNN, LSTM, RNN, Fully-Connected & Structural and training hyper-parameters & Tree of Parzen Estimators & Python, TensorFlow \\
        \midrule
        \cite{bohrer2020neuroevolution} & Keras-CoDeepNEAT & 2020 & CNN, RNN & Topology, structural hyper-parameters, training hyper-parameters & CoDeepNEAT & Python, Keras \\
        \midrule
        \cite{baker2017designing} & MetaQNN & 2017 & CNN & Topology, structural hyper-parameters & Reinforcement Learning & Python, Caffe \\
        \midrule
        \cite{davison2017devol} & DEvol & 2017 & CNN & Topology, structural hyper-parameters & Genetic Algorithm  & Python, Keras\\
        \midrule
        \cite{such2017deep, conti2018improving} & AI Labs Neuroevolution Algorithms  & 2017-2018 & CNN  & Trainable parameters (weights and biases) &  Genetic Algorithm + Novelty Search & Python, TensorFlow \\
        \midrule
        \cite{suganuma2017genetic} & CGP-CNN & 2017 & CNN & Topology, structural hyper-parameters & Cartesian Genetic Programming  & Python, Chainer \\
        \midrule
        \cite{cortes2017adanet} & AdaNet & 2017 & Architectures that can be represented as a directed acyclic graph & Topology, structural hyper-parameters, training hyper-parameters, trainable parameters & AdaNet Algorithm & Python, TensorFlow \\
        \midrule
        \cite{mendoza-automlbook18a} & Auto-Pytorch  & 2018 & CNN (+ shallow learning) & Topology, structural hyper-parameters, training hyper-parameters & Random-forest-based Bayesian optimization & Python, Pytorch \\
        \bottomrule
    \end{tabular}}
    \label{tab:frameworks_evo}
\end{table}

Since its first appearance, neuroevolution approaches (with NEAT at their forefront) have been applied to multitude of tasks and problems. A major fraction of them relate to reinforcement learning, such as car controllers \cite{cardamone2009evolving} or first-person agents control \cite{Stanley2005}. Interestingly, years after these applications were reported, the community shifted its research focus towards bio-inspired algorithms as an efficient replacement to train Deep Reinforcement Learning methods \cite{Petroski2017}, showing up competitive results. Bio-inspired optimization was also applied to other Deep Learning tasks at the time, particularly for CNNs \cite{verbancsics2013generative}, and showcased in a diversity of applications including the classification of epileptic EEGs \cite{gao2012bp}, time series forecasting \cite{donate2013time} or scheduling deicing tasks in airports \cite{Mao}. 

As a result of these findings, a growing literature corpus started to explore the potential of neuro-evolution strategies for the topology and structural hyper-parameter optimization of different Deep Learning models, including AE, DBM and GANs. Naturally, the research community soon flowed into a further use of bio-inspired algorithms: the optimization of the trainable parameters of Deep Learning models \cite{morse2016simple}. An early approach was explored in \cite{Mason2017}, where a genetic algorithm is used for architecture optimization, and differential evolution for weight optimization. The main scientific interest of this work and those published thereafter is to assess whether meta-heuristics can avoid the convergence limitations of gradient descent methods when facing strongly non-convex search spaces as those characterizing the problem of training complex Deep Learning architectures. Nonetheless, most agreed on the dilated computation time and enormous computational resources needed by meta-heuristic solvers, which outweigh in practice the eventual convergence gains reported in experimental studies. As a result, gradient back-propagation approaches have dominated as the Deep Learning training solvers of choice over the years. Recently, this tendency is again emerging, stimulated by advances in highly parallel computing paradigms (including GPU and distributed asynchronous computation), prospecting new ways to train Deep Learning models with bio-inspired algorithms appearing frequently as promising alternatives. 
\begin{table}[ht]
    \centering
    \caption{Recent overviews on evolutionary Deep Learning and related topics.}
    \resizebox{\linewidth}{!}{\begin{tabular}{|C{1cm}|C{2cm}|C{2cm}|C{4cm}|C{4cm}|C{2cm}|L{7cm}|}
        \toprule
        \textbf{Survey} & \textbf{Period} & \textbf{\# reviewed works} & \textbf{Taxonomy} & \textbf{Coverage (DNN models/tasks)} & \textbf{Empirical study}  & \multicolumn{1}{c}{\textbf{Lessons learned and challenges}}\\
        \midrule
        
        \cite{ojha2017metaheuristic} & 1987-2016 & $\sim 15$ & Yes (optimization domain of the neural network$^{\ast}$)) & CNN, RNN, RL, DBN & No & Relevance of data quality. Evolutionary techniques good at exploration and exploitation but no single method for all optimization tasks.\\\midrule
        
        \cite{baldominos2019automated}  & 2014-2018  & $\sim 50$ & No (temporal analysis) & CNN, RL & No & High computational resources are required. Special emphasis on ensembles, transfer learning, multiobjective and modular evolutionary approaches. \\ \midrule
        
        \cite{surveyEvoML} & 2011-2018  & $\sim 15$  & Yes (NN-based or GP-based and optimization problem(Architecture, Training, Multi-objective)) & CNN & No & Lack of mathematical foundations, computational costs, scalability, poor generalization ability of the evolved model, lack of interpretability\\ \midrule
        
        \cite{darwish2020survey} & 2011-2019  & $\sim 90$ & Yes (Evolutionary/Swarm Intelligence and Deep Learning model) & CNN, DBN, RNN, AE & No & Lack of rigurosity by the community. Costs of implementation, run time and overfitting.  \\ \midrule
        
        \cite{Chiroma} & 2012-2019  & $\sim 20$ & Yes (meta-heuristic and Deep Learning Architecture) & CNN, DBM, DBN  & No & Time, more efforts on enhancing convergence speed and complexity (meta optimization)  \\ \midrule
        
        \textbf{Ours} & 2014-2020 & $\sim 160$ & Yes (Deep Learning model/task and optimization problem) & CNN, AE, DBM, DBN, RNN, GAN, RL & Yes & See Sections \ref{sec:lessons}, \ref{sec:challenges} and \ref{sec:conclusions}\\\bottomrule
    \multicolumn{7}{L{20cm}}{Note: The column ``\# reviewed works'' only takes into account the number of papers related to the optimization of Deep Learning problems. Any other non-related reference has been filtered out and not considered in the reported quantities.}\\
    \multicolumn{7}{L{23cm}}{$\ast$: weights, architecture + weights, input layer, node, learning algorithm, combination of domains.}
    \end{tabular}}
    \label{tab:survey_comp}
\end{table}

Although it is still an incipient research field, some huge companies like Google, Facebook or Uber have glimpsed the power of this hybridization, and have started investing massively in software frameworks towards optimizing their Deep Learning models with meta-heuristics. The research community has also joined this momentum by open-sourcing software packages for this same purpose. EvoDeep \cite{Comput2018}, AutoKeras \cite{jin2018efficient} or Google Cloud AutoML are noteworthy platforms used for autonomously optimizing Deep Learning models, which we collect in Table \ref{tab:frameworks_evo} together with other alternatives from the literature. 

The complexity of these problems, given by the cardinality of their search spaces and/or the large number of variables to be optimized, has stimulated a ever-growing corpus of literature that lasts to date. Advances held within this area have been reviewed in a number of surveys on this topic, listed in Table \ref{tab:survey_comp}. However, our critical inspection of the achievements in this area reported over the years reveals poor methodological practices, unsolved technical caveats and research challenges that deserve a detailed analysis of where we currently stand in this effervescent area.

\section{Deep Learning: Fundamentals and Optimization Problems} \label{sec:preliminaries}

Before delving into the rest of this work, it is first convenient to settle the mathematical formulation of the optimization problems that lie at the core of this review. In this way, we establish what we understand by the different criteria in which the subsequent literature study is organized. 

A Deep Learning model can be seen as a black-box optimizer where some parameters can be manually selected so that the model behaves in a different way. In fact, almost all parameters that can be tuned in a Deep Learning model can be treated as a task to be optimized. Therefore, depending on the parameters to be solved, we can differentiate different optimization problems. This being said, a Deep Learning model can be conceived mathematically as a composition of $N$ different functions (\emph{layers}) that maps its input $\mathbf{x}_n\in\bm{\mathcal{X}}_n$ to an output $\mathbf{y}_n=f_n(\mathbf{x}_n;\bm{W}_n; T_n,\bm{\theta}_n)\equiv f_{n,\bm{W}_n}^{T_n,\bm{\theta}_n}(\mathbf{x}_n)$, where:
\begin{itemize}[leftmargin=*]
\item $T_n\in \mathcal{T}$ denotes the \emph{type of layer}, with $\mathcal{T}$ denoting the set of possible layer types (e.g. convolutional, LSTM, GRU).
\item $\bm{\theta}_n$ is the vector of \emph{structural hyper-parameters} of the layer. The specific parameters in this vector depend on the type $T_n$ of the layer (e.g. $\bm{\theta}_n$ will specify the sizes of the convolutional filters only if $T_n=\texttt{convolutional}$).
\item $\bm{W}_n$ denotes the \emph{trainable parameters} (weights/filter coefficients and biases) of layer $n$, whose type and cardinality depend on $T_n$ and $\bm{\theta}_n$. For instance, if $\mathbf{x}_n$ represents RGB images ($3$ channels), $T_n=\texttt{convolutional}$ and $\bm{\theta}_n$ establishes that layer $n$ comprises five $3\times 3$ convolutional filters, $\bm{W}_n$ will comprise $3\times 3\times 5 \times 3$ weights and $5$ biases, yielding a total of $|\bm{W}_n|=140$ trainable parameters.
\end{itemize}
It is important to highlight, at this point, that the values of the trainable parameters $\bm{W}_n$ must be learned by the model to efficiently perform a given task. 

For the sake of simplicity in subsequent derivations, we will assume that we deal with a supervised learning task in which we assume a training dataset $\mathcal{D}_{tr}=\{(\mathbf{x}_1^m,\mathbf{y}_N^m)\}_{m=1}^{M_{tr}}$, with $\mathbf{y}_N^m\in \bm{\mathcal{Y}}$ denoting the supervised output of input $\mathbf{x}_1\in\bm{\mathcal{X}_1}$, and $M_{tr}$ representing the number of training instances. The trainable parameters of a Deep Learning model are learned from $\mathcal{D}_{tr}$ by means of a \emph{training} algorithm $\{\bm{W}_n\}_{n=1}^N=ALG(\mathcal{D}_{tr},\{T_n,\bm{\theta}_n\}_{n=1}^N; \bm{\vartheta})$, where we refer to $\bm{\vartheta}$ as the set of \emph{training hyper-parameters} of the training algorithm. In general, the training algorithm is driven by the minimization of a task-dependent loss function $L(\widehat{\mathbf{y}}_N^m,\mathbf{y}_N^m)$ that provides a measure of error between the supervision $\mathbf{y}_N^m$ of input $\mathbf{x}_1^m\in\mathcal{D}$ and the corresponding output of the Deep Learning model:
\begin{equation}
\widehat{\mathbf{y}}_N^m = F(\mathbf{x}_1^m; \{\bm{W}_n\}_{n=1}^N; \{T_n\}_{n=1}^N, \{\bm{\theta}_n\}_{n=1}^N)\doteq f_{N,\bm{W}_N}^{T_N,\bm{\theta}_N} \circ f_{N-1,\bm{W}_{N-1}}^{T_{N-1},\bm{\theta}_{N-1}} \circ \ldots \circ f_{1,\bm{W}_1}^{T_1,\bm{\theta}_1}(\mathbf{x}_1^m),
\end{equation}
where $\circ$ denotes composition of functions, i.e. $f\circ g(x) = f(g(x))$. Such a measure of loss computed for every training instance can be averaged to yield a numerical estimation of the performance of the DL model when approximating the supervised instances in $\mathcal{D}_{tr}$:
\begin{equation} \label{eq:losstotal}
L(F;\mathcal{D}_{tr}) = \frac{1}{M_{tr}} \sum_{m=1}^M L(F(\mathbf{x}_1^m; \{\bm{W}_n\}_{n=1}^N; \{T_n\}_{n=1}^N, \{\bm{\theta}_n\}_{n=1}^N),\mathbf{y}_N^m).
\end{equation}

With this notation in mind, we define the following optimization problems that underlie the construction of Deep Learning models:
\begin{problem}
(Topological Optimization) Given a learning task defined on a training dataset $\mathcal{D}_{tr}$, the topological optimization of a DL model refers to the search for the topology of the DL model that best solves the task at hand, wherein topology involves the discovery of the optimal number of layers $N$ and their types $\{T_n\}_{n=1}^N$. This problem assumes fixed values for $\{\bm{\theta}_n\}_{n=1}^N$ (e.g. standard values), and relies on a training algorithm $ALG(\mathcal{D}_{tr},\{T_n,\bm{\theta}_n\}_{n=1}^N; \bm{\vartheta})$ to optimize the trainable parameters $\{\bm{W}_n\}$. Mathematically:
\begin{equation}
\min\limits_{N,\{T_n\}_{n=1}^N} L(F;\mathcal{D}_{tr}),
\end{equation}
where the dependence of the aggregate loss function with respect to $N$ and $\{T_n\}_{n=1}^N$ comes through \eqref{eq:losstotal}, and $\{\bm{W}_n\}_{n=1}^N$ are optimized by means of $ALG(\mathcal{D}_{tr},\{T_n,\bm{\theta}_n\}_{n=1}^N; \bm{\vartheta})$.
\end{problem}

Topology optimization is rarely conceived in isolation with respect to the rest of variables that define a DL model. Instead, topology is often optimized along with the values of their structural hyper-parameters. However, we define this second problem separately so as to allow for a fine-grained literature analysis:
\begin{problem}
(Structural Hyper-parameter Optimization) Given a learning task defined on a training dataset $\mathcal{D}_{tr}$, and a fixed topology of the DL model ($N$ and $\{T_n\}_{n=1}^N$), the optimization of the structural hyper-parameters of the DL model aims to find the best value of $\bm{\theta}_n$ (structural hyper-parameters) for each of their compounding layers. Mathematically:
\begin{equation}
\min\limits_{\{\bm{\theta}_n\}_{n=1}^N} L(F;\mathcal{D}_{tr}),
\end{equation}
where the dependence of the aggregate loss function with respect to variables $\{\bm{\theta}_n\}_{n=1}^N$ comes through \eqref{eq:losstotal}, and $\{\bm{W}_n\}_{n=1}^N$ are optimized by means of $ALG(\mathcal{D}_{tr},\{T_n,\bm{\theta}_n\}_{n=1}^N; \bm{\vartheta})$.
\end{problem}

Finally, the third optimization problem that can be formulated is the training process itself, which aims at finding the values of the parameters $\{\bm{W}_n\}_{n=1}^N$ that minimizes the loss in \eqref{eq:losstotal}. This is indeed the purpose of $ALG(\mathcal{D}_{tr},\{T_n,\bm{\theta}_n\}_{n=1}^N; \bm{\vartheta})$. However, we note at this point that two different formulations of this problem can be made depending on whether variables to be optimized include the set of \emph{training hyper-parameters} $\bm{\vartheta}$:
\begin{problem}
(Training Hyper-parameter Optimization) Given a learning task defined on a training dataset $\mathcal{D}_{tr}$, a fixed topology of the DL model ($N$ and $\{T_n\}_{n=1}^N$), fixed values of their structural hyper-parameters $\bm{\theta}_n$, and a training algorithm $ALG(\mathcal{D}_{tr},\{T_n,\bm{\theta}_n\}_{n=1}^N; \bm{\vartheta})$, the training hyper-parameter optimization problem of a DL model aims to find the best value of $\bm{\vartheta}$ (training hyper-parameters) as:
\begin{equation}
\min\limits_{\bm{\vartheta}} L(F;\mathcal{D}_{tr}),
\end{equation}
where the dependence of the aggregate loss function with respect to $\bm{\vartheta}$ comes through the application of $ALG(\mathcal{D}_{tr},\{T_n,\bm{\theta}_n\}_{n=1}^N; \bm{\vartheta})$ to solve for $\{\bm{W}_n\}_{n=1}^N$ as per \eqref{eq:losstotal}.
\end{problem}
\begin{problem}
(Trainable Parameter Optimization) Given a learning task defined on a training dataset $\mathcal{D}_{tr}$, a fixed topology of the DL model ($N$ and $\{T_n\}_{n=1}^N$), and fixed values of their structural hyper-parameters $\bm{\theta}_n$, the trainable parameter optimization problem of a DL model seeks the best value of $\{\bm{W}_n\}_{n=1}^N$ (trainable parameters) as:
\begin{equation}
\min\limits_{\{\bm{W}_n\}_{n=1}^N} L(F;\mathcal{D}_{tr}),
\end{equation}
for which an optimization (training) algorithm $ALG(\mathcal{D}_{tr},\{T_n,\bm{\theta}_n\}_{n=1}^N; \bm{\vartheta})$ is utilized.
\end{problem}
   \begin{figure}[ht!]
        \centering
        \includegraphics[width=0.75\columnwidth]{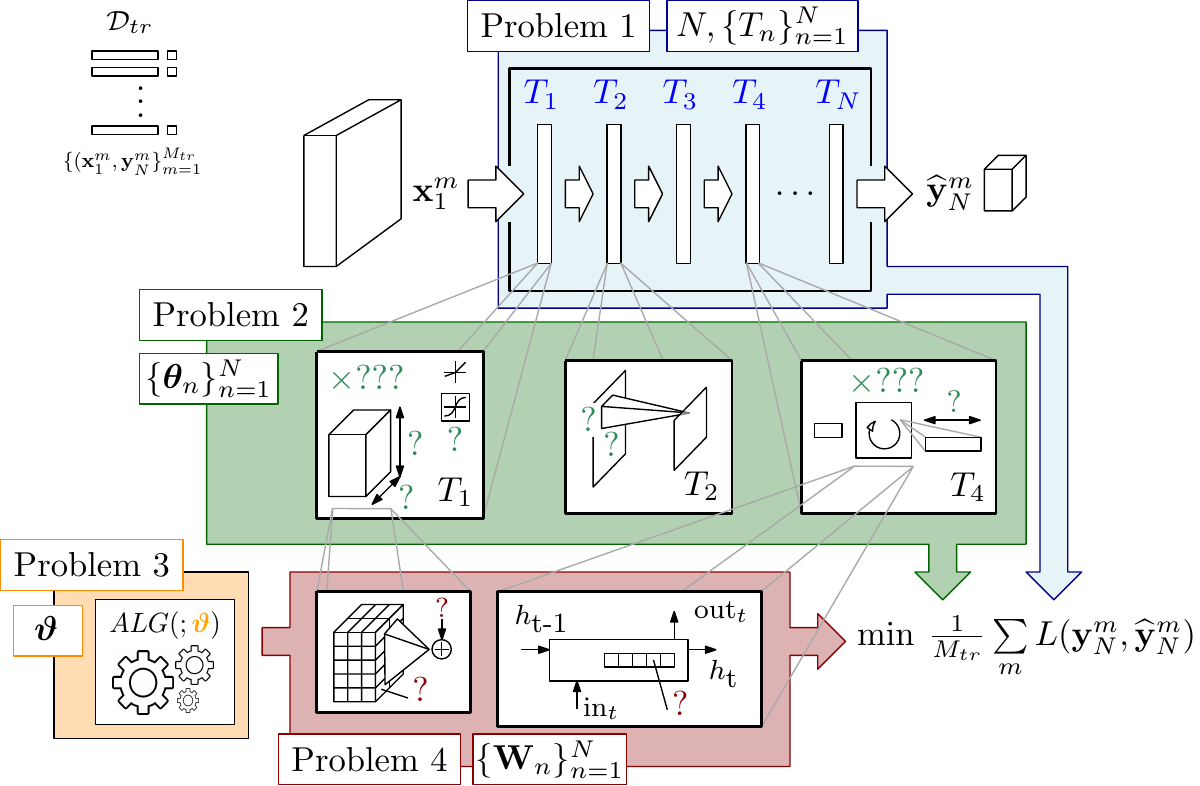}
\caption{Optimization problems in Deep Learning for a generic model comprising, among others, a convolutional layer, a max-pooling layer and a recurrent layer.}
\label{fig:DLProblems}
\end{figure}
   
A visual summary of the above problems is sketched in Figure \ref{fig:DLProblems}. The above $4$ optimization problems can represent the majority of contributions so far elaborating on new algorithms to address them efficiently. However, several practical remarks must be made on these definitions:
\begin{itemize}[leftmargin=*]
\item First of all, it is important to highlight that other measures are often used in the objectives of these problems to replace the aggregate loss $L(F;\mathcal{D}_{tr})$, particularly in Problems 1, 2 and 3. This is the case of cross-validated task-dependent performance metrics (e.g. accuracy or $F_1$ for classification problems), or even the same measures computed over a validation holdout so as to reduce the computational burden of repeatedly evaluating different solution candidates. In Problem 4, however, differentiable loss functions like the ones typically used by gradient-based back-propagation algorithms are often used disregarding the training algorithm adopted.
\item Since the evaluation of the aggregate loss function in Expression \eqref{eq:losstotal} depends on all optimization variables considered in Problems 1 (topology), 2 (structural hyper-parameters) and 3 (training hyper-parameters), it is often the case that studies in the literature consider several problems jointly (e.g. joint topology and structural hyper-parameter optimization). However, as stated previously we find it very convenient to explicitly differentiate among all problems to set clear the optimization domain in which each contributes to the general understanding of the field.
\item Given the scope of our study it is relevant to emphasize that the goal of the reviewed literature strand is to devise an algorithm that, when solving for the variables of each of the above problems, produces lower loss values than other solvers with respect to the dataset and task at hand. However, the design target (the sought \emph{optimization algorithm}) varies depending on the problem under consideration:
\begin{itemize}[leftmargin=*]
\item In Problems 1, 2 and 3, the design target is an optimization algorithm that solves for $N,\{T_n\}_{n=1}^N$ (Problem 1), $\{\bm{\theta}_n\}_{n=1}^N$ (Problem 2) and $\bm{\vartheta}$ (Problem 3). This algorithm under target operates as a wrapper of the overall model, working in parallel to the training algorithm $ALG(\mathcal{D}_{tr},\{T_n,\bm{\theta}_n\}_{n=1}^N; \bm{\vartheta})$. As the latter falls out from the target of of the design process, the training algorithm in use is often set to a naive gradient back-propagation solver (e.g. stochastic gradient descent, Adam and the like).
\item In Problem 4, the design target is the optimization algorithm $ALG(\mathcal{D}_{tr},\{T_n,\bm{\theta}_n\}_{n=1}^N; \bm{\vartheta})$ itself solving for the trainable parameters $\{\bm{W}_n\}_{n=1}^N$, hence there is only one single optimization algorithm. 
\end{itemize}
\end{itemize}

\section{Taxonomy and Literature Review} \label{sec:taxonomy_literature}

In light of the past history between bio-inspired optimization and Deep Learning, a need arises for properly organizing contributions so far in a taxonomy that covers which problems are addressed, which Deep Learning models are involved, and which bio-inspired algorithms are in use. In this section we perform this analysis, centering the discussion around a taxonomy that sorts the literature according to the three aforementioned criteria. 

The main purpose of this taxonomy (Subsection \ref{ssec:taxonomy}) and the literature analysis made over each of its categories (Subsections \ref{ssec:topology}, \ref{ssec:hyp} and \ref{ssec:trainable}) is to highlight those areas where the community has so far placed most research efforts. This literature analysis settles a firm stepping stone towards a critical discussion of poor methodological practices and points of improvement observed in related contributions to date: the \emph{shadows} in which this field is held nowadays. Such a discussion will be held in Section \ref{sec:critic}.

\subsection{Taxonomy} \label{ssec:taxonomy}

As has been stated in Section \ref{sec:preliminaries}, we distinguish among four main optimization tasks: topological optimization (Problem 1), structural hyper-parameter optimization (Problem 2), training hyper-parameter optimization (Problem 3) and trainable parameter optimization (Problem 4). Our taxonomy gathers Problems 2 and 3 under the general hyper-parameter tuning category, discriminating between them in a lower level of the taxonomy. 

The main reason for this special arrangement of the taxonomy is to highlight that as per the reviewed literature ($>$160 references), there is little explicit distinction between structural hyper-parameter and training hyper-parameter in related contributions. Our thorough examination of this corpus has discriminated interesting research opportunities in the extrapolation of studies and frameworks, from training to structural hyper-parameter tuning and vice versa. When it comes to Problem 4, a distinction is made between i) bio-inspired algorithms that do not incorporate any problem-specific knowledge in their design; and ii) bio-inspired solvers that are hybridized with local search solvers or combined with gradient back-propagation techniques.  
\nocite{real2018regularized, kim2017nemo, Xie, Suganuma2017}
\nocite{lu2018nsga, lorenzo2018memetic, chen2019auto, evans2018evolutionary}
\nocite{shafiee2018deep, zhu2020real, desell2017large}
\nocite{wang2019, wang2019particle}
\nocite{wang2019hybrid}
\nocite{Hu2018, rawal2018nodes, Angeline}
\nocite{desell2015evolving}
\nocite{juang2004hybrid}
\nocite{assuncao2018automatic, lander2015evoae}
\nocite{Liu2014}
\nocite{mehta2019neuroevolutionary, costa2019coevolution, costa2019coegan}
\nocite{Stanley2005, pham2018efficient, poulsen2017dlne, pham2018playing}
\nocite{hausknecht2014neuroevolution,stanley2002efficient, etor2020adaptive}
\nocite{Real2017, assunccao2019denser, liang2019evolutionary, suganuma2017genetic}
\nocite{fujino2017deep, Baldominos2018, neuralEvolution2020:ch8, Dufourq2018}
\nocite{akut2019neuroevolution, assunccao2019fast, bochinski2017hyper}
\nocite{prellberg2018lamarckian, sun2018automatically, zhang2019}
\nocite{miikkulainen2019evolving, Elsken2017, ma2020autonomous}
\nocite{neuralEvolution2020:ch7, gu2019esae, suganuma2020evolution}
\nocite{zhu2019multi, loni2020deepmaker, lu2019multi}
\nocite{sun2019evolving, sun2019completely, Rakhshani2020adaptive}
\nocite{lu2020nsganetv2}
\nocite{wang2018evolving, neuralEvolution2020:ch6}
\nocite{wang2018hybrid}
\nocite{miikkulainen2019evolving, Peng2018, Nakisa2018}
\nocite{rawal2016evolving, akut2019neuroevolution, 10.1007/978-3-030-50417-5_25}
\nocite{bento2018short, Nakisa2018}
\nocite{charte2020evoaaa, van2018evolutionary, charte2019automating}
\nocite{sun2018particle} 
\nocite{papa2015model, passos2019metaheuristic}
\nocite{Kuremoto, Passos2018, wang2019deep}
\nocite{passos2019metaheuristic}
\nocite{Sabar2017, hossain2017evolution, de2019soft}
\nocite{neuralEvolution2020:ch3}
\nocite{horng2017fine, Li2019, Goudarzi2018}
\nocite{Ma2017, neuralEvolution2020:ch5, Rodrigues2016}
\nocite{Garciarena2018,Lu2018}
\nocite{fujino2017deep,Dahou2019,Young2015}
\nocite{bingham2020evolutionary, 8790354, gonzalez2020improved}
\nocite{lorenzo2017particle,Yamasaki2017}
\nocite{ortego2020evolutionary}
\nocite{ElSaid2018,ElSaid2018a, Wang2015a}
\nocite{silhan2019evolution}
\nocite{Sabar2017, Papa2016, Papa2017}
\nocite{Li2019,Rosa2012}
\nocite{Dahou2019, ul2018optimising}
\nocite{Holzinger2016, DeRosa2018}
\nocite{guo2019tabu_genetic}
\nocite{Papa2016, Sabar2017, Papa2017}
\nocite{ismail2019evolutionary}
\nocite{horng2017fine, Li2019, Rosa2012}
\nocite{Jaderberg2017}
\nocite{verbancsics2013generative, Pawelczyk2018, rere2016metaheuristic}
\nocite{fedorovici2012embedding,Garcia2020hybrid,zhang2018gadam}
\nocite{cui2018evolutionary}
\nocite{Zang2018, banharnsakun2018towards, Khalifa2018}
\nocite{li2019ea}
\nocite{Nawi2015}
\nocite{Alvernaz2017,david2014genetic}
\nocite{levy2014genetic}
\nocite{al2018towards}
\nocite{poulsen2017dlne,Alvernaz2017,khadka2018evolutionary}
\nocite{khadka2019collaborative,NIPS2018_7395,koutnik2014evolving}
\nocite{verbancsics2013generative,akut2019neuroevolution,miikkulainen2019evolving}
\nocite{Pawelczyk2018, rere2016metaheuristic, rere2015simulated}
\cite{de2019detection,ayumi2016optimization,rosa2015fine}
\nocite{Zang2018, banharnsakun2018towards, Khalifa2018}
\nocite{de2019detection}
\nocite{akut2019neuroevolution,miikkulainen2019evolving,rawal2016evolving}
\nocite{risi2019deep,Rashid2018,rashid2019improvement}
\nocite{van2009hierarchical,duchanoy2017novel,jana2019reconstruction}
\nocite{biswas2018bi, Angeline}
\nocite{ibrahim2018particle, 6900625, nawi2015weight}
\nocite{juang2018optimization}
\nocite{risi2019deep,Song2017}
\nocite{wang2019evolutionary,toutouh2019spatial}
\nocite{Stanley2005,Petroski2017,such2017deep,pham2018playing}
\nocite{hausknecht2014neuroevolution,stanley2002efficient,Igel2003}
\nocite{martinez2020simultaneously,mason2018maze,chrabaszcz2018back}
\nocite{etor2020adaptive}
\begin{figure}[h!]
\centering
\includegraphics[width=0.95\columnwidth]{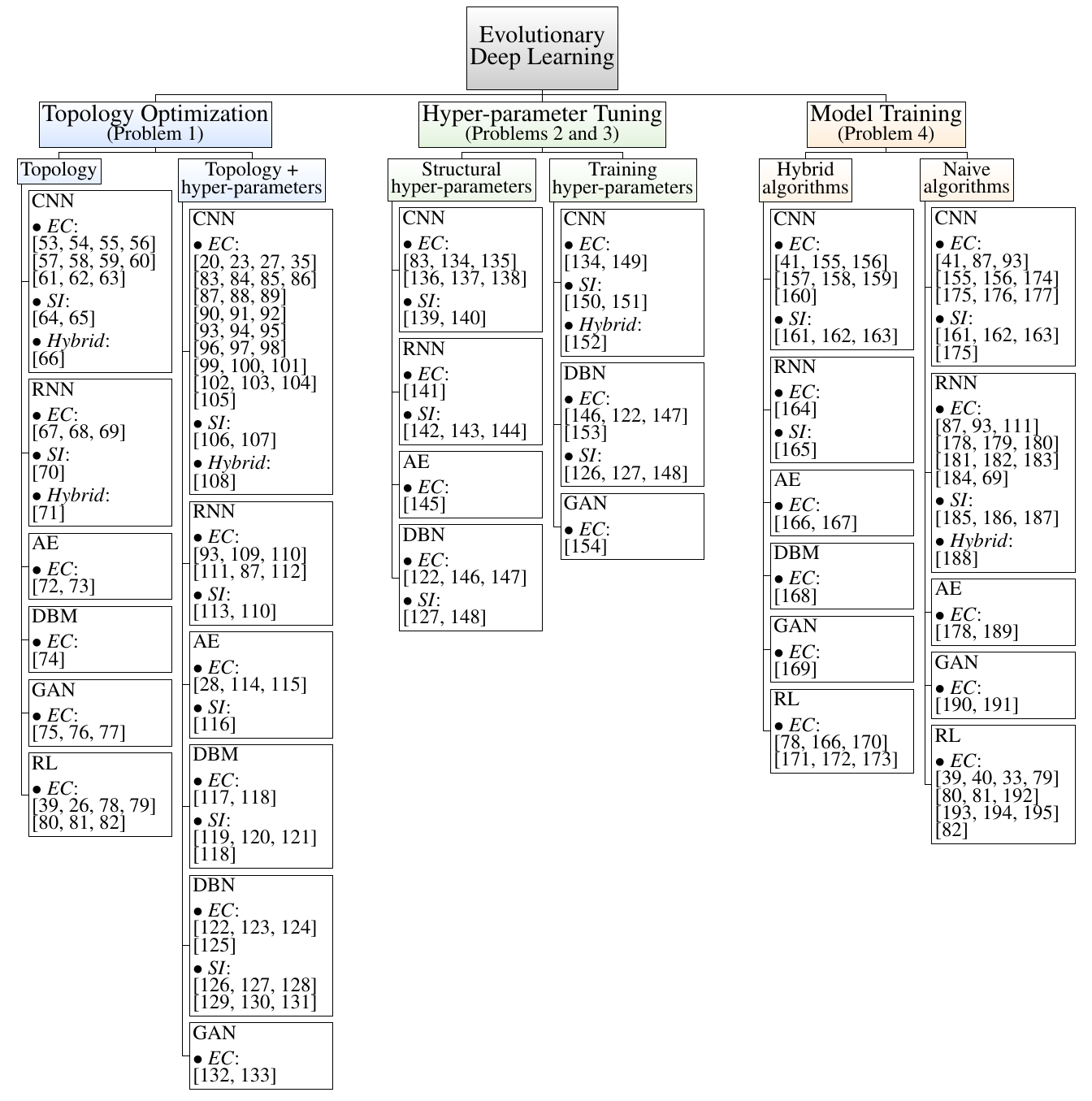}
\caption{Taxonomy of the reviewed literature on Evolutionary Computation and Swarm Intelligence algorithms applied to the optimization of Deep Learning models. The taxonomy is structured by the domain of the Deep Learning model under focus (topology, hyper-parameters and trainable parameters), further discriminated by the specific Deep Learning model under consideration and the type of bio-inspired algorithm in use (\emph{EC}: Evolutionary Computation; \emph{SI}: Swarm Intelligence; \emph{Hybrid}: a mixture of both).}
\label{fig:treeCat}
\end{figure}

Figure \ref{fig:treeCat} depicts graphically the taxonomy considered for the literature analysis. In the first level we consider the type of optimization problem under consideration (topology, hyper-parameter and trainable parameter optimization), followed by contributions sorted as per the Deep Learning model (\ref{ssec:DLmodels}) and kind of bio-inspired solver (\ref{ssec:metah}) under choice. In what follows we analyze in depth on the contributions classified within each of these categories.

\subsection{Topology optimization} \label{ssec:topology}

It is widely acknowledged that the topology or architecture of Deep Learning models have a direct impact on their performance. For this reason researchers have traditionally striven to develop automated methods for generating topologically small yet well-performing network architectures. In this context, the set of algorithms gathered under the \textit{Neuroevolution} (NE) label aim at progressively augmenting the complexity of neural network topologies to attain increasingly better generalization properties while keeping its complexity to its minimum required. Originally applied to  ANN models, different NE variants have been applied in the last few years to optimize Deep Learning models, not only in terms of their topology, but also jointly with their structural and training hyper-parameters (e.g. kernel size, activation function, dropout and learning rate). In some few cases, trainable parameters have also been considered in the set of variables to be optimized via NE \cite{verbancsics2013generative}. Furthermore, since they resort to evolutionary algorithms at its core, NE approaches have stimulated over the years a manifold of other bio-inspired approaches, in a way to assess whether the same optimization problem can be tackled more effectively with alternative search strategies and operators. 

All in all, in terms of topological optimization CNNs are arguably the most targeted Deep Learning models to date. CNNs' topology optimization is faced by scientific community in two ways; layer by layer or by blocks. In layer-wise optimization, hyper-parameters are fixed and networks are fully evolved using bio-inspired solvers, such as GA \cite{Xie} and customized versions of Evolutionary Algorithms \cite{real2018regularized}. In this last work, the so-called AmoebaNet-A model settled a state-of-the-art landmark score on the ImageNet dataset ($83.9\%$ accuracy), including comparisons to other search strategies (random search and reinforcement learning). Another approach proposed in \cite{wang2019particle} resorts to Particle Swarm Optimization (PSO) -- a popular Swarm Intelligence solver -- to optimize a block formed by dense layers. Once optimized, this block is stacked along with convolutional and pooling layers configured with fixed hyper-parameters, and ultimately used to address a image classification task. This work exemplifies a research trend focused on optimizing the topology of certain parts of the entire Deep Learning architecture, in an attempt at reducing the cardinality of the search space and speeding up the search process, at the cost of being much less exploratory in terms of network configurations than other counterparts. There is another important matter to be taken in account when performing topological optimization: the encoding of solutions, which impacts directly on the dimensionality of the search space. Actually, initial improvements of NE approaches were achieved thanks to novel network encoding strategies, which allowed for an easier exploration and less computational cost than preceding alternatives.

Another strand of literature has elaborated on more complex problem formulations by jointly addressing the optimization of the topology of the network along with its hyper-parameters. Again, CNNs have become central in related studies. An illustrative work is the one in \cite{Real2017}, where an Evolutionary Algorithm is used with different mutation operators operating on topological variables, structural and training hyper-parameters such as the filter size, the convolution stride, learning rate or the insertion/removal of convolutional layers, among others. Studies in this field tend to be similar to each other in terms of the complexity of the optimization problem under consideration. Thus, new proposals are usually made by customizing the operators (mutator, selector) of optimization algorithms or by developing custom encoding strategies, as in \cite{wang2018evolving} where a PSO variant is introduced based on an IPv4 based codification scheme with varying length.

Despite the predominance of CNNs in topological optimization, RNNs (and in particular, LSTM networks) have also been a subject of study in this research area, In \cite{Hu2018} a Differential Evolution (DE) solver is proposed for achieving this purpose and efficiently undertaking a wind forecasting regression task. It is relevant to observe that when both architecture and hyper-parameters are evolved for RNNs, certain hyper-parameters are recurrently considered in related studies, such as learning rate, dropping frequency factor \cite{bento2018short} or batch size \cite{Nakisa2018, Peng2018}. In general terms, the aforementioned DE appears to be the most applied meta-heuristics in LSTM. A few exceptions can be found, such as \cite{bento2018short} (Bat Algorithm), \cite{desell2015evolving} (Ant Colony Optimization) and \cite{Nakisa2018}, where a comparison is made between DE, PSO and SA, concluding that DE reaches better performance levels. Hybrid approaches have been also explored for the optimization of RNNs, as in \cite{juang2004hybrid} where the architecture (i.e. connection pattern) is optimized by means of a hybrid PSO-GA solver. An interesting point arises when inspecting in detail this set of studies: the creation of custom objective functions to allocate different (usually conflicting) optimization goals. The work in \cite{Hu2018} is an example of how a customized objective function can yield topologically optimized network designs that achieve a balance between performance and model complexity, being the latter of particular interest for the deployment of the model in resource-constrained embedded devices. 

Other Deep Learning models have grasped a remarkable attention in topological optimization. AEs have been optimized topologically, often along with structural hyper-parameters in those cases where convolutional layers are involved. In their original formulation, AEs are composed by stacked dense layers (\emph{encoder}) producing a low-dimensional representation of the input, which is reconstructed by another set of stacked dense layers (\emph{decoder}). A contribution from 2015 \cite{lander2015evoae} presented a way to generate promising AE architectures by mutating candidates by using a customized Evolutionary Algorithm, whose mutation operator is based on the reconstruction error achieved by the decoder. Also in this vein, a mini-batch variant training method was proposed (\emph{evo-batches}) aimed at reducing the computational cost when a large number of candidate networks have to be evaluated in large datasets. Decoder and encoder topologies of studies related to AEs are often assumed to be symmetrical \cite{lander2015evoae}. However, in \cite{assuncao2018automatic} a more flexible architecture was proposed, where the decoder is evolved along with the encoder and does not have to mimic its architecture. In this reference several operations were applied to topological variables, such as layer addition (random number of neurons), layer removal, application of Gaussian perturbation to the number of neurons, or layer swapping. To wrap up the activity noted in AEs, we highlight the work in \cite{sun2018particle} and \cite{van2018evolutionary}, where PSO and GA are respectively used to evolve topology and structural hyper-parameters of AEs comprising convolutional layers.

We proceed forward with our analysis by pausing at DBMs, whose architecture can be very similar to DBNs in terms of the structural hyper-parameters involved in the optimization process. Given this similarity and the relative scarcity of studies related to these models, we analyze them jointly in what follows. The majority of works related to the topological optimization of DBMs and DBNs pay special attention to the process of optimizing both architecture and some hyper-parameters. Moreover, most of them rely on Swarm intelligence algorithms, such as PSO or ACO. An exception can be found in \cite{horng2017fine} where Artificial Bee Colony (ABC) is used to optimize DBNs' structure, learning rate, momentum and weight decay. The results are compared to those yielded by other bio-inspired solvers: Firefly Algorithm (FA), CS and Harmony Search (HS). FA for 2- and 3-layered DBN, and ACO for single-layer DBN, resulted to yield the best performing network architectures for the image reconstruction task under consideration. The rest of contributions consider a combination of all or some of learning rate, momentum or weight decay hyper-parameters, focusing the application of the optimized model to different practical problems such as traffic flow prediction \cite{Li2019} or the detection of turbine failures \cite{Ma2017}. The work in \cite{Passos2018} introduces a novel way to optimize structure and hyper-parameters of DBMs, and compares the performance of DBMs for image classification when optimized with different flavors of the PSO solver, random search and several HS variants. They concluded that bio-inspired techniques are suitable to optimize DBMs, beating Random Search in all considered datasets. Nevertheless, network topologies optimized via PSO and HS solvers scored similar performance levels. This last observation connects directly with one of the points remarked in our critical analysis of Section \ref{sec:critic}.

In terms of algorithmic variants, some hyper-heuristic techniques like \cite{Sabar2017} proposed an approach to optimise DBNs' structural hyper-parameters, i.e. number of hidden units, along with non-structural hyper-parameters like learning rate, and some hyper-parameters related to the heuristic algorithm (number of epochs or iterations). Hyper-heuristics have also been used to optimize CNNs, \cite{ul2018optimising} presents a method to select the best heuristics, where batch size, number of epochs, neurons on the fully connected layer, dropout and learning rates, rho and epsilon factors are evolved. 

There are also a few works dealing with the topological optimization of GANs. In \cite{Garciarena2018}, a meta-heuristic approach to evolve GANs' discriminator and generator is introduced. Specifically, a GA is used to evolve the architecture, activation functions of each layer and initialization mode in both generator and discriminator. Furthermore, an optimization of the loss function is done, taking them from a bunch of well-known formulations. Besides, training hyper-parameters are also mentioned in this work as potentially evolvable variables (yet not optimized in practice), such as the gradient-based solver used to learn the trainable parameters, the batch size and the number of epochs. Shortly thereafter, Costa et al. \cite{costa2019coevolution} proposed an approach to optimize the architecture and parameters of both the generator and discriminator modules of a GAN. The approach was based on DeepNEAT and adapted to the context of GAN optimization. Linear, convolutional and transpose convolutional layers were directly mapped to a phenotype -- an array of genes -- representing the final network. In all layers the activation function was evolved, and in the case of convolutional and transpose convolutional layers, the output channels were also considered.

Finally, in the field of RL, the tendency observed in our literature analysis is to use NE approaches to optimize both topology and trainable parameter (weights) of the neural network mapping the output of the environments to the actions to be taken by the agent. Commonly, NEAT is used for this purpose \cite{hausknecht2014neuroevolution, pham2018playing, poulsen2017dlne, stanley2002efficient}, which becomes in charge of optimizing the neural network involved in Deep RL approaches. A real-time adaptation of NEAT was used in \cite{Stanley2005} to evolve agents for the NERO videogame, placing an emphasis on the need for efficient workarounds to alleviate the complexity of neuro-evolution methods. 

On a summarizing note, the literature on bio-inspired algorithms applied to architecture optimization has a long history departing from NE, which was originally applied to evolve ANN architectures. Since then, many other meta-heuristics have been applied to optimize architecture and hyper-parameters of Deep Learning models. Given that networks can have variable-length topologies, a good solution encoding strategy is essential to lessen computational costs and the time of execution without hindering the representability of all network configurations. Remarkably, modern bio-inspired solvers such as FA, BA and CS have been lately used with competitive results with respect to classical solvers (EA, PSO and ACO).

\subsection{Hyper-parameter optimization} \label{ssec:hyp}

Arguably, one of the optimization tasks where bio-inspired methods have been traditionally applied within the Machine Learning field is hyper-parameter tuning. It is well-known that hyper-parameter tuning usually yields better performance levels than off-the-shelf model configuration. When shifting the focus towards the hyper-parametric optimization of Deep Learning models, two major fields are spotted. On the one hand, literature focused on optimizing parameters related to the training algorithms, such as \textit{learning rate}, \textit{batch size} or \textit{momentum}. On the other hand, architectural hyper-parameters, which are layer-type-dependent, e.g. filter size, number of kernels, activation functions or stride size in CNNs. Actually, given the high number of structural hyper-parameters of convolutional layers, CNNs have protruded as one of the most explored Deep Learning models for hyper-parameter optimization, in many cases jointly with training hyper-parameters such as the learning rate, momentum or weight decay of gradient solvers. Although there are some contributions focused exclusively on the optimization of structural hyper-parameters, the mainstream is to jointly address the optimization of topology and hyper-parameters, as the literature on topological optimization examined in the previous subsection has clearly revealed.

Let us start from 2016, when \cite{Holzinger2016} proposed a set of bio-inspired meta-heuristics (BA, FA and PSO) to optimize the aforementioned hyper-parameters in a CNN used for Parkinson disease identification from image data. Likewise, \cite{Dahou2019} proposed a DE-based approach to optimize filter sizes, number of neurons in the fully-connected network end, weight initialization policy and dropout rate for sentiment analysis. Comparisons were made to GA and PSO, achieving better results in terms of accuracy and computational efficiency. The optimization of dropout probability is other common approach that some authors have tackled using different solvers, including CS, BA, FA and PSO \cite{DeRosa2018} or hybrid GA and Tabu Search algorithms \cite{guo2019tabu_genetic}. In this latter work random search and Bayesian optimization were proven to perform worse than the proposed hybrid meta-heuristic algorithm over the considered image classification datasets. Batch size and learning rate were also regarded as optimization variables. 

Moving on to RNNs, very few papers are focused only on hyper-parameter optimization, conforming to the general trend observed in the analyzed literature. Dropout optimization is tackled for LSTM networks in \cite{ElSaid2018} and \cite{ElSaid2018a}, where ACO is used for engines vibration prediction. The connections between neurons are activated/deactivated to accomplish the task, which is very similar to the approach carried out in \cite{Wang2015a}. However, the aspect to be highlighted in this last reference is that a PSO is utilized to optimize the output connections of an Echo State Network (ESN), a randomization-based RNN model belonging to the family of Reservoir Computing models. We will later revolve on the possibilities that we have found in the extrapolation of advances in bio-inspired optimization for Deep Learning models to other less studied models.

\subsection{Trainable parameter optimization} \label{ssec:trainable}

In recent years the advent and progressive maturity of new parallel and distributed computing paradigms have reignited the interest of the scientific community in applying bio-inspired optimization algorithms to train Deep Learning models. Although each model type is different in terms of topology, they share some disadvantages resulting from the adoption of gradient back-propagation training methods, such as gradient vanishing/exploding and proneness to get stuck in local optima. Evidently, the complexity of the problem grows up as the number of parameters involved in the optimization process increase, yielding largely non-convex search landscapes. These acknowledged issues have been extensively studied by the community by proposing different workarounds. Nonetheless, an increasing trend towards the use of bio-inspired solvers for this purpose can be noticed in recent literature, as their search operators do not rely on gradient back-propagation anyhow, and therefore avoid its drawbacks effectively. Based on this rationale, we now delve into how the community has adapted different bio-inspired solvers for training Deep Learning models.

A first examination of the literature exposes two main tendencies followed by the community, which are reflected in the second level of the corresponding taxonomy branch of Figure \ref{fig:treeCat}:
\begin{itemize}[leftmargin=*]
\item Approaches that combine bio-inspired solvers with traditional training algorithms, which aim to overcome the disadvantages we have just introduced. Almost the entirety of studies adopting this hybrid design strategy are focused on CNNs, implementing the aforementioned combination in many different ways. A straightforward way to overcome falling in local optima is to evolve an initial set of values for the trainable parameters (weights, bias) that sets the gradient back-propagation solver on a promising path towards the global minimum of the loss function. In \cite{banharnsakun2018towards} this approach is adopted for training a CNN using the ABC algorithm. Other works \cite{Pawelczyk2018} combine GA and SGD: GA evolves new candidates through its search operators, but the fitness function is evaluated after some training epochs of stochastic gradient back-propagation (SGD). Similarly, in \cite{Khalifa2018} PSO is used to evolve the trainable parameters of the last layer of a CNN, while the parameters of the rest of the layers are learned via SGD. Comparisons with the CNN trained exclusively with SGD rendered an enhanced convergence speed and final accuracy on image classification tasks. Last but not least, in \cite{Nawi2015} the CS algorithm is used to train RNNs following two strategies: one trained using only this solver, and the other combining CS and gradient back-propagation. A benchmark comparison to networks trained with conventional gradient back-propagation and different variants of the ABC algorithm discovered that all models trained using bio-inspired optimization techniques performed better than the RNN trained with gradient back-propagation. 

\item Approaches in which training is performed completely using bio-inspired optimization methods. Most references embracing this second strategy deal with RNNs and CNNs, and differ from each other mostly in the search algorithm being considered. All in all, a common approach is to evolve the trainable parameters of the model (via the search operators of the bio-inspired solver at hand), and evaluate it in terms of loss value or any other performance estimator linked to the task at hand. In this line, in \cite{rere2015simulated} and \cite{ayumi2016optimization} two SA-based solvers are proposed and assessed for optimizing the parameters of a CNN, achieving better performance scores and better convergence speed than the same model trained via gradient back-propagation. LSTM network training has also been tackled by using different bio-inspired optimization techniques. A good exponent is the work in \cite{Rashid2018}, where HS, Gray Wolf Optimizer (GWO), Sine Cosine Algorithm (SCA) and Ant Lion Optimization  (ALO) were compared to each other when used to learn the trainable parameters of different LSTM model configurations. We emphasize that despite the diversity of methods considered in this work, no comparisons to traditional training solvers were reported, uncovering one of the critical points discussed in Section \ref{sec:critic}.
\end{itemize}

Before proceeding with this critical analysis, we briefly comment on Deep RL models. Bio-inspired algorithms have been lately postulated as efficient alternatives to solve several optimization problems underlying these models. A first approach is to train parts of the architecture via evolutionary algorithms, and the rest using SGD. This is indeed what the study in \cite{Alvernaz2017} proposes: a Covariance Matrix Adaptation Evolution Strategy (CMA-ES) is used to evolve weights for the behavior-generating network, while the rest of the Deep RL architecture (a convolutional AE) is trained by using a gradient based method. This work continued the research line started years before in \cite{Igel2003}, where CMA-ES was used to train simpler RL networks. All in all, Evolutionary Algorithms have largely demonstrated to be efficient solvers to train Deep RL models, as shown in renowned studies such as \cite{such2017deep, Petroski2017} (with GA), \cite{such2017deep} (GA with Novelty Search) and \cite{mason2018maze} (DE and Novelty Search). Recent works \cite{martinez2020simultaneously} have also explored the capabilities of multi-task optimization evolving multiple related RL tasks at the same time, taking advantage of the transfer of genetic information between tasks. In other works, NEAT is used to optimize the trainable parameters and the topology of a Deep RL model \cite{hausknecht2014neuroevolution, pham2018playing}. On the other hand, in \cite{khadka2018evolutionary, khadka2019collaborative} an hybrid EA algorithm is proposed, where a population of networks is trained and periodically evolved, exploiting Lamarckian transfer capabilities. This same procedure is used in \cite{NIPS2018_7395} as a way to inject information about the gradient in the population of individuals maintained by the Evolutionary Algorithm during the search. Other hybridization techniques consist of the use of different networks in the same architecture, where some of them are trained via SGD, and others evolved with evolutionary operators. This is the case of \cite{koutnik2014evolving}, where the parameters of a RNN used for determining the actions of an agent are evolved using Cooperative Synapse Neuroevolution (CoSyNE), an evolutionary algorithm that enforces subpopulations at the level of a single trainable parameter. This work built upon the findings in \cite{gomez2008accelerated} and extrapolated them to complex Deep RL models, showcasing how evolutionary algorithms can evolve small networks capable of reaching competitive performance levels.

\section{Critical Methodological Analysis} \label{sec:critic}

The above corpus of reviewed literature sheds evidence on the vibrant activity of the intersection between bio-inspired optimization and Deep Learning. So far the community has reported interesting findings in what refers to hyper-parameter optimization, topological search and small/medium-sized network training. Notwithstanding this noted activity, our critical analysis of these contributions has disclosed a number of poor practices and methodological shortages that should be underlined to set them down in black and white. We now discuss briefly about these issues, settling the necessary rationale for the experimental part of this survey:
\begin{itemize}[leftmargin=*]
\item The lack of benchmark datasets/tasks to validate new advances (Subsection \ref{ssec:lack_benchmarks}).
\item The unrealistic scales of the Deep Learning models optimized via bio-inspired methods (Subsection \ref{ssec:unrealistic_scales}).
\item The need for good methodological practices when comparing among different solvers used for Deep Learning optimization (Subsection \ref{ssec:methodological}).
\item The limited utility of software implementations (Subsection \ref{ssec:software_limited}).
\item The existence of metaphor-based publication series (Subsection \ref{ssec:metaphors}). 
\end{itemize}

\subsection{Lack of benchmark datasets/tasks} \label{ssec:lack_benchmarks}

A major problem observed in the literature is the heterogeneity of datasets used to validate new algorithmic approximations for the optimization problems under analysis. Even when the task is clearly defined (e.g. image classification or time series forecasting), the possibility to compare the results obtained by different studies becomes unfeasible since the considered datasets are not the same. 

For the community to gain verifiable evidence about the claimed gains of upcoming proposals, consensus should be reached about the datasets/tasks that should be utilized for comparison purposes in the future. Unfortunately, the diversity of datasets/tasks over which some of the new contributions are assessed seem to go in the opposite direction, calling into question whether the reported performance improvements can be extrapolated to other learning problems.

\subsection{Unrealistic complexity of Deep Learning models} \label{ssec:unrealistic_scales}

Besides the heterogeneity of datasets/tasks discussed above, it is often the case that the Deep Learning models under consideration do not meet the complexity levels of the state of the art for the task under consideration. This is particularly concerning in works related to model training (Problem 4), where the cardinality of the search space faced by the meta-heuristic solver is in the order of several thousands to millions of optimization variables (trainable parameters). For instance, a recent work on image classification using the well-known MNIST dataset has recently established a new record in accuracy (0.1\% test error rate) with a model comprising 57.02 millions of trainable parameters \cite{tabik2020mnist}. However, most of the reviewed literature on training via bio-inspired solvers rarely considers Deep Learning models that surpass a few thousands of trainable parameters. 

This issue again calls for a major reflection on whether research advances are missing the real challenge underneath the use of bio-inspired algorithms in such large search spaces (scalability, exploitation of the correlation among decision variables), to instead focus on minor aspects of doubtful scientific contribution.

\subsection{Comparison methodology} \label{ssec:methodological}

Even if addressing and effectively solving the preceding two issues, several methodological aspects still remain often overseen when comparing among different solvers for a given task/dataset/optimization problem scenario:
\begin{itemize}[leftmargin=*]
\item Baseline schemes: our literature analysis revealed that a fraction of contributions discussed on extensive experiments with several new meta-heuristic algorithms for a given optimization problem, without including in the benchmark standard solvers utilized in the past for the same problem. This, again, is particularly worrying in regards to Problem 4 (model training): comparisons should compulsorily include gradient back-propagation based solvers widely used for the same purpose (e.g. SGD, Adam). Overlooking the analysis of whether bio-inspired algorithms perform competitively with respect to established solvers for the same purpose is counterproductive for the potentiality of this research area. 

\item Objective function(s): we noticed that relevant divergences emerge in how the optimization algorithms proposed over the years are guided when attempting to solve a given optimization problem. For instance, a common practice is to reserve a validation data subset over which a measure of performance related to the task at hand is computed (e.g. accuracy in image classification). This measure is used as the objective function guiding the search of the proposed optimization algorithm. However, depending on whether this validation subset is kept fixed or shuffled, a partition bias might affect the generalization capabilities of the evolved network, specially when dealing with small datasets. 

On a similar reasoning, when dealing with imbalanced datasets the standard definition of accuracy is known to be not adequate to quantify the performance of the model in the minority class, and could exacerbate further the aforementioned problems. In what refers to Problem 4 (trainable parameter optimization), this issue becomes even more serious because derivatives of the loss function are not needed any longer, hence widening the portfolio of possible objective functions. To date, there is no clear answer whether differentiable loss functions should be selected at the objective function of bio-inspired optimization algorithms used for model training, or, instead, alternative task-dependent objectives should be formulated.  

\item Parameter tuning of solvers: additional issues arise in how different solvers are compared to each other for a given optimization problem. To begin with, it is often the case that no evidence is provided about the optimality of the parameters controlling the behavior of the search algorithm itself. In the research community working in bio-inspired optimization, it is largely accepted that a good parameter tuning is crucial for ensuring fair comparisons among algorithms \cite{latorre2020fairness}. Since the objective function evaluation of candidates in problems related to Deep Learning is usually costly in terms of computational effort, the parameters of the bio-inspired algorithm are often set equal to values retrieved from past works, or conforming to common practice. This unfairly biases the discussion, usually leaving unclear whether the reported performance gaps are incidental. We acknowledge that the research record noted around the usage of hyper-heuristics \cite{Sabar2017,ul2018optimising} avoids to an extent this comparison bias, but the problem still prevails in most works proposing new bio-inspired methods.

\item Assessing the impact of randomness on the results: another methodological aspect that has not been properly considered in most related literature is the fact that several sources of randomness can collide into a optimization problem. For instance, in Problems 1, 2 and 3 not only the search algorithm comprises a number of stochastic operators, but the training algorithm in use can also induce randomness in the obtained results. For instance, the values of trainable parameters optimized by the SGD solver depends on the composition of the mini-batches over which gradient estimates are computed. Such mini-batches comprise a number of examples from the training set, which is usually shuffled between successive epochs. This source of randomness could justify training the optimized network several times (\emph{runs}) and aggregating the results for a more reliable fitness computation. Otherwise, this should be conceived as an additional factor motivating a proper statistical assessment of the significance of performance gaps between different optimization algorithms, adding to the randomness induced by their search operators. 

Surprisingly, only a few exceptions have embraced the usage of statistical tests for this purpose, leaving most of the experiments reported in this field dubious and inconclusive. Furthermore, experiments with several datasets, tasks and optimization algorithms should embrace the methodological practices for multiple comparisons deeply rooted on the scientific community, such as critical distance plots \cite{demvsar2006statistical} or Bayesian tests \cite{carrasco2020recent}.

\item Reproducibility of results: although this is a claim that emerges in almost any field of research, the need for reproducibility becomes particularly pressing in this area. Reasons go far beyond the verification of the contribution reported in emerging studies: the community can expedite the achievement of novel advances in the field if the software and experimental results of preceding works are shared within the community. The current status in this matter is not that concerning, as many open-source software frameworks exist with the functionalities required to develop new approaches and experiments. Unfortunately, many contributions published lately still do not provide any access to the software implementing their proposals. When given, it is also frequent to see that the source code is made ad-hoc for the problem at hand, thereby dilating the time needed for building new proposals on top of existing ones.
\end{itemize}

\subsection{Software implementations of limited practical utility} \label{ssec:software_limited}

Nowadays, Deep Learning architectures scoring competitively in tasks defined over real-world datasets are usually composed by millions of trainable parameters. When addressing Problem 4 (trainable parameter optimization), the huge search space faced by the optimization algorithm under consideration requires intelligent means to exploit the relationships existing among the optimization variables. Actually, gradient based solvers adopt this strategy by the back-propagation of the gradients throughout all layers compounding the Deep Learning model, which gives rise to an implicit mechanism to exploit the correlations between the optimization variables. In this context, there is an entire research area dedicated to the large-scale global optimization, plenty of algorithmic proposals where synergies among optimization variables are exploited by assorted means. Nonetheless, many works still revolve on the naive application of standard bio-inspired solvers, neglecting such interactions between variables. 

Further along this line, we note that the vibrant activity of the area is not in accordance with the ultimate goal for which bio-inspired algorithms are being proposed for training Deep Learning models. As told by the currently available implementations in the literature, the computational cost of population-based meta-heuristics is enormous, and yields far longer training times than off-the-shelf gradient back-propagation approaches. Even if the software implementation of algorithmic proposals reported to date may have been restricted to experimental settings, a complexity analysis should have been performed to accounting for both their benefits and drawbacks, so that the community can delimit realistic boundaries for the practical utility of these advances. 

\subsection{Metaphor-based publication series} \label{ssec:metaphors}

Last but not least, we emphasize on the claims of recent studies about the justification of new meta-heuristic algorithms just by the biological metaphor in which it is allegedly inspired \cite{del2019bio}. In our literature review we identified several publication series in which the same optimization problem were tackled by different bio-inspired solvers within a short time period. By no means these contributions provide any scientific value for the general knowledge of the field, even if publication workflows run at different speeds. 

Disregarding the reasons for these practices, results with optimization algorithms inspired by different biological behaviors and phenomena should not be shattered over different short-elapsed publications. They can provide much more valuable insights when presented and discussed together.

\section{Case of study I: Architecture and Hyper-parameter Optimization of Deep Learning Models with Bio-inspired Algorithms} \label{ssec:exp1}

In this first case of study we focus on the topology and structural hyper-parameter optimization of Deep Learning models, which has been an open challenge in the last few years. In particular, this experimental study focuses on finding the best CNN architecture along with their structural hyper-parameters for solving image classification tasks. We focus the case of study on two recent frameworks:
\begin{itemize}[leftmargin=*]
\item EvoDeep, proposed in \cite{Comput2018} as an evolutionary algorithm specially designed to optimize both hyper-parameters and architecture of a Deep Learning model, selecting the type and size of layers compounding the evolved architecture.
\item AutoKeras \cite{jin2019auto}, which is another modern AutoML systems built on top of the renowned Keras Python library, that is able to automatically generate highly-optimized neural network.
\end{itemize}

Taking into account the recent activity noted in this area, we herein compare and discuss on the performance of these two consolidated frameworks for topology and hyper-parameter optimization of Deep Learning models. The comparison is made using three important measures: accuracy, time and model complexity. 

To this end, in what follows we describe EvoDeep (Subsection \ref{ssec:evodeep}) and AutoKeras (Subsection \ref{ssec:autokeras}), present the designed experimental setup (Subsection \ref{ssec:experimental_setup_I}). The obtained results are discussed in Subsection \ref{ssec:discussion_I}, analyzing them in terms of accuracy, computation time and model complexity.

\subsection{EvoDeep: Evolutionary Computation for Deep Neural Network topology and hyper-parameter tuning} \label{ssec:evodeep}

EvoDeep is a framework based on an evolutionary algorithm that follows a $(\lambda +  \mu)$ strategy, where $\lambda$ indicates the number of new individuals produced at each generation, and $\mu$ represents the number of individuals selected for the next generation. Each individual of the population is a network architecture with its respective hyper-parameters. The fitness for each individual is the accuracy of the neural network when solving a classification problem. At every generation, the recombination and mutation operators are applied to generate a new individual for the next generation. 

EvoDeep finds increasingly better neural network architectures for CNN using elementary convolutional and fully connected layers. Specifically, EvoDeep evolves a network by using only the original data and an set of permitted layers (i.e. \texttt{Convolution2D}, \texttt{Flatten}, \texttt{MaxPooling2D}, \texttt{Reshape}, \texttt{Dense} and \texttt{Dropout} as per Keras notation). The number of layers represents the depth of the evolved network, and its search range is set to $[3,20]$ with a step size of $1$ layer. Moreover, the training hyper-parameters can be also specified in its configuration. These parameters are as follows:
\begin{itemize}[leftmargin=*]
    \item Optimizer, chosen among \texttt{Adam}, \texttt{SGD}, \texttt{Rmsprop}, \texttt{Adagrad}, \texttt{Adamax} or \texttt{Nadam}.
    \item Number of epochs: an integer value in the range $[2,20]$, with a step size of $2$ epochs.
    \item Batch size: this value has a range of $[100,5000]$ with a step size of $100$ examples.
\end{itemize}

Another characteristic of EvoDeep is the fact that it requires the data in a specific manner. A two-dimensional matrix is the input for the algorithm. The number of row matches the number of examples of the database and the number of columns is the size of the images. With size of the image we mean the product of width, height and channels (three channels for RGB and one for grayscale images). Unfortunately, EvoDeep is more oriented towards optimizing grayscale rather than RGB images.

EvoDeep allows the user to specify the parameters of its evolutionary strategy. Table \ref{tab:evodeep_config} shows the parameters and values used in this case of study.
\begin{table}[ht]
    \centering
    \caption{Parameter configuration set for the EvoDeep framework.}
    \resizebox{\linewidth}{!}{\begin{tabular}{|C{2cm}|C{3cm}|C{3cm}|C{2cm}|C{2cm}|C{3cm}|C{4cm}|C{1.5cm}|}
        \toprule
        \textbf{Parameter}  & $\lambda$  & $\mu$ & \texttt{cxpb}, $p_c$ & \texttt{mutpb}, $p_m$ & \texttt{newpb} & \texttt{ngen}, $n_{gen}$ \\ \midrule
        \textbf{Description} & 
        Number of newly produced individuals per generation &
        Number of selected individuals for the next population &
        Crossover probability &
        Mutation probability &
        Probability of adding a new layer to the network &
        Number of generations \\
        \midrule
        \textbf{Value}      & 10         & 5     & 0.5   &  0.5  & 0.5   &   20\\\bottomrule
    \end{tabular}}
    \label{tab:evodeep_config}
\end{table}

\subsection{AutoKeras: an AutoML reference} \label{ssec:autokeras}

AutoKeras is a software that also allows to finding the best neural network architecture for a given data set and task \cite{jin2019auto}, offering several search engines for this purpose. In our study, the version of AutoKeras used is 1.0.2, which features the following search methods:
\begin{itemize}[leftmargin=*]
    \item Random: it performs a random search of the models in relation to their depth and layer type.
    \item Greedy: it groups the parameters into several categories. For each category, the tuner uses a greedy strategy to generate new values for the hyper-parameters and to generate new values for one of the categories of hyper-parameters. It uses the best trial so far for the rest of hyper-parameter values.
    \item Hyperband: departing from a given model, this bandit-based algorithm searches the best hyper-parameter values for this model by running several random configurations for a scheduled number of iterations, using earlier results to retain good candidate configurations that are evaluated for longer runs.
    \item Bayesian optimization: the search space is explored using morphing. This optimization method is based on the three basis of Bayesian optimization: update, generation and observation.
    \item Task-specific: AutoKeras tries with a task-specific tuner of the model, and evaluates the most commonly used models for the task at hand, which in our case is image recognition. This is the default configuration. AutoKeras has two trial blocks: Vanilla and ResNet. The best model is taken from one of these two.
\end{itemize}

\subsection{Experimental setup} \label{ssec:experimental_setup_I}

In our study we have chosen 4 diverse and representative data sets according to their complexity: Horses or Humans (HORSEHUMAN \cite{horsesorhumans}); the \texttt{vangogh2photo} dataset from the so-called \texttt{cycle\_gan} repository (VANGOGH \cite{cyclegan}), MNIST \cite{lecun-mnisthandwrittendigit-2010} and CIFAR-10 \cite{cifar10dataset}. On the one hand, Horses or Humans and Van Gogh or Photo have two classes, but Van Gogh or Photo is unbalanced (only 400 images belonging to the minority class). On the other hand CIFAR-10 and MNIST are more complex databases in terms of number of examples and classes: both databases have 10 different classes. Moreover, in our experiment we have considered both color (RGB) and grayscale versions of these datasets so as to assess the influence of the color space in the complexity and performance of the models evolved by the compared frameworks.
\begin{table}[H]
\centering
\caption{Utilized datasets in Case of Study I.}
\label{tbl:datasets_use_case_I}
    \resizebox{\linewidth}{!}{\begin{tabular}{C{3cm}|c|C{1cm}|C{3cm}|L{8cm}}
    \toprule
         \textbf{Dataset} & \textbf{Shape} & \textbf{\# classes} & \textbf{\# Instances (train/test)} & \multicolumn{1}{c}{\textbf{Characteristics}} \\
        \midrule
        HORSEHUMAN       & $300\times 300\times 3$ & 2 & 1,027 / 256 & Balanced dataset, binary classification, RGB \\
        HORSEHUMAN-G   & $300\times 300\times 1$ & 2 & 1,027 / 256 & Balanced dataset, binary classification, grayscale \\
        VANGOGH        & $256\times 256\times 3$ & 2 & 6,687 / 1,151 &  Imbalanced dataset, binary classification, RGB \\
        VANGOGH-G        & $256\times 256\times 1$ & 2 & 6,687 / 1,151 &  Imbalanced dataset, binary classification, grayscale \\
        CIFAR-10    & $32\times 32\times 3$ & 10 & 50,000 / 10,000 &  Balanced dataset, multi-class classification, RGB \\
        CIFAR-10-G    & $32\times 32\times 1$ & 10 & 50,000 / 10,000 &  Balanced dataset, multi-class classification, grayscale \\
        MNIST       & $28\times 28\times 1$ & 10 & 60,000 / 10,000 &  Balanced dataset, multi-class classification, grayscale \\
        \bottomrule
    \end{tabular}}
\end{table}

In order to account for the statistical behavior of the frameworks under comparison, we have carried out 5 independent runs of AutoKeras, EvoDeep and random models, from which we have chosen the best model in terms of its accuracy in validation. Random models are obtained using EvoDeep without the evolutionary process. The number of tested individuals is the number of evaluations that EvoDeep would make should the search be done via its evolutionary strategy. The training phase has been done by splitting the training set as follows: 80\% for training and 20\% for validation. After that, the model is trained again with the whole training set, and evaluated on the test set.

\subsection{Results and discussion} \label{ssec:discussion_I}

Table \ref{tab:results_acc_casestudy1} shows the obtained accuracy scores for models evolved with EvoDeep, AutoKeras and random search over the color and grayscale datasets under consideration. The best results for each dataset are highlighted in bold. On the one hand, it is straightforward to observe that EvoDeep outperforms random models over both train and test datasets. However, AutoKeras still offers a better performance than EvoDeep. AutoKeras features the best test results in each dataset. To sum up, the results of EvoDeep are close to AutoKeras, which it is one of the best AutoML tools in this area. In Vangogh or Photo and MNIST the difference between them is less than 1\% in test and 2\% in Horses or Humans. Therefore, EvoDeep shows a great performance on these databases.

When shifting the scope towards the results obtained for the grayscale databases, similar conclusions can be drawn. In relation to the HORSEHUMAN dataset, the results issued by EvoDeep are similar to the previous ones in terms of accuracy over test, whereas AutoKeras improves its performance by 3\%. The results for CIFAR-10-G are worse when compared to the color ones. The exception in this trend is VANGOGH-G, due to the bad results obtained by the random models and EvoDeep. The accuracy in test decreases approximately 17\% in EvoDeep and 27\% in random models. By contrast, the results obtained for this dataset by AutoKeras are similar to the ones obtained for its colored counterpart. 
\begin{table}[H]
    \centering
    \setlength{\tabcolsep}{0.7em} 
    \caption{Results in terms of accuracy for color datasets and algorithms (in \%).}
    \resizebox{\linewidth}{!}{\begin{tabular}{c|ccc|ccc|ccc|cc}
    \cmidrule[1pt]{1-9}
    \multirow{2}{*}{\makecell{Color\\datasets}} & \multicolumn{2}{c}{HORSEHUMAN} & & \multicolumn{2}{c}{VANGOGH} & & \multicolumn{2}{c}{CIFAR-10} & & & \\
    \cmidrule{2-3} \cmidrule{5-6} \cmidrule{8-9}
    & \multicolumn{1}{c}{Train} & \multicolumn{1}{c}{Test} & & \multicolumn{1}{c}{Train} & \multicolumn{1}{c}{Test} & & \multicolumn{1}{c}{Train} & \multicolumn{1}{c}{Test} & & & \\
    \cmidrule{1-9}
    \textbf{Random}  & 89.19 & 89.06 & & 98.10 & 92.96 & & 73.06 & 53.63 & & & \\
    \textbf{EvoDeep} & 97.07 & 89.40 & & 100.00 & 98.87 & & 99.99 & 60.55 & & &\\
    \textbf{AutoKeras} & \textbf{100.00} & \textbf{91.40} & &  \textbf{100.00} & \textbf{99.48} & & \textbf{95.28} & \textbf{73.26} & & & \\
    \cmidrule[1pt]{1-12}
    \multirow{2}{*}{\makecell{Grayscale\\datasets}} & \multicolumn{2}{c}{HORSEHUMAN-G} & & \multicolumn{2}{c}{VANGOGH-G} & & \multicolumn{2}{c}{CIFAR-10-G} & & \multicolumn{2}{c}{MNIST} \\
    \cmidrule{2-3} \cmidrule{5-6} \cmidrule{8-9} \cmidrule{11-12}
    & \multicolumn{1}{c}{Train} & \multicolumn{1}{c}{Test} & & \multicolumn{1}{c}{Train} & \multicolumn{1}{c}{Test} & & \multicolumn{1}{c}{Train} & \multicolumn{1}{c}{Test} & & \multicolumn{1}{c}{Train} & \multicolumn{1}{c}{Test} \\
    \midrule
    \textbf{Random} & 94.06 & 89.84 & & 94.02 & 65.24 & & 73.44 & 49.35 & & 99.87 & 98.40 \\
    \textbf{EvoDeep} & 100.00 & 89.84 & & 96.66 & 81.58 & & 85.52 & 56.79 & & 100.00 & 98.69\\
    \textbf{AutoKeras} & \textbf{100.00} & \textbf{94.53} & & \textbf{100.00} & \textbf{99.30} & & \textbf{94.42} & \textbf{71.67} & & \textbf{99.99} & \textbf{99.40} \\
    \bottomrule
    \end{tabular}}
    \label{tab:results_acc_casestudy1}
\end{table}


After the comparison of EvoDeep, AutoKeras and random models in terms of accuracy, we now examine in Table \ref{tab:dataset_results_time} the execution time of the previous results, in minutes. A first inspection of the results in this table reveals that AutoKeras has the best performance with lower times than EvoDeep. We note that EvoDeep stops if no improvement is made over 5 consecutive generations. This is what occurs in MNIST: EvoDeep needs more computation time than AutoKeras and random models, but its accuracy results lie in between. This fact makes EvoDeep a reliable software because, even though it needs more computation time, the quality of the models in terms of predictive accuracy are close to those of AutoKeras in most cases. When it comes to grayscale datasets, in general the computation time is lower than the time taken by the experiments with color datasets. Runtimes of AutoKeras are almost the same except for HORSEHUMAN-G, in which the time is approximately $2.6$ times faster than that of HORSEHUMAN. EvoDeep and random models require less computation time in all the cases when compared to the colored ones. For example, VANGOGH-G is around $2.25$ times faster than the previous experiments.
\begin{table}[H]
    \centering
    \caption{Results in terms of time (minutes) for all datasets and frameworks.}
    \setlength{\tabcolsep}{0.5em}
    \resizebox{0.75\textwidth}{!}{
    \begin{tabular}{c|c|c|c|c|}
    \cmidrule[1pt]{1-4}
    Color & HORSEHUMAN & VANGOGH & CIFAR-10-G & \\
    \cmidrule{1-4}
    \textbf{Random}  & \textbf{5.5}  & \textbf{11} & \textbf{92} & \\
    \textbf{EvoDeep} & 22.0 & 50 & 322 &  \\
    \textbf{AutoKeras} & 13.5 & 14 & 110 & \\
    \cmidrule[1pt]{1-5}
    Grayscale & HORSEHUMAN-G & VANGOGH-G & CIFAR-10 & MNIST \\
    \cmidrule{1-5}
    \textbf{Random}  & \textbf{3.08}  & \textbf{7.14} & \textbf{43.44} & \textbf{46} \\
    \textbf{EvoDeep} & 13.17 & 22.25 & 295.99 & 230 \\
    \textbf{AutoKeras} & 5.19 & 13.98 & 109.98 & 262 \\
    \bottomrule
    \end{tabular}}
    \label{tab:dataset_results_time}
\end{table}

We now proceed in our analysis with Table \ref{tab:dataset_models}, where we compare random, EvoDeep and AutoKeras in terms of the complexity of the models produced over the search (in terms of the number of layers and the number of trainable parameters of the best model). As proven by the results included in this table, random model produces very similar network architectures across the colored datasets: they all comprise 5 layers with varying size (between $10^3$ and $2\cdot 10^6$ trainable parameters). EvoDeep has better results than AutoKeras in terms of number of layers for $3$ datasets, and in terms of the number of trainable parameters for $2$ datasets (HORSEHUMAN and MNIST). The best model for HORSEHUMAN is achieved by EvoDeep, comprising a simple neural architecture with 3 layers and approximately $1.9\cdot 10^6$ parameters. AutoKeras' model for the VANGOGH dataset has 9 layers and fewer trainable parameters in comparison to the models produced by the other frameworks. EvoDeep produces a model with only 5 layers for CIFAR-10, but AutoKeras' model has less parameters. Finally, for the MNIST dataset, EvoDeep discovers a better model than AutoKeras in terms of model complexity: fewer layers and parameters. To sump up, in terms of model complexity EvoDeep can be declared to perform better than AutoKeras, at least in some of the considered colored datasets.
\begin{table}[H]
    \centering
    \setlength{\tabcolsep}{0.5em} 
    \caption{Results in terms of model complexity for all datasets and algorithms.}
    \resizebox{\linewidth}{!}{\begin{tabular}{c|ccc|ccc|ccc|cc}
    \cmidrule[1pt]{1-9}
    \multirow{2}{*}{\makecell{Color\\datasets}} & \multicolumn{2}{c}{HORSEHUMAN} & & \multicolumn{2}{c}{VANGOGH} & & \multicolumn{2}{c}{CIFAR-10} & & & \\
    \cmidrule{2-3} \cmidrule{5-6} \cmidrule{8-9}
    & Layers & Parameters & & Layers & Parameters & & Layers & Parameters & & & \\
    \cmidrule{1-9}
    \textbf{Random}  & 5 & \textbf{179,922} & & \textbf{5} & 1,158,402 & & \textbf{5} & 2,270,100 &  & & \\
    \textbf{EvoDeep} & \textbf{3} & 1,895,012 & & 9 & 18,756,012 & & \textbf{5} & 10,821,230 & & & \\
    \textbf{AutoKeras} & 194 & 23,566,856 & & 10 & \textbf{31,944} & & 10 & \textbf{144,849} & & & \\
    \cmidrule[1pt]{1-12}
    \multirow{2}{*}{\makecell{Grayscale\\datasets}} & \multicolumn{2}{c}{HORSEHUMAN-G} & & \multicolumn{2}{c}{VANGOGH-G} & & \multicolumn{2}{c}{CIFAR-10-G} & & \multicolumn{2}{c}{MNIST} \\
    \cmidrule{2-3} \cmidrule{5-6} \cmidrule{8-9} \cmidrule{11-12}
    & Layers & Parameters & & Layers & Parameters & & Layers & Parameters & & Layers & Parameters \\
    \midrule
    \textbf{Random}  & \textbf{4} & 853,322 & & \textbf{3} & 318,372 & & \textbf{5} & 239,650 & & \textbf{5} & \textbf{211,245} \\
    \textbf{EvoDeep} & 7 & \textbf{129,182} & & 6 & 783,352 & & 12 & 1,883,220 & & 8 & 5,409,980 \\
    \textbf{AutoKeras} & 194 & 23,560,580 & & 10 & \textbf{31,364} & & 10 & \textbf{144,269} & & 194 & 23,579,021 \\
    \bottomrule
    \end{tabular}}
    \label{tab:dataset_models}
\end{table}

When analyzing the complexity of models discovered for the grayscale datasets, the results in Table \ref{tab:dataset_models} unveil a huge improvement in the number of parameters in random models and EvoDeep. Results by AutoKeras remain very similar to those for the colored datasets, with minimal differences in terms of the number of parameters. If we observe the number of layers, random models have almost the same number of layers as for the colored ones. However, more remarkable changes are noticed for EvoDeep. The models discovered for the HORSEHUMAN-G dataset has more layers, but significantly less parameters. The same statement can be said for CIFAR-10-G. The results for the VANGOGH-G dataset are the exception: the best model has fewer layers and fewer parameters. These observations support the aforementioned claims on the complexity of models produced by AutoKeras with respect to EvoDeep and Random.

The previous experiments have shown some differences between the color and the grayscale databases. In fact, when focusing the analysis on the best model found by EvoDeep across all datasets, remarkable differences arise between the test accuracy and the complexity of such models corresponding to color and grayscale datasets. As summarized in Table \ref{tab:dataset_results_evodeep}, accuracy results are similar in both cases. The accuracy achieved by the best EvoDeep model over CIFAR-10 in higher than that of its grayscale version (CIFAR-10-G). The case with the largest difference in terms of accuracy is VANGOGH, which has a 17\% gap. In terms of complexity, the best model of each grayscale database has fewer neurons than the corresponding colored dataset. In HORSEHUMAN-G, the model has approximately 14.69 times fewer neurons than that of HORSEHUMAN. For VANGOGH and CIFAR-10, this ratio gets close to 24 times and 5.75 times, respectively. 
\begin{table}[H]
    \centering
    \setlength{\tabcolsep}{0.3em}
    \caption{Results in terms of accuracy and complexity for the best model encountered by EvoDeep for both color and grayscale datasets.}
    \resizebox{\textwidth}{!}{
    \begin{tabular}{cccccccc}
    \toprule
    & HORSEHUMAN & HORSEHUMAN-G & VANGOGH & VANGOGH-G & CIFAR-10 & CIFAR-10-G & MNIST\\
    \midrule
    \textbf{Test accuracy (\%)}       & 89.45 & 89.94 & 98.87 & 81.58 & 60.55 & 56.79 & 98.69 \\
    \textbf{\# of neurons}        & 1,895 & 129 & 18,756 & 783 & 10,821 & 1,883 & 5,409 \\
    \bottomrule
    \end{tabular}}
    \label{tab:dataset_results_evodeep}
  \end{table}

We summarize now the main conclusions drawn from this discussion, leaving a further elaboration on the general lessons learned in regards to topology and hyper-parameter optimization for Section \ref{sec:lessons}:
\begin{itemize}[leftmargin=*]

\item EvoDeep has similar results to AutoKeras in the color databases HORSEHUMAN, VANGOGH and CIFAR-10. Nevertheless, the computation time that EvoDeep requires is much higher than the one taken AutoKeras during its search. If we consider the grayscale datasets, the difference between AutoKeras and EvoDeep increases. In particular, accuracy gaps over the VANGOGH dataset is particularly large: 81.58\% in VANGOGH-G and 98.87\% in VANGOGH. In terms of accuracy, EvoDeep performs better with the colored databases.

\item If we take a closer look at the results in terms of accuracy and model complexity, EvoDeep needs less computation time in the grayscale datasets. Furthermore, this statement also holds in terms of model complexity. Although there are some models that comprise more layers when dealing with grayscale datasets, the number of total trainable parameters for all the models is much lower. All these facts contribute to a better overall performance of EvoDeep with grayscale databases.
\end{itemize}

To summarize, although AutoKeras has better performance in terms of accuracy in all datasets, EvoDeep gives competitive results and requires less computation time for simple datasets (in our case, MNIST). For several datasets, EvoDeep produces models with fewer parameters, while AutoKeras yield very complex models that may be unsuitable for application scenarios with stringent memory restrictions. In other datasets, the simplicity of the layers supported by EvoDeep enforces a higher number of neurons (and parameters) than AutoKeras for the same performance level.

The empirical results reported in this first case of study underscore the potential of Evolutionary Algorithms for the topological and hyper-parameter optimization of CNN networks, suggesting several possible improvements to frameworks appearing in the literature in forthcoming years. First, frameworks should incorporate sophisticated layers at their core, so that they become eligible for the evolutionary algorithm in use and ultimately lead to a reduced overall complexity of the optimized models. The overly complex models encountered by EvoDeep in some of the considered image classification datasets is a clear evidence that, for the sake of fair comparisons, all frameworks should ensure that the search spaces explored by their optimization engines are comparable to each other as well. Another possible improvement is to formulate topological and hyper-parameter optimization as a multi-objective problem, embracing as conflicting objectives the accuracy of the model, and a measure of its complexity (layers, parameters, computation time, etc). We believe that such an output could allow the community to examine the behavior of new frameworks and solvers in regards to the performance and complexity of their produced solutions, making their output more flexible to implement the discovered models by taking into account both objectives. We will later revolve on these research directions on Sections \ref{sec:lessons} and \ref{sec:challenges}. 

\section{Case of Study II: Training Deep Learning Models with Bio-inspired Algorithms} \label{ssec:exp2}

This second case of study aims to shed light on the performance of bio-inspired optimization algorithms when applied to model training (\emph{trainable parameter optimization} as per our nomenclature in this survey). The ultimate goal is to check whether bio-inspired algorithms can be a competitive alternative to gradient-based methods when undertaking image classification tasks over well-known datasets, ensuring that Deep Learning architectures of realistic complexity are in use. To this end, we design an experimental setup to provide an informed response to the following research questions (RQ):
\begin{itemize}[leftmargin=*]
\item RQ1: Should bio-inspired solvers exploit the layered structure of Deep Learning models during the search?
\item RQ2: Do bio-inspired solvers perform competitively with respect to gradient-based solver for trainable parameter optimization, in terms of predictive accuracy and computational efficiency?
\end{itemize}

In the remainder of this second case of study, Subsection \ref{ssec:SHADEILS} introduces the evolutionary algorithm selected for the experiments, underscoring several changes made to its original definition to make it better suited to the optimization of trainable parameters. Subsection \ref{ssec:experimental_setup_II} provides further details on the experimental setup, including the Deep Learning models and datasets under consideration. Subsections \ref{ssec:RQ1} and \ref{ssec:RQ2} discuss in depth on the results obtained from the experiments, with a focus on RQ1 and RQ2, respectively.

\subsection{SHADE-ILS: A reference evolutionary algorithm for large-scale global optimization} \label{ssec:SHADEILS}

Given the large number of variables to be optimized, we select SHADE with Iterative Local Search (SHADE-ILS) as the evolutionary algorithm selected for our experiments. SHADE-ILS is a renowned large-scale global optimization algorithm that has won recent international competitions in the field \cite{molina2018shade}. In particular, SHADE-ILS resorts to the global exploration capability of an adaptive variant of the Differential Evolution algorithm (SHADE), which is helped by two local search methods -- namely, a limited-memory version of the Broyden–Fletcher–Goldfarb–Shanno (BFGS) algorithm, and multiple trajectory search -- that improve the candidate solutions encountered during the search. At every iteration, SHADE applies evolutionary operators on a population of individuals, followed by the application of one of the local search methods to the best individual of the population. The selection of which local search to apply is driven by the best expected relative improvement of each of the considered local search operator, which is given by the results produced by each local search alternative during recent generations. Moreover, SHADE-ILS incorporates a restart mechanism to avoid stagnation. These key algorithmic aspects and the excellent results scored in benchmarks and competitions make SHADE-ILS one of the baseline algorithms for large-scale global optimization problems, just like the one addressed in this second case of study.

Several changes have been done to the original SHADE-ILS algorithm to make it better suited to the optimization of the trainable parameters of a Deep Learning model. The objective function to be minimized over the search is the same than that used by gradient back-propagation approaches for the same model, namely, binary cross-entropy for binary classification and categorical cross-entropy for multi-class classification. Both are measured for all examples belonging to the training set. Secondly, we take into account the layer-wise structure of Deep Learning models by devising several scheduling strategies. Each strategy establishes the criteria, order and number of generations for which the optimization variables (trainable parameters) belonging to each layer are evolved via the global search operators and local search techniques of SHADE-ILS. The design of the above strategies is motivated by RQ1, where the focus is placed on whether the layered structure of Deep Learning models should be exploited anyhow by the bio-inspired solver. 
\begin{figure}[ht!]
	\centering
	\includegraphics[width=\columnwidth]{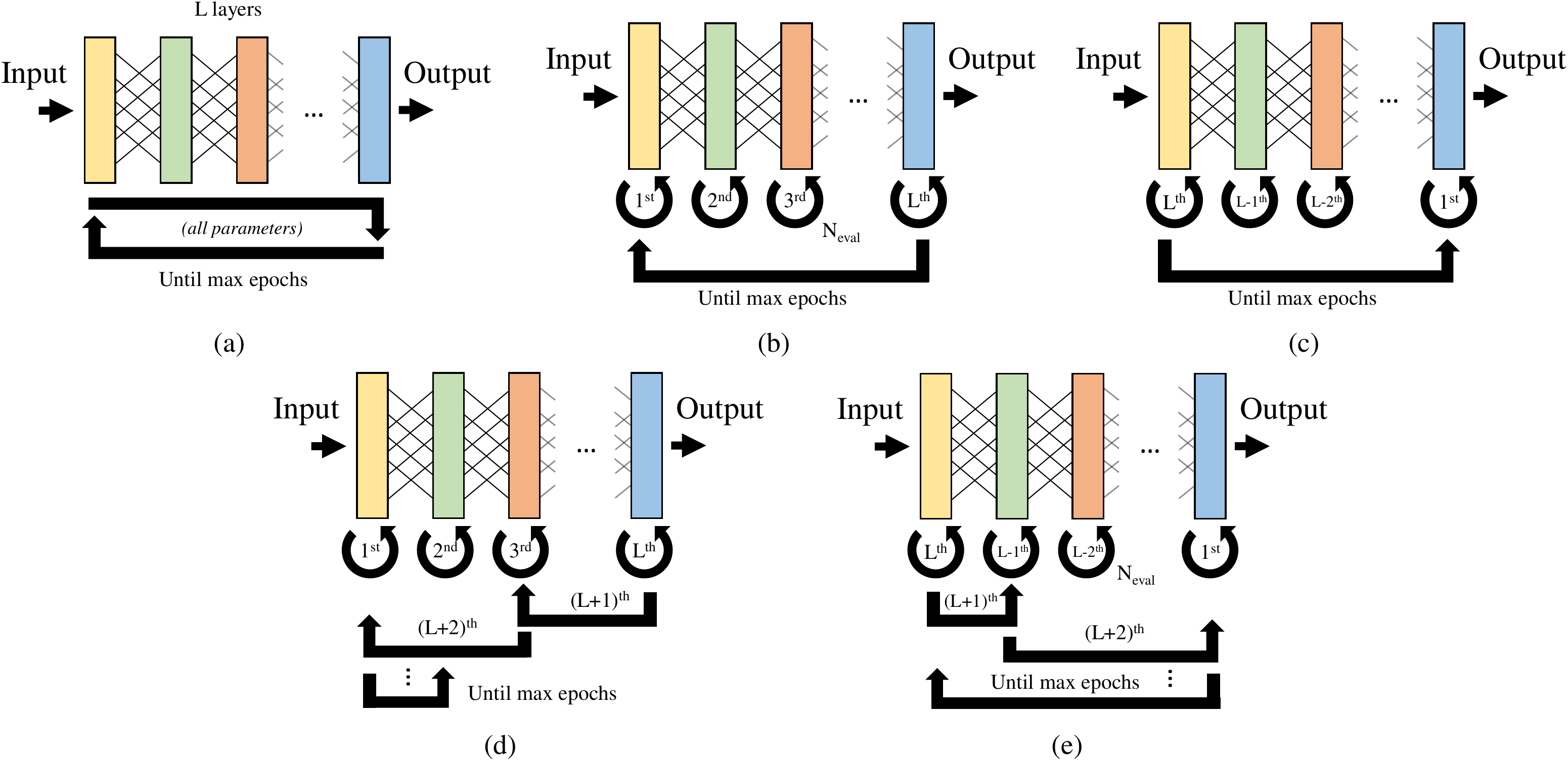}
	\caption{Diagram showing the different scheduling strategies proposed for optimizing Deep Learning models with the SHADE-ILS algorithm: (a) FULL-SHADE-ILS; (b) DOWN-SHADE-ILS; (c) UP-SHADE-ILS; (d) A-DOWN-SHADE-ILS; (e) A-UP-SHADE-ILS.}
	\label{fig:shade_schedules}
\end{figure}

Specifically, the strategies considered in the simulations discussed later are as follows (see Figure \ref{fig:shade_schedules} for a visual explanation):
\begin{itemize}[leftmargin=*]
\item FULL-SHADE-ILS: all trainable parameters are optimized jointly, without considering the layered structure of the model to be trained. This is actually the strategy followed by most contributions to the literature dealing with the application of bio-inspired optimization techniques for Deep Learning models. As we will later show, this strategy only works for network architectures of relatively limited size. 
\item DOWN-SHADE-ILS: trainable parameters are optimized starting from those belonging to the first layer of the network. Such parameters are evolved via SHADE-ILS for a certain number of generations, keeping the values of the remaining parameters fixed to their Glorot-based initialized values. The optimization schedule is repeated in order from the first to the last layer of the network for a maximum number of epochs. 
\item UP-SHADE-ILS: this schedule is similar to the previous DOWN-SHADE-ILS strategy, but departing from the last layer of the network, and proceeding upwards until the first layer of the network.
\item A-DOWN-SHADE-ILS: an automated variant in which after a first iteration of the DOWN-SHADE-ILS optimization strategy, a relative improvement ratio of the network predictive accuracy is computed for every layer optimization step. This ratio is used to select which layer to optimize in subsequent epochs, so that layers whose optimization yielded larger improvements in the last epoch are more likely to be selected for optimization.
\item A-UP-SHADE-ILS: this last strategy is similar to A-DOWN-SHADE-ILS, the difference being that the first layer-wise application of SHADE-ILS is done from the last to the first layer of the network (namely, under the DOWN-SHADE-ILS strategy). Once this initial stage is completed, SHADE-ILS proceeds analogously to the previous strategy, automatically selecting the layer with most potential margin of improvement.
\end{itemize}

When addressing RQ2, it is important to stress on the complexity of ensuring a fair comparison between gradient back-propagation solvers and bio-inspired optimization algorithms. The main reason is their essentially different search behavior. On one hand, gradient back-propagation approaches maintain a single solution to the problem, which is enhanced over epochs by exploit the mathematical relationships between the optimization variables (trainable parameters) through the tailored computation of their gradients. However, the search operator of gradient back-propagation techniques is simple, yet effective (gradients are personalized for every single variable) and efficient (gradient computations are highly parallelizable). By contrast, most bio-inspired algorithms rely on a population of individuals, which are evolved jointly by means of a series of search operators, so that the best individuals survive between generations. In summary, comparisons between both approaches should be fair not only in terms of accuracy performance, but also in terms of computational complexity. 

For ensuring fairness in these terms, a straightforward decision is to define how \emph{epoch} and \emph{generation} relate to each other. First of all, it is clear that both concepts indicate when a full update of the network's parameters is complete: in gradient back-propagation, an epoch implies the application of a number of gradient updates to the whole network parameters. The number of updates depends on the size of the training set and the chosen batch size. Given that an epoch is defined as the optimization of all parameters composing the model, in SHADE-ILS we will establish than an \emph{epoch} corresponds to the optimization of all layers of the network under any of the strategies described previously. If we assume $N_{eval}$ evaluations of the network per layer and $L$ layers, an epoch for SHADE-ILS will comprise $L\cdot N_{eval}$ total evaluations of the network per epoch. Each evaluation of the network involves predicting the entire $N_{train}$-sized training set and computing the loss function. As a result, in general the number of evaluated training instances per epoch differs between SHADE-ILS ($N_{train}\cdot N_{eval} \cdot L$) and gradient back-propagation ($N_{train}$). Nevertheless, we proceed forward with the experiments disregarding this issue, and analyze whether SHADE-ILS, even if endowed with more computational budget per epoch, can beat the accuracy of networks optimized via Adam, one of the most renowned gradient back-propagation solvers. 

\subsection{Experimental setup} \label{ssec:experimental_setup_II}

The above two questions are tackled by considering 6 datasets for image classification: Hand gesture recognition (HANDS, \cite{mantecon2016hand}), Blood Cells classification dataset (BCCD, \cite{BCCD}), MNIST \cite{lecun-mnisthandwrittendigit-2010}, Fashion MNIST (F-MNIST) \cite{xiao2017fashion}, GTSRB \cite{Stallkamp2012} and CIFAR-10 \cite{cifar10dataset}. Details of these datasets are given in Table \ref{tab:exp_config}. Given the amount of computational resources required to complete the experiments, we consider a subset of the examples for each dataset. For the same reason and except for the WBC dataset, images have been converted to grayscale to reduce the number of channels of the input image.
\begin{table}[ht!]
    \centering
    \caption{Datasets and models utilized for the second case of study. $\texttt{C2D}_{N,x\times y}$ denotes a convolutional layer with $N$ filters of $n$ rows and $m$ cols; $\texttt{D}_{N}$ represents a fully-connected (\emph{dense}) layer with $N$ output neurons; $\boxplus$ is a $2\times 2$ max pooling layer; $\bigodot$ is a $2\times 2$ average pooling layer; and $\texttt{Drop}_{p}$ is a dropout layer with rate $p$. Layers enclosed within $(\cdot)^L$ are concatenated $L$ times.}
    \resizebox{\textwidth}{!}{%
    \begin{tabular}{C{2cm}|c|C{1.2cm}|C{2.2cm}|C{2cm}|l}
    \toprule
         \textbf{Dataset} & \textbf{Shape} & \textbf{\# classes} & \textbf{\# Instances ($N_{train}$/$N_{test}$)} & \textbf{\# trainable parameters} & \multicolumn{1}{c}{\textbf{Network topology and structural hyper-parameters}} \\
        \midrule
        HANDS       & $30\times 40\times 1$ & 10 & 10,000 / 10,000    &   3,854 &  $\texttt{C2D}_{8,4\times4}-\boxplus-(\texttt{C2D}_{16,2\times2}-\boxplus)^2-\texttt{D}_{20}-\texttt{D}_{10}$ \\
        BCCD         & $30\times 40\times 3$ & 2 & 17,000 / 5,416  &   9,065  &  $\texttt{C2D}_{30,3\times3}-\boxplus-(\texttt{C2D}_{16,3\times3}-\boxplus)^2-\texttt{D}_{16}-\texttt{Drop}_{0.7}-\texttt{D}_{1}$\\
        MNIST       & $28\times 28\times 1$ & 10 & 10,000 / 5,000    &   19,063  &  $\texttt{C2D}_{28,3\times3}-\boxplus-\texttt{C2D}_{14,3\times3}-\boxplus-\texttt{C2D}_{7,2\times2}-\boxplus-\texttt{D}_{128}- \texttt{Drop}_{0.2} -\texttt{D}_{80}- \texttt{Drop}_{0.3} -\texttt{D}_{10}$\\
        F-MNIST     & $28\times 28\times 1$ & 10 & 10,000 / 5,000    &   36,188  &  $\texttt{C2D}_{64,4\times4}- \texttt{Drop}_{0.25} -\bigodot-\texttt{C2D}_{16,4\times4}- \texttt{Drop}_{0.25} -\bigodot- \texttt{Drop}_{0.15} -\texttt{D}_{70}-\texttt{D}_{10}$ \\
        GTSRB       & $32\times32\times 1$  & 43 & 20,000 / 10,000   &   83,999  &  $\texttt{C2D}_{6,3\times3}-\bigodot-\texttt{C2D}_{16,3\times3}-\bigodot-\texttt{D}_{120}-\texttt{D}_{84}-\texttt{D}_{43}$ \\
        CIFAR-10-G  & $32\times 32\times 1$ & 10 & 10,000 / 5,000    &   1,658,570 &  $\texttt{C2D}_{32,3\times3}-\texttt{Drop}_{0.1}-\texttt{C2D}_{64,5\times5}-\texttt{Drop}_{0.2}-\texttt{D}_{128}-\texttt{Drop}_{0.3}-\texttt{D}_{10}$\\
        \bottomrule
    \end{tabular}
    }
    \label{tab:exp_config}
\end{table}

For each of these datasets a fixed Deep Learning architecture is considered, featuring a realistic level of complexity (given by its number of trainable parameters), and rendering a good prediction performance when trained via gradient back-propagation. The layer type and structural hyper-parameters of every layer compounding such models are also specified in the table. Thus, the aim of this section is not to evolve very precise, state-of-the-art networks, but to assess the limitations faced by Evolutionary Algorithms when used for training these deep neural networks. Table \ref{tab:training_hyp} summarizes the training hyper-parameters utilized for all image classification tasks under study.
\begin{table}[ht!]
    \centering
    \setlength{\tabcolsep}{0.4em}
    \caption{Training hyper-parameters used in our experiments.}
    \resizebox{0.7\textwidth}{!}{%
    \begin{tabular}{ccccccccc}
    \toprule
        & \multicolumn{2}{c}{Adam} & & \multicolumn{2}{c}{SHADE-ILS} & & & \\
        \cmidrule{2-3} \cmidrule{5-6}
         \textbf{Dataset} & \textbf{Batch size} & \textbf{Learning rate} & & \textbf{Population size} & $\mathbf{N_{eval}}$ & & \textbf{Initializer} & \textbf{Epochs}\\
         \midrule
        HANDS & 128 & 0.01 & & 10 & 200 & & Glorot & 20 \\
        BCCD & 64 & 0.02 & & 10 & 200 & & Glorot & 20 \\
        MNIST & 512 & 0.01 & & 10 & 200 & & Glorot & 20 \\
        F-MNIST & 512 & 0.01 & & 10 & 200 & & Glorot & 20 \\
        GTSRB & 64 & 0.02 & & 10 & 200 & & Glorot & 22 \\
        CIFAR-10-G & 256 & 0.01 & & 10 & 200 & & Glorot & 30 \\
        \bottomrule
    \end{tabular}
    }
    \label{tab:training_hyp}
\end{table}

\subsection{Addressing RQ1: On the importance of the layered neural structure in the design of SHADE-ILS} \label{ssec:RQ1}

Once the experimental setup has been described, we start our discussion by addressing RQ1, namely, a quantitative analysis of the impact and viability of exploiting the layered structure of Deep Learning models by SHADE-ILS, similarly to what gradient-based solvers do when back-propagating the gradients. To this end, we evaluate the aforementioned scheduling strategies devised for SHADE-ILS over the datasets and Deep Learning models under consideration, and compare its performance to that of a naive application of SHADE-ILS that does not take into account any structure of the problem.

Table \ref{tab:dataset_results} summarizes the results obtained in regards to RQ1. Specifically, the average loss and accuracy measured over train and test subsets are reported for every dataset and SHADE-ILS schedule. Values are averaged over 5 independent runs of every (dataset,schedule) combination. In addition, results corresponding to the Adam gradient-based solver are also included as a reference. The best results among those yielded by SHADE-ILS are highlighted in bold.
\begin{table}[ht!]
    \centering
    \setlength{\tabcolsep}{0.3em} 
    \caption{Average accuracy/loss (over 5 independent runs) corresponding to different schedules of the SHADE-ILS algorithm over the datasets under consideration. Results corresponding to the Adam gradient-based solver are also included as a reference.}
    \resizebox{\textwidth}{!}{%
    \begin{tabular}{c|ccccccccccccccccc}
    \toprule
    & \multicolumn{2}{c|}{Adam} & & \multicolumn{2}{c|}{\makecell{FULL\\SHADE-ILS}}& & \multicolumn{2}{c|}{\makecell{DOWN\\SHADE-ILS}}& & \multicolumn{2}{c|}{\makecell{UP\\SHADE-ILS}}& & \multicolumn{2}{c|}{\makecell{A-DOWN\\SHADE-ILS}}& & \multicolumn{2}{c}{\makecell{A-UP\\SHADE-ILS}}\\
    \cmidrule{2-3} \cmidrule{5-6} \cmidrule{8-9} \cmidrule{11-12}\cmidrule{14-15} \cmidrule{17-18} 
    & \multicolumn{1}{c}{\textbf{Train}} &\multicolumn{1}{c}{\textbf{Test}} & & \multicolumn{1}{c}{\textbf{Train}} &\multicolumn{1}{c}{\textbf{Test}} & & \multicolumn{1}{c}{\textbf{Train}} &\multicolumn{1}{c}{\textbf{Test}} & &\multicolumn{1}{c}{\textbf{Train}}  &\multicolumn{1}{c}{\textbf{Test}} & &\multicolumn{1}{c}{\textbf{Train}} &\multicolumn{1}{c}{\textbf{Test}} & &\multicolumn{1}{c}{\textbf{Train}} & \multicolumn{1}{c}{\textbf{Test}} \\
    \cmidrule{2-3} \cmidrule{5-6} \cmidrule{8-9} \cmidrule{11-12}\cmidrule{14-15} \cmidrule{17-18} 
    & \makecell{\textbf{Acc}\\\textbf{Loss}} & \makecell{\textbf{Acc}\\\textbf{Loss}} & & \makecell{\textbf{Acc}\\\textbf{Loss}} & \makecell{\textbf{Acc}\\\textbf{Loss}} & & \makecell{\textbf{Acc}\\\textbf{Loss}} & \makecell{\textbf{Acc}\\\textbf{Loss}} & & \makecell{\textbf{Acc}\\\textbf{Loss}} & \makecell{\textbf{Acc}\\\textbf{Loss}} & & \makecell{\textbf{Acc}\\\textbf{Loss}} & \makecell{\textbf{Acc}\\\textbf{Loss}} & & \makecell{\textbf{Acc}\\\textbf{Loss}} & \makecell{\textbf{Acc}\\\textbf{Loss}}\\
    \midrule
    HANDS & \makecell{0.9983\\0.0082} & \makecell{0.9932\\0.0296} & &
                    \makecell{\textbf{0.9708}\\\textbf{0.1082}}& \makecell{\textbf{0.9553}\\\textbf{0.1589}} & &
                    \makecell{0.7038\\2.6608} & \makecell{0.6910\\2.6894} & &
                    \makecell{0.7150\\0.8194} & \makecell{0.7072\\0.8705} & &
                    \makecell{0.6492\\3.5419} & \makecell{0.6470\\3.5750} & &
                    \makecell{0.5253\\1.3179} & \makecell{0.5189\\1.3533}\\
    \midrule                
    BCCD  & \makecell{0.9611\\0.1314} & \makecell{0.9815\\0.0566} & &
                    \makecell{\textbf{0.8558}\\\textbf{0.3321}} & \makecell{\textbf{0.8489}\\\textbf{0.3399}} & &
                    \makecell{0.7499\\0.5306} & \makecell{0.7400\\0.5377} & &
                    \makecell{0.7527\\0.5227} & \makecell{0.7439\\0.5322} & &
                    \makecell{0.7852\\0.4861} & \makecell{0.7768\\0.4962} & &
                    \makecell{0.8212\\0.4149} & \makecell{0.8087\\0.4301}\\
    \midrule                    
    MNIST & \makecell{0.9480\\0.1672} & \makecell{0.9534\\0.1534} & &
                    \makecell{0.9524\\0.1590} & \makecell{0.9342\\0.2326} & &
                    \makecell{0.9747\\0.0851} & \makecell{0.9504\\0.1685} & &
                    \makecell{0.9748\\0.0856} & \makecell{0.9490\\0.1798} & &
                    \makecell{0.9719\\0.0938} & \makecell{0.9452\\0.1828} & &
                    \makecell{\textbf{0.9772}\\\textbf{0.0764}} & \makecell{\textbf{0.9508}\\\textbf{0.1627}}\\
    \midrule
    F-MNIST & \makecell{0.9636\\0.1190} & \makecell{0.9672\\0.1100} & &
                            \makecell{0.9265\\0.2539} & \makecell{0.9196\\0.2940} &  &
                            \makecell{0.9489\\0.1773} & \makecell{0.9384\\0.2157} &  &
                            \makecell{0.9422\\0.2077} & \makecell{0.9311\\0.2407} &  &
                            \makecell{\textbf{0.9561}\\\textbf{0.1576}} & \makecell{\textbf{0.9447}\\\textbf{0.1966}} & &
                            \makecell{0.9435\\0.2012} & \makecell{0.9336\\0.2317} \\  
    \midrule
    GTSRB  & \makecell{0.7437\\0.6957} & \makecell{0.7046\\0.7807} & &
                    \makecell{0.2329\\3.0994} & \makecell{0.2355\\3.1196} & &
                    \makecell{0.3636\\2.4532} & \makecell{0.3473\\2.4931} & &
                    \makecell{0.3596\\2.4304} & \makecell{0.3504\\2.4672} & &
                    \makecell{\textbf{0.3956}\\\textbf{2.2916}} & \makecell{\textbf{0.3868}\\\textbf{2.3323}} & &
                    \makecell{0.3204\\2.6405} & \makecell{0.3133\\2.6622} \\
    \midrule
    CIFAR-10-G  & \makecell{0.8347\\0.4519} & \makecell{0.6542\\1.2418} & &
                        \makecell{0.2628\\2.0952} & \makecell{0.2602\\2.1087} & &
                        \makecell{0.3696\\1.7878} & \makecell{0.3612\\1.8023} & &
                        \makecell{0.3708\\1.7737} & \makecell{0.3642\\1.7867} & &
                        \makecell{\textbf{0.3806}\\\textbf{1.7483}} & \makecell{\textbf{0.3790}\\\textbf{1.7619}} & &
                        \makecell{0.3765\\1.7612} & \makecell{0.3681\\1.7787}\\    
    \bottomrule
    \end{tabular}}
    \label{tab:dataset_results}
\end{table}

Several interesting observations can be made after inspecting the above table. To begin with, we see that for relatively small-sized Deep Learning models (i.e. those for the HANDS and BCCD datasets), FULL-SHADE-ILS suffices for obtaining good scores, even superior than those rendered by its scheduled SHADE-ILS counterparts. This goes in line with our examination of the related literature, in which many contributions deal with model training using naive bio-inspired solvers, without taking into account the structure of the network. The above results confirm that when the number of network parameters is low, a powerful optimization algorithm can effectively (albeit not efficiently) find optimal values for the task at hand. 

However, when increasing the complexity of the network, the trend changes, and the exploitation of the layered structure of the Deep Learning model becomes essential to maintain a good performance. This is specially remarkable in the GTSRB and CIFAR-10-G datasets, in which the accuracy values of FULL-SHADE-ILS degrade severely with respect to those attained by A-DOWN-SHADE-ILS. For networks of moderate size (MNIST, F-MNIST), accuracy differences between the scheduled and non-scheduled versions of SHADE-ILS become neglibigle. Nevertheless, in terms of loss metric values the difference results to be larger, showing evidence that SHADE-ILS performs better in terms of optimized loss when endowed with the automated schedule mechanism.

When comparing the accuracy and loss values measured over train and test, a quick glimpse at the table confirms that in general, the trained models do not overfit excessively. We pause briefly at the case of the MNIST dataset, arguably one of the most utilized databases in this field. If we compare the average loss values achieved by the Adam optimizer ($0.1672$ over train, $0.1534$ over test) to those of A-UP-SHADE-ILS ($0.0764$ over train, $0.1627$ in test), one can state that A-UP-SHADE-ILS achieves lower loss values than Adam, thereby concluding that this scheduled SHADE-ILS variant is a better optimization algorithm for model training than Adam. However, we must bear in mind that the final goal of predictive modeling is to provide models that \emph{generalize nicely}, namely, models that perform as expected when predicting unseen data instances. When placing our attention in the losses measured over the test set, networks evolved by A-UP-SHADE-ILS for the MNIST dataset seems to overfit, thereby providing lower accuracy scores than expected. A similar conclusion can be drawn in other datasets (e.g. F-MNIST), yet at a lower extent than MNIST. This leads to an interesting insight on the influence of overfitting that we will elaborate in depth in Section \ref{sec:lessons}.

We conclude the discussion on this first set of results by emphasizing on the low scores obtained by all SHADE-ILS variants when the complexity of the network is very high (models corresponding to the GTSRB and CIFAR-10-G datasets). In these cases the large gaps to the accuracy and loss scores of the network optimized by the Adam solver indicate without doubts that this meta-heuristic algorithm is of no practical use for this level of complexity. Given that SHADE-ILS is specially tailored to deal with high-dimensional optimization problems, it is fair to conclude that Evolutionary Computation and Swarm Intelligence methods are still far from being a realistic replacement for gradient-based solver. Instead, these empirical findings should drive the interest of the community towards hybridizing bio-inspired algorithms with gradient-based information and/or solvers. We will later revolve on this postulated research line.

\subsection{Addressing RQ2: comparing bio-inspired optimization algorithms to gradient-based solvers} \label{ssec:RQ2}

Experiments discussed in the previous subsection have concentrated on the performance comparison between different layer-wise optimization schedules of the SHADE-ILS solver. In our analysis of the results shown in Table \ref{tab:dataset_results} we also highlighted the large gap between the gradient-based Adam solver and the best performing SHADE-ILS schedule, specially for Deep Learning models of moderate-to-high levels of complexity. Although this remark provides a partial answer to RQ2, in this second part of the study we delve into the convergence of the training process when undertaken via SHADE-ILS and Adam, aiming to discern the reasons for these identified performance gaps.
\begin{figure}[ht!]
	\centering
	\includegraphics[width=\textwidth]{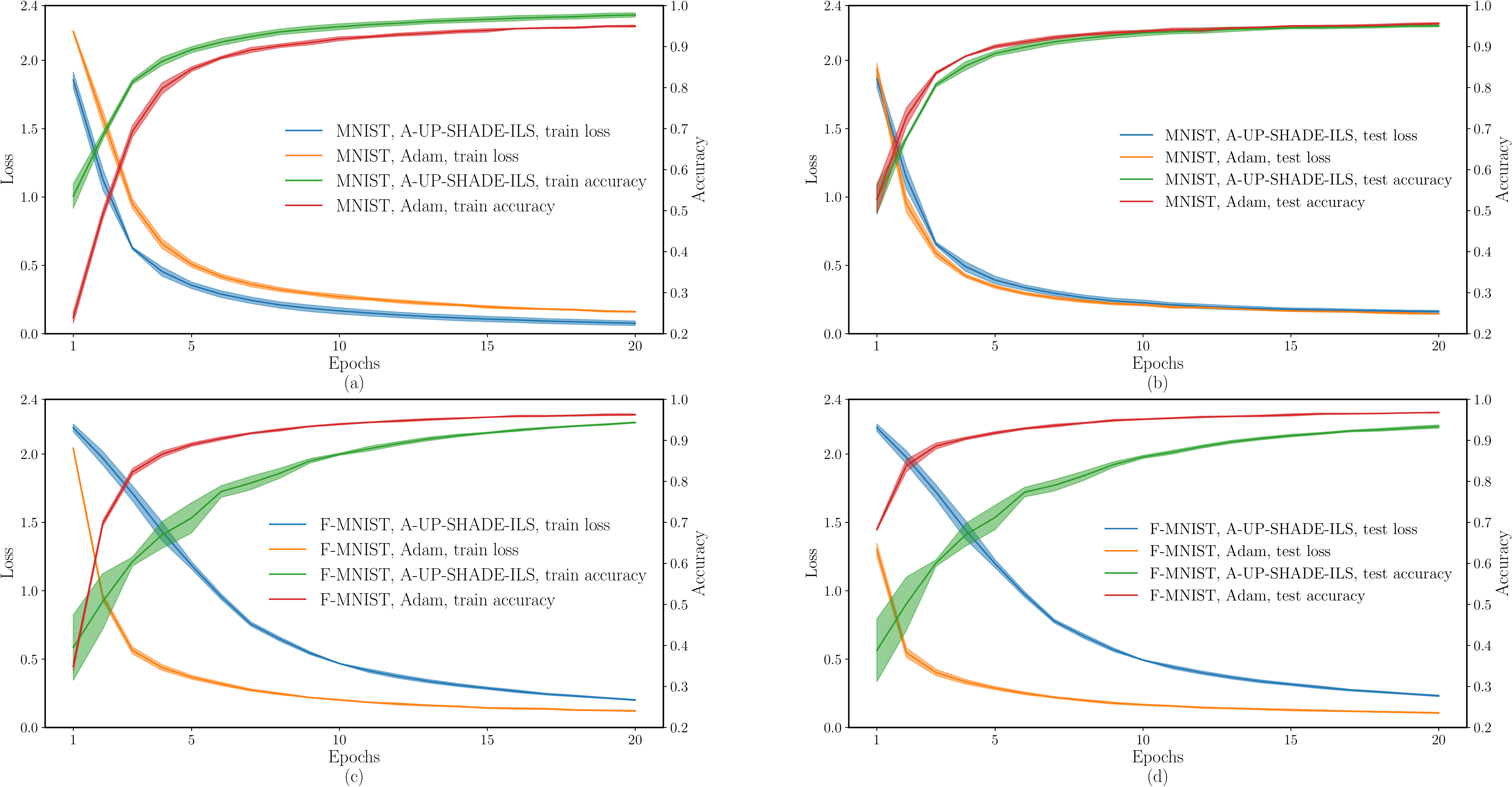}
	\caption{Accuracy and loss Convergence plots of Adam and A-UP-SHADE-ILS corresponding to (a) MNIST, measured over train set; (b) MNIST, measured over test set; (a) F-MNIST, measured over train set; (a) F-MNIST, measured over test set.}
	\label{fig:mnist_fmnist}
\end{figure}

This being said, we focus on the results corresponding to MNIST and F-MNIST, which are illustrative of the conclusions that can be drawn from the overall set of performed experiments. The dual plots in Figures \ref{fig:mnist_fmnist}.a to \ref{fig:mnist_fmnist}.d depict the loss/accuracy convergence plots measured over train and test subsets corresponding to Adam and the scheduled A-UP-SHADE-ILS variant of the SHADE-ILS algorithm. Plotted lines depict the average loss/accuracy over epochs computed over 5 experiments, whereas the shaded overlay areas denote their standard deviation. Net loss values are indicated in the left axis of every plot, whereas accuracy score values are indicated in the right axis.

Our discussion departs from Figures \ref{fig:mnist_fmnist}.a and \ref{fig:mnist_fmnist}.b, corresponding to the convergence plots over the MNIST dataset over train and test subsets, respectively. It is straightforward to observe that when measured over the train dataset (Figure \ref{fig:mnist_fmnist}.a, both loss and accuracy scores of the A-UP-SHADE-ILS approach are better than those rendered by Adam over all epochs. This fact is conclusive in regards to the comparable or superior performance of SHADE-ILS when optimizing the trainable parameters of relatively small-sized networks. However, when shifting the focus on the test set (which reflects the generalization capability the evolved Deep Learning models), the curves plotted in Figure \ref{fig:mnist_fmnist} reveal that A-UP-SHADE-ILS yields trainable parameter values that are slightly overfitted, thus generalizing worse than the model optimized via the Adam solver. This observation suggests that the apparently worse performance of Adam as an optimization algorithm is actually an advantage (a sort of \emph{implicitly regularization} mechanism) that yields models of better generalization properties. 

When increasing the complexity of the network to be evolved, Figures \ref{fig:mnist_fmnist}.c and \ref{fig:mnist_fmnist}.d illustrate a rather different behavior of the convergence plots. At this point we recall that the model selected to deal with the F-MNIST image classification problem has 36,188 trainable parameters, almost twice the complexity of the model designed for the MNIST dataset (19,063 trainable parameters). The convergence curves in these plots show that although the accuracy and loss values of the networks evolved with both solvers get close to each other after all epochs are completed,  A-UP-SHADE-ILS perform steadily worse than Adam over all intermediate epochs. In light of the results for the rest of datasets with models of larger complexity (Table \ref{tab:dataset_results}), the case of the F-MNIST dataset must be understood as an inflection point, beyond which SHADE-ILS fails to perform competitively with respect to gradient-based methods. Furthermore, this worse performance occurs even if SHADE-ILS is allowed to execute more network evaluations per epoch. 

These experiments and our conclusions drawn therefrom reinforce even further our belief that for the time being, current bio-inspired optimization algorithms do not constitute a feasible replacement for gradient-based solvers. In the next section we collect and summarize the lessons learned through our literature analysis and experiments, and prescribe several good practices and recommendations that should be followed to achieve significant advances in evolutionary Deep Learning.

\section{Learned Lessons, Good Practices and Recommendations} \label{sec:lessons}

As it follows from our experiments and the performed literature analysis, lights and shadows still remain in the application of Evolutionary Computation and Swarm Intelligence algorithms to the diverse optimization problems arising from Deep Learning. 

We now summarize the lessons learned throughout this study, classifying them depending on the optimization domain in which each lessons is most recurrently noted:
\begin{itemize}[leftmargin=*]
\item General lessons and recommended practices that hold for all Deep Learning optimization problems (Subsection \ref{ssec:general_lessons}).
\item Lessons and recommendations suited for studies related to topology optimization (Subsection \ref{ssec:topology_lessons}).
\item Lessons and good practices related to structural and training hyper-parameter optimization (Subsection \ref{ssec:hyper_lessons}).
\item Lessons and recommendations for the optimization of trainable parameters (Subsection \ref{ssec:trainable_lessons}).
\end{itemize}

\subsection{General lessons} \label{ssec:general_lessons}

The first lesson we summarize at this point is in close accordance with the general need for more methodological principles in meta-heuristic research across all the application scenarios where these solvers are applied nowadays. Unfortunately, the optimization problems tackled in this study are not an exception to this claim. The guidelines and procedures to be followed to reach solid and conclusive studies in the use of meta-heuristics are known to the community, specially in regards to the identification of the novel aspects of newly emerging algorithms and the proper design of comparison benchmarks. In this latter research stage, the assessment of the statistical significance is a must when dealing with problems related to Deep Learning topology and/or hyper-parameter optimization. Since several sources of uncertainty may coexist in the same problem statement (e.g. the operators of the search algorithm, the initialization of weights, and the stochasticity of the gradient-based training algorithm), reporting on the statistical significance via hypothesis testing should be considered a necessary step. Despite not an exclusive recommendation of the research area under study, leaving all code and results available in public software repositories for the research community is of utmost necessity, due to the dilated computation times usually required to run experiments with Deep Learning models.

Another general lessons that stems from our study is that the goal in Evolutionary Deep Learning is to yield models of improved generalization properties, namely, to find models that perform \emph{better} when fed with unseen data. To an extent, in our literature review we noted that many contributions in the recent past dismisses this target goal and conclude that the solver at hand performs better since it achieves a more optimal objective value than gradient-based methods. Statements alike should be avoided in prospective studies, as there is no practical value in a model that generalizes worse \emph{in the wild}, e.g. when predicting new data instances.

Another missing point in past contributions is a quantitative evaluation of the complexity of the solver, not only a mere indication of the quality of its produced output. We have shown in the cases of study that important complexity gaps exist between the solvers under study. Neglecting to inspect this important aspect of optimization solvers yields biased conclusions about the practical value of new proposals. Convergence plots like the ones depicted in the case of study II can provide a hint about the relationship between performance (accuracy) and complexity (number of epochs) of the solvers under comparison. Along these plots, a clear definition of an \emph{epoch} should be provided, so as to establish a reference and ensure fair complexity comparisons.

Furthermore, other optimization objectives beyond predictive performance should also be considered in the future, such as the complexity of the optimized model in topology and/or structural hyper-parameter optimization. There are many reasons for considering such objectives in addition to predictive performance, ranging from an easier deployment of the optimized models in constrained computation hardware (e.g. cell phones) or a potentially more interpretable network structure \cite{arrieta2020explainable}. However, most studies seem to focus just on the predictive performance of the optimized model, leaving unanswered relevant matters for model deployment such as the tradeoff between model's performance and complexity.

Finally, we advocate for a global consensus on the dimensions of realistic benchmark dataset and models. A Deep Learning model is not just a layered structure of perceptrons, but rather a hierarchical composite of neural layers of different nature. It is their capability to extract increasingly specialized features from large-dimensional data what should bestow the label \emph{Deep Learning}. Unfortunately, we have observed many works where \emph{Deep Learning} refers to a neural network with very few trainable parameters and fed with already handcrafted features. The same can be said about the datasets used for validation, which in many studies lack the complexity that could argue the adoption of Deep Learning models. The community should agree on the minimum characteristics (task difficulty, diversity of datasets, model complexity) of experimental setups devised for reaching meaningful results and valuable conclusions.

\subsection{Topology optimization} \label{ssec:topology_lessons}

We now focus on the lessons learned from the literature that has so far tackled the optimization of the topology of Deep Learning models. First of all, in the first case of study we highlighted that the EvoDeep framework featured a diversity of layer types lower than that of AutoKeras, which restricts the search domain of the evolutionary algorithm that runs at its core. This fact can imprint a subtle yet impacting bias in prospective comparisons between new topology optimization frameworks. Such comparisons should ensure that the counterparts in the benchmark should utilize the same number and diversity of layer types when optimizing Deep Learning models for a given task. Otherwise, there is no certainty whether gaps found among the compared frameworks are due to the differences between their optimization algorithms or to the fact that some of them are in unfair disadvantage. Similar recommendations can be issued in regards to the formulation of the objective to be optimized, which should be set the same among frameworks.
In what refers to the optimization algorithm, we have shown that bio-inspired solvers (in particular, evolutionary algorithms) are still far from the performance offered by other ad-hoc methods. Actually, the default optimization approach for the AutoKeras framework is a parameter-wise greedy strategy, and no Evolutionary Computation nor Swarm Intelligence methods are considered whatsoever. This observation implicitly suggests that bio-inspired solvers are still far from performing competitively, as our results in the first case of study have clearly shown. However, it is known that in other application domains, greedy search methods usually fall into local optima unless proper algorithmic countermeasures are included along the search. This opens up an opportunity to hybridize such greedy methods with global search heuristics or, alternatively, to design ad-hoc search algorithms that incorporate any of the ingredients featured by such greedy methods.

Another aspect in which much research effort has been invested is in the design of variable-size encoding strategies for the representation of evolved network architectures. This has been a subject of intense research for years since the advent of the first neuro-evolution frameworks, in particular the adaptation of compositional pattern producing networks made by Hyper-NEAT to represent and evolve neural networks. Despite the numerous works on neural representation strategies published thereafter to date, in many cases the selected encoding approach does not account for the validity of the sequence of neural layers it represents. To this end, EvoDeep resorts to finite state machines to model all possible transitions between layers, followed by a two-part encoding to represent global (training) hyper-parameters and layers' types and structural hyper-parameters, respectively. It is our belief that a key aspect for an efficient heuristic search is to tightly couple the solution encoding strategy to the design of the search operators. 

Finally, we dedicate some words of reflection about the level of granularity at which topology optimization should be performed. While in some recent works Deep Learning models are optimized at the level of the computation graph, namely, without assuming any particular type of neural layer. This extreme is interesting should the goal be to trascend conventional layer types and seek more diverse neural structures. At the other end of the scale, other contributions have capitalized on the mixture of pretrained modules, aiming to enhance the structure of the Deep Learning model while leveraging, at the same time, high-level knowledge acquired in other related tasks. To the best of our knowledge, there is no clear consensus on whether high-level or low-level topological optimization strategies are more promising for Deep Learning topology optimization.  

\subsection{Structural/training hyper-parameter optimization} \label{ssec:hyper_lessons}

When it comes to the optimization of the hyper-parameters of the Deep Learning model, a first recommendation to elicit is to clearly specify the hyper-parameters to be optimized, as well as their search ranges. Echoing the aforementioned need for fairness in comparison benchmarks, and following our conclusions drawn from the Case of Study I (Section \ref{ssec:exp1}), it is crucial to guarantee that the compared solvers explore search spaces of equal complexity so as to remove any bias due to differences in this matter. A good practice for this purpose is to include a table listing each parameter with its corresponding search range, so that fairness can be guaranteed and reproducibility eased for the interested audience. 

Another recommendation in hyper-parameter optimization is to estimate the impact of different hyper-parameter values in terms of the predictive accuracy and overall complexity of the optimized model. By reporting on the correspondence between different hyper-parameters values and the overall performance and complexity of the model, the community can discern potentially good search ranges for such hyper-parameters, thereby reducing the time needed to perform new hyper-parametric optimization tasks.

Finally, our taxonomy of the existing literature shown in Figure \ref{fig:treeCat} revealed that the number of works simultaneously tackling structural and hyper-parameter optimization surpassed those dealing only with structural or training hyper-parameter optimization in isolation. This fact calls into question whether the results obtained so far for the latter cases are conclusive. The selected topology for the Deep Learning model affects directly the search space of a structural hyper-parameter optimization problem defined over it, so it remains uncertain whether the conclusions drawn for the particular network topology under choice can be extrapolated to any other network topology. This is why we suggest increasing the number of experiments with different network topologies and datasets when performing hyper-parameter optimization. Otherwise, the claims delivered by prospective studies can be in doubt due to the lack of enough empirical evidence.

\subsection{Trainable parameter optimization} \label{ssec:trainable_lessons}

To end with, the optimization of the trainable parameters of Deep Learning models is arguably the one grasping most interest from the community in recent times. Several learned lessons and recommendations can be issued in this regard.

First of all, the research community should come to an agreement and understand that currently, we are far from fully replacing gradient-based solvers with bio-inspired optimization algorithms. The results discussed in the Case of Study II (Section \ref{ssec:exp2}) buttress this statement with solid findings: one of the most renowned and competitive algorithms for large-scale global optimization (SHADE-ILS) has not been able to perform better than the gradient-based Adam solver. Besides, when increasing the complexity of the Deep Learning model, the performance of SHADE-ILS degrades severely even if granted more computational budget (number of loss evaluations) than its gradient-based counterpart. In summary, for the time being bio-inspired optimization algorithms cannot rival the computational efficiency and the quality of solutions produced by gradient-based methods.

The main reason for this conclusion can arguably be found in the structure of the Deep Learning model, which establishes relationships among the trainable parameters of consecutive layers that \emph{should} be exploited by the optimization algorithm. This is actually what gradient back-propagation realizes in a clever yet computationally efficient fashion, even though creating other known issues (e.g. gradient vanishing). In our experiments we noticed that when adapting the search behavior of SHADE-ILS to the layered structure of the network, simple layer-wise schedules of this algorithm yields remarkable performance boosts, specially for networks of relatively small size. This suggests that more sophisticated hybridization strategies should be further investigated to embed problem-specific knowledge (namely, the structure of Deep Learning models or gradient information) within the search procedure of bio-inspired solvers.

Notwithstanding this noted performance gap, gradient-based solvers restrict the spectrum of loss functions to those for which derivatives can be computed. However, bio-inspired algorithms do not impose any requirement on the objective function to be optimized, nor do they require it to be differentiable. This fact could tilt the scale towards the use of bio-inspired algorithms in singular learning tasks that require a tailored definition of the objective function, as in cases with severe class imbalance or multilabel classification. 

When it comes to computational efficiency, the higher complexity of population-based meta-heuristics when compared to that of gradient-based solvers should stimulate more parallel and distributed implementations of Evolutionary Computation and Swarm Intelligence methods. There are modern programming languages and frameworks that can be utilized to accelerate bio-inspired search algorithms, even if still lagging behind the typical runtimes of gradient-based techniques. Recent works aim indeed at this direction, reviewing implementations available so far and prescribing recommendations and guidelines for the implementation of meta-heuristics in GPU \cite{essaid2019gpu,tan2015survey} and asynchronous distributed computing architectures \cite{schryen2019parallel}. Experiments with large datasets and realistic Deep Learning models should capitalize on already available software packages that ease the seamless deployment of meta-heuristics in massively parallel computing hardware, such as jMetalPy \cite{benitez2019jmetalpy} (Apache Spark and Dask) and \textit{libCudaOptimize} \cite{nashed2012libcudaoptimize} (CUDA for GPU). Interestingly, Tensorflow (the computation engine that underlies well-known software libraries for Deep Learning models) also provides a naive implementation of Differential Evolution as one of its functionalities \cite{abadi2016tensorflow}. Parallel, federated or distributed computation frameworks for Deep Learning models are also spreading fast \cite{zhu2020real, desell2017large, balaprakash2018deephyper}. Definitely new studies should leverage the availability of these tools to undertake experiments at realistic complexity scales. 

We end with our learned lessons on trainable parameter optimization by emphasizing several good methodological practices that should be followed in prospective studies. First, the accuracy achieved by the optimized models over the test set should be informed jointly with the usual objective function statistics reported in experiments with bio-inspired meta-heuristics. This is particularly relevant in trainable parameter optimization to assess whether performance gaps identified between solvers do not come along with a penalty in the generalization of the evolved model. Furthermore, conventional gradient-based solvers should be always included in the benchmark, even if their lower complexity makes the comparison unfair in such terms. Finally, we recommend the use of convergence plots such as the ones depicted in Figures \ref{fig:mnist_fmnist}.a to \ref{fig:mnist_fmnist}.d as a visual tool to examine the relative differences between algorithms over epochs. This information can be very valuable for the deployment of the solver(s) in real hardware, as well as for the detection of overfitting issues like the one identified in our experiments.

\section{Challenges and Research Directions} \label{sec:challenges}

Our exhaustive literature review and performed experiments have unveiled several promising facts and unsolved caveats of Evolutionary Computation and Swarm Intelligence algorithms when used for addressing optimization problems related to Deep Learning. Nonetheless, many proposals have been contributed to date for assorted learning tasks, not only supervised and unsupervised learning, but also other paradigms relying on Deep Learning models (e.g. deep reinforcement learning). Despite this noted activity, several research niches still remain uncharted or insufficiently addressed in this fusion of technologies.
\begin{figure}[ht!]
	\centering
	\includegraphics[width=\textwidth]{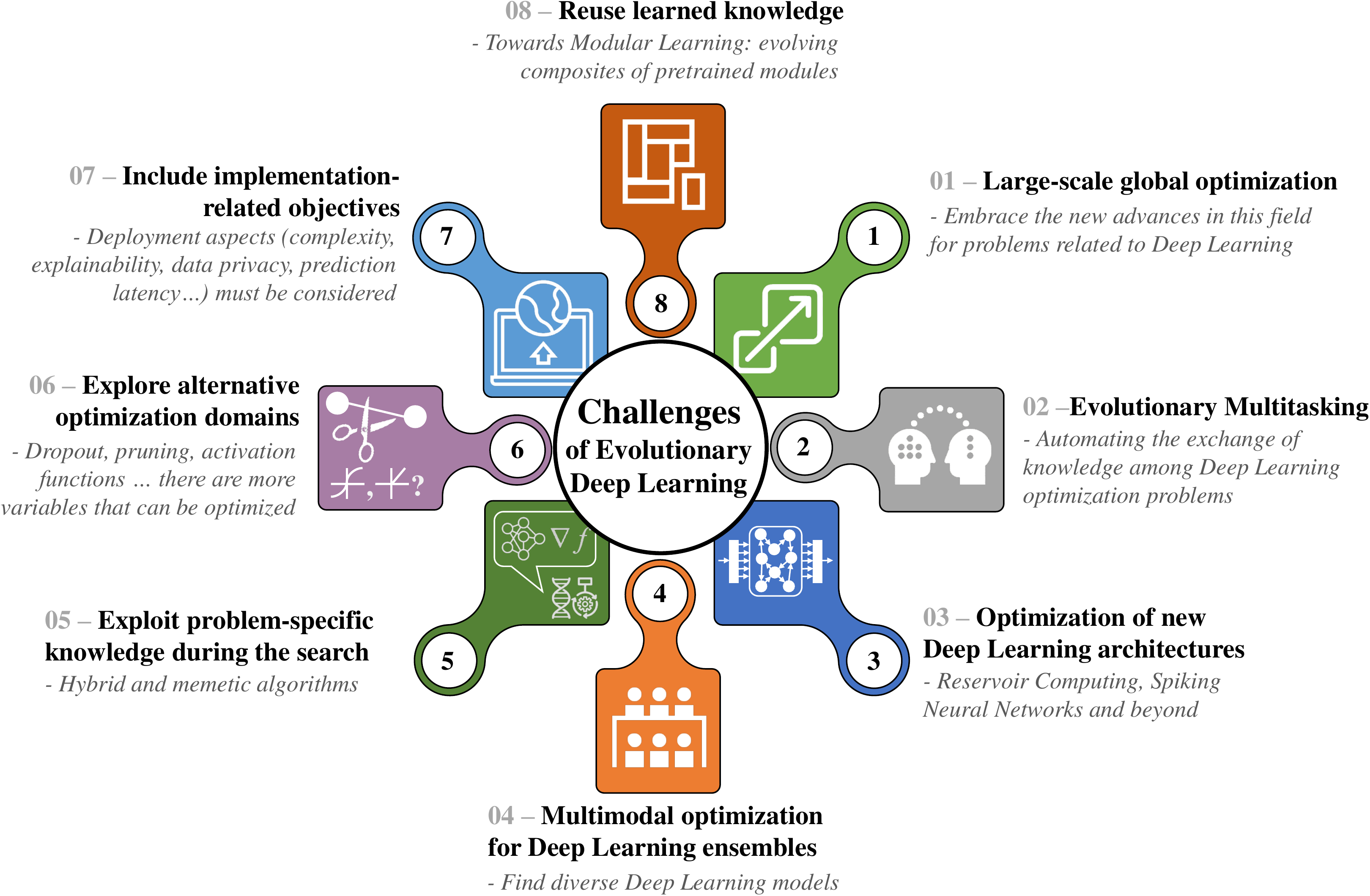}
	\caption{Challenges and research directions envisioned for Evolutionary Computation and Swarm Intelligence for the optimization of Deep Learning models.}
	\label{fig:challenges}
\end{figure}

In this section we summarize several challenges and research directions that should be under the target of future efforts conducted in this area. Such challenges are schematically depicted in Figure \ref{fig:challenges}, and contribute to the last two questions targeted by this overview: what can be done in future investigations on the confluence between bio-inspired optimization and Deep Learning, and what should future research efforts be conducted for?

\subsection{Large-scale optimization for model training} \label{ssec:cha_1}

The design of bio-inspired algorithms capable of efficiently tackling large-scale optimization problems seems to be one of the critical points that require further developments to train Deep Learning models of realistic complexity levels. This is the reason for the selection of SHADE-ILS as the search algorithm in the second case of study. Indeed, SHADE-ILS remains nowadays as one of the most competitive proposals for large-scale global optimization, and is regularly considered as a baseline for competitions and benchmarks. 

However, as in other research areas related to meta-heuristics, many advances in large-scale global optimization are regularly contributed to the community, featuring sophisticated ways to infer and exploit the correlation between variables during the search process (\emph{interaction learning}). Improving this feature in large-scale solvers is often the main target of new proposals, either in an implicit fashion (as in Estimation of Distribution Algorithms, and Bayesian Optimization) or explicitly via grouping, statistical correlation-based methods, decomposition or other assorted means \cite{mahdavi2015metaheuristics}. 

Unfortunately, our analysis has revealed that most works related to trainable parameter optimization have resorted to off-the-shelf variants of bio-inspired solvers. Consequently, no consideration is made about the interactions between variables (weights, biases) that are known to occur due to the neural connections throughout the multiple neural layers. This motivates a closer look to be taken at new advances in large-scale global optimization, for both single- and multi-objective optimization problems \cite{yi2020behavior}. Given the upsurge of Deep Learning problems in which more than one objective is established \cite{kim2017nemo, lu2018nsga, lu2019multi, liang2018evolutionary, li2020parallel}, the use of multi-objective solvers for large-scale optimization seems to be a natural choice.

\subsection{Evolutionary multitasking} \label{ssec:cha_2}

An interesting research area has revolved lately around the design of evolutionary multitasking algorithms capable of simultaneously addressing several optimization problems within a unique search process that exploits the complementarities and synergies existing among such problems \cite{ong2016evolutionary}. The challenge in this area is to develop intelligent optimization methods that not only promote the exchange of knowledge among candidate solutions corresponding to related problems, but also prevents the convergence of the search from being affected by \emph{counterproductive} knowledge transfers among unrelated tasks.

When framed within the current study, the adoption of large-scale evolutionary multitasking to optimize simultaneously different Deep Learning models can boost even further the possibilities foreseen for the intersection between Transfer Learning and bio-inspired optimization. For instance, the transfer of pretrained modules between tasks can be conceived as a crossover strategy between networks partially evolved for undertaking different tasks. Similarly, the exchange of the parameters values between Deep Learning models can be also automated via evolutionary multitasking towards evolving behavioral policies for different reinforcement learning tasks \cite{martinez2020simultaneously}. Evolutionary multitasking has been also used to achieve modular network topologies \cite{chandra2018evolutionary}. The relative youth and promising results shown by evolutionary multitasking techniques are a sign of objective evidence that optimization problems related to Deep Learning should be explored via these techniques, e.g. by leveraging the straightforward exchange of knowledge among networks allowed by their hierarchically layered structure. Furthermore, developments in multi-objective evolutionary multi-tasking \cite{gupta2016multiobjective,yao2020multiobjective,bali2020cognizant} open up further opportunities towards considering other objectives beyond accuracy of relevance for Deep Learning, such as the complexity of evolved topologies.

\subsection{Optimization of new Deep Learning architectures} \label{ssec:cha_3}

Most of the reviewed literature on bio-inspired algorithms for Deep Learning has focused on traditional forms of neural computation, including convolutional filters and recurrent units. However, this major activity has set apart the optimization of other neural network flavors, for which \emph{Deep} (multi-layered) versions have been proposed over the years. Such alternative deep architectures have fewer optimization variables, hence favoring the use of naive bio-inspired solvers for the different problems that can be formulated on them.

One of such neural families is \emph{Reservoir Computing}, which comprises a number of recurrent neural networks where only the parameters of the output layer (the readout layer) are learned. The parameters of the rest of recurrent neurons (the \emph{reservoir} are randomly initialized subject to some stability constraints, and kept fixed while the readout layer is trained \cite{lukovsevivcius2009reservoir}. Some works have been reported in the last couple of years dealing with the optimization of Reservoir Computing models, such as the composition of the reservoir, connectivity and hierarchical structure of Echo State Networks via Genetic Algorithms \cite{dale2018neuroevolution}, or the structural hyper-parameter optimization of Liquid State Machines \cite{zhou2019evolutionary,zhou2020surrogate} and Echo State Networks \cite{liu2020nonlinear} using an adapted version of the Covariance Matrix Adaptation Evolution Strategy (CMA-ES) solver. The relatively recent advent of Deep versions of Reservoir Computing models \cite{gallicchio2017deep} unfolds an interesting research playground over which to propose new bio-inspired solvers for topology and hyper-parameter optimization.

Despite more scarcely, optimization problems related other families of neural computation models have also been approached via Evolutionary Computation and Swarm Intelligence. The most remarkable case is the family of Spiking Neural Networks, in which topology and structural hyper-parameters have been addressed by means of different bio-inspired solvers \cite{vazquez2011training,schuman2016evolutionary}. Training of synapses in spiking neural architectures has been also tackled in \cite{vazquez2015training,carino2016spiking}. We fully concur with the prospects outlined in the recent overview on training methods for Spiking Neural Networks \cite{wang2020supervised}: efficient large-scale training methods should be investigated for new variants of these models, such as spiking deep belief networks and spiking convolutional neural networks. 

\subsection{Multimodal optimization for Deep Learning ensembles} \label{ssec:cha_4}

Another interesting research path stems from the adoption of niching methods used in bio-inspired algorithms for multi-modal problems for the construction of Deep Learning ensembles (also referred to as \emph{committees}. Indeed, the evolved population of candidate networks can be employed to retain near-optimal yet diverse Deep Learning model configurations. Such a diversity can emerge from different evolved topologies and/or values of their (hyper-)parameters. Such retained network configurations can be assembled into a committee, allowing for a robust fusion of their issued decisions. 

Work in this direction has been published recently in \cite{baldominos2019hybridizing}, where a niching mechanism is used to penalize individuals that are more similar to others in the population are penalized. Network configurations remaining in the population after the search are then combined together via majority voting, showing a significant improvement in performance with respect to the use of a single evolved Deep Learning model. Multi-modal optimization methods can also be combined with diversity induction techniques (e.g. Novelty Search \cite{martinez2019hybridizing}) to promote an efficient exploration and discovery of multiple global optima over strongly multi-modal search spaces, just like the ones known to characterize Deep Learning optimization problems. 

\subsection{Exploitation of problem-specific knowledge during the search} \label{ssec:cha_5}

In light of our experiments, we believe that the research community working on bio-inspired optimization should \emph{go memetic} and exploit the rich structural properties of neural networks when designing new algorithmic approaches for any of the problems related to Deep Learning. A diversity of strategies can be followed for this purpose, from simplest layer-wise search schedules as the ones proposed in Case of Study I, to more sophisticated means like iterating between meta-heuristic and gradient-based search, or the use of similarity measures between neural layers \cite{kornblith2019similarity} for controlling the behavior of search operators in topology and/or structural hyper-parameter optimization. 

By blending together the global search capabilities of bio-inspired meta-heuristics and the problem-specific knowledge embedded in e.g. back-propagated gradients, a major performance improvement can be obtained, effectively overcoming known issues such as gradient vanishing or slow convergence. This is actually the approach followed in \cite{munoz2018framework}, yet validated with small network sizes. Another problem-specific aspect hybridized with bio-inspired algorithms can be found in \cite{conti2018improving}. In this work, the need for inducing curiosity in reinforcement learning models (specially in environments with sparse rewards) is realized by not driving the search of the learning agent with the reward objective, but rather with a measure of the \emph{behavioral novelty} of the resulting policy. Other examples of hybrid methods are \cite{pourchot2018cem,maziarz2018evolutionary,cui2018evolutionary,Garcia2020hybrid}.

This can be conceived as another example of the importance of considering the particularities of the learning problem in the design of bio-inspired algorithms. We firmly advocate for more proposals along this direction when addressing optimization problems in Deep Learning.

\subsection{Exploration of alternative optimization domains} \label{ssec:cha_6}

The taxonomy around which our literature analysis has been organized considers three possible optimization domains on which a problem related to Deep Learning can be formulated: 1) topological variables, namely, the number and type of layers that compose the model; 2) hyper-parameters, which establish the details of the layers (structural hyper-parameters) and the optimization algorithm chosen for training the model (training hyper-parameters); and 3) trainable parameters (weights and biases). This threefold categorization collectively reflects most contributions reported to date in this research area. 

However, there are more optimization domains related to Deep Learning that can be tackled with bio-inspired solvers. One of them is pruning, e.g. the selective removal of connections among neurons for different purposes, from regularization against overfitting to a lower computational burden of gradient back-propagation solvers. Different pruning strategies can be developed to select which connections to drop at every layer of the Deep Learning model, which have been reviewed in recent comprehensive surveys on this topic \cite{blalock2020state,labach2019survey}. To the best of our knowledge, the use of Evolutionary Computation and Swarm Intelligence for neural pruning has been only done at the level of optimizing dropout rates. When turning the focus on more fine-grained pruning strategies, large-scale optimization algorithms can be used to determine the subset of neural connections that must be discarded to achieve a good trade-off between generalization performance and network compactness. Recent findings in the application of genetic algorithms to convolutional channel selection \cite{wang2020network} should be followed by further studies evaluating the scales at which network pruning with meta-heuristics can be realized.

Other optimization domains for which Evolutionary Computation and Swarm Intelligence methods can be applied include the tailored design of activation functions \cite{bingham2020evolutionary}, or the fusion of decisions issued within Deep Learning ensembles \cite{bochinski2017hyper}. Definitely, these problems and other ones still to be proposed lay a magnificent panorama for bio-inspired optimization. 

\subsection{Inclusion of multiple implementation-related objectives} \label{ssec:cha_7}

When evolving Deep Learning models, objectives and constraints related to the application scenario on which they are to be deployed should be considered in the optimization problem. For instance, many software libraries and embedded electronic chips can be found nowadays in the market for the implementation and execution of neural network models in constrained computing devices, such as Internet of Things (IoT) sensors \cite{mohammadi2018deep}. Likewise, real-world applications such as autonomous vehicular driving, remote precision surgery or wearable sensors restrict severely which Deep Learning models can be rolled out in their equipment. This suggests that a closer look should be taken at the complexity of evolved Deep Learning models during the optimization of their topology and hyper-parameters, among other domains.

A similar elaboration can be also made in regards to the progressive maturity of new paradigms such as Federated Learning \cite{yang2019federated} and Edge Computing \cite{chen2019deep}. In these paradigms not only local models lack the amount of computational resources needed to run complex Deep Learning models, but also additional objectives are imposed. Efficient (incremental) training algorithms, energy consumed by the model \cite{garcia2019estimation}, data privacy preservation \cite{nasr2018comprehensive,rodriguez2020federated}, robustness against adversarial attacks \cite{bhagoji2019analyzing}, or the explainability and accountability of decisions \cite{arrieta2020explainable} are some illustrative examples of the eventual confluence of multiple implementation-related objectives in Deep Learning optimization. However, these factors are rarely taken into account in current approaches for evolving Deep Learning models. Most of them rely just on accuracy or any other measure of predictive performance.

This being said, Deep Learning models should be evolved by considering together several objectives and constraints as the ones exemplified above. To this end, we foresee that bio-inspired algorithms for multi-objective optimization \cite{coello2007evolutionary} can play a differential role in the future of Deep Learning. This branch of bio-inspired optimization, along with techniques devised to handle constraints during the search \cite{mezura2011constraint}, can catalyze the practical deployment of Deep Learning models by addressing the improvement of the model's generalization performance together with implementation-related objectives and constraints.

\subsection{Reuse of learned knowledge: towards modular learning} \label{ssec:cha_8}

The reutilization of pretrained modules between models corresponding to related learning tasks is at the core of Transfer Learning \cite{weiss2016survey}, whose most straightforward strategy is to reuse parts of a model developed for a task as the starting point for a model devised for another task. Such parts are often conceived as structural fragments of the network, particularly those capturing high-level features from the input to the Deep Learning model. Features corresponding to those parts are more easily reusable among tasks due to their high-level nature. In image classification, for instance, features learned by the first layers of the network are borders and broad shapes that could be of help for many different tasks. Therefore, it is intuitive to think that the output of such layers can be of help in other tasks for which annotated data instances are scarce.

The availability of Deep Learning models trained on different datasets and the effectiveness of just transferring layers between models corresponding to different tasks suggest a very interesting research path: expanding further the search space of topology and structural hyper-parameter search to also consider pretrained modules. The inclusion of such modules in the alphabet of possible layers could yield a major boost of Deep Learning models, specially for those cases with few labeled data. Furthermore, the flexibility of bio-inspired algorithms when designing the encoding strategy that represents networks during the search could allow achieving finer levels of granularity in the knowledge imported from such pretrained modules, to the scales of convolutional filters or recurrent units. There is a great opportunity to bring together transfer learning and topology/structural hyper-parameter optimization around the same goal: to evolve and discover Deep Learning models of superior performance.

\section{Conclusions and Outlook} \label{sec:conclusions}

This paper has presented a comprehensive and critical review on the use of Evolutionary Computation and Swarm Intelligence approaches to the topological, hyper-parametric and/or trainable parameter optimization of Deep Learning models. As we have previously indicated, the paper is focused on three axes: a) definition of optimization problems in Deep Learning and taxonomy; b) a critical methodological analysis of the related literature and two cases of study, allowing to prescribe learned lessons and recommendations for good practices; and c) an enumeration of challenges and new directions of research. We highlight the aforementioned two cases of study, providing factual results on the performance of bio-inspired optimization algorithms when applied to the architectural design, hyper-parameter tuning and training of Deep Learning models. 

Our elaborations made throughout these three axes have yielded informed conclusions and insights about the four fundamental questions posed in the introduction, which round up the critical review of the field targeted in this overview. We synthesize below our responses to such questions, in the form of reflections stemming from the aforementioned axes: 
\begin{enumerate}[leftmargin=*]
\item \textbf{Why} are bio-inspired algorithms of interest for the optimization of Deep Learning models?

The increased scales and diversity of neural layers of modern Deep Learning approaches have lately reactivated the global interest in Deep Learning optimization with bio-inspired algorithms, as a means to automate efficiently the processes of designing their topology, tuning their hyper-parameters and learning their parameters. Such processes can be formulated as complex optimization problems, motivating the adoption of bio-inspired algorithms for solving them efficiently. Furthermore, the renowned global search capability of Evolutionary Computation and Swarm Intelligence methods makes them a suitable choice to deal with complex search spaces as those characterizing Deep Learning problems. Finally, the flexibility of bio-inspired solvers to be hybridized with problem-specific search methods is another reason supporting the hypothesis that Deep Learning optimization can largely benefit from them. 

In conclusion, we observe solid grounds for this synergistic fusion of technologies, which has so far stimulated the community to tackle optimization problems in Deep Learning using bio-inspired algorithms. However, we recognize that this fusion has not yet achieved results that are truly a step forward in terms of quality and objective achievement. This is still a lost race of bio-inspired optimization algorithms with respect to gradient-based solvers, which remain as the horse at the head of the race.
    
\item \textbf{How} should research studies falling in the intersection between bio-inspired optimization and Deep Learning be made? 
    
Our discussion on the results obtained in our cases of study suggest that bio-inspired algorithms can be used for topological and/or hyper-parameter optimization, yet performing worse than other methods when comparisons are fair in terms of search space and complexity. Furthermore, our experiments have also revealed that even competitive bio-inspired solvers for large-scale global optimization are outperformed by conventional gradient-based solvers for trainable parameter optimization. These results, along with several issues detected in the literature (most notably, the unrealistic scales of the evolved model and the dataset/task under consideration), support our claims that there is a large space for improvement in this research area. 
    
On a prescriptive note, we have identified several learned lessons and recommendations that trace \emph{how} research should be done to reach solid conclusions and sound achievements. We highlight below the most relevant ones:
\begin{itemize}[leftmargin=*]
\item Good methodological practices when designing experiments and benchmarks between different solvers, including realistic datasets and models, fairness in terms of computational complexity, assessment of the significance between performance gaps and reported performance scores over test instances, among others.

\item A closer attention at encoding strategies for topology optimization that account for the validity of the composition of layers that they represent.

\item A clear definition of the variable search ranges in structural and training hyper-parameter optimization, so that differences emerging between solvers can be attributed exclusively to their search efficiency.

\item The exploitation of problem-specific knowledge in trainable parameter optimization: the interactions between trainable parameters imposed by the hierarchical structure of neural connections should be exploited further by bio-inspired solvers for them to step out from the shadows of their application to the training of Deep Learning models. 
\end{itemize}

\item \textbf{What} can be done in future investigations on this topic?

In this regard we have underscored the need for overcoming the computational inefficiency observed in current bio-inspired optimization algorithms with respect to gradient-based solvers. It is known that these latter solvers have also their own drawbacks: they require differentiable loss formulations, they are sensitive to vanishing and exploding gradients and they are prone to local optima in non-convex search spaces. There are well-founded reasons why bio-inspired solvers can be a firm alternative to gradient-based methods, but more efficient designs and/or better performing implementations of bio-inspired approaches should be under active investigation in the future for them to become a practical choice for Deep Learning model training.

We have also stressed on other subareas in Evolutionary Computation and Swarm Intelligence of utmost interest for their application to Deep Learning optimization. Large-scale global optimization for training Deep Learning models of realistic complexity, or multi-modal optimization for the automated construction of Deep Learning ensembles, are just a few examples of the myriad of research niches in bio-inspired optimization that have not been explored yet. Evolutionary multitasking is another interesting direction to follow when approaching several Deep Learning optimization problems at the same time, mostly when such problems are related to each other and share a significant degree of overlap in their solutions (as occurs in transfer learning). Finally, optimization problems formulated on alternative Deep Learning models seem to have received less attention to date, and unleash further opportunities for bio-inspired optimization.

An additional research direction in this fusion of technologies has been identified in the existence of other variables and domains in which optimization problems can be formulated. Ensemble construction, network pruning or selective dropout strategies can also be described as optimization problems, favoring the adoption of bio-inspired optimization algorithms for their efficient solving. 

\item \textbf{What} should future research efforts be conducted \textbf{for}?

There is a strong incentive for which Deep Learning optimization should be approached via bio-inspired algorithms in the future: the consideration of additional objectives and constraints linked to the application scenario on which the model is to be deployed. Aspects such as the available computational resources, the need for periodically updating the model, or the time taken by the model for issuing a prediction should be considered during the design of the model for it to be of practical value. 

Furthermore, new paradigms such as Edge Computing, explainable Artificial Intelligence and Federated Learning have underpinned the need for taking into account other objectives beyond the accuracy of the model. Aspects such as data privacy preservation, the explainability of decisions issued by the model, the non-stationary nature of data at the edge, or the complexity of the learning algorithm can be formulated as additional objectives for the design of the model, impacting on its topology, hyper-parameter values and other optimization variables. 

These arguments, combined with the flexibility of bio-inspired algorithms to deal with multiple conflicting objectives, can be a primary \emph{what for?} driver for adopting them to evolve Deep Learning models in practical settings. Specific scenarios and contexts that require ad-hoc designs to be evaluated with multiple objectives can and should also open the door to evolutionary Deep Learning models towards meeting imposed goals in terms of efficiency, performance and other application-related objectives.

\end{enumerate}

All in all, we hope that the material and insights given in this work serve as a reference material for readers willing to arrive at this research area with a clear and thorough understanding of its recent past, current status, and potential in years to come. There are promising evidences that Evolutionary Computation and Swarm Intelligence can tackle optimization problems related to Deep Learning, but we firmly believe that they are not close to maturity, nor do they justify yet the replacement of other solvers used for the same purposes. Nevertheless, this is the role of research itself: to build upon the shadows of knowledge and bring light through scientific achievements. This survey has just lit a candle to illuminate this path through the field of Evolutionary Deep Learning.

\section*{Acknowledgments}

Aritz D. Martinez, Esther Villar-Rodriguez, Eneko Osaba and Javier Del Ser acknowledge the funding support received from the Basque Government through the EMAITEK and ELKARTEK programs (3KIA project, KK-2020/00049), and the Spanish \textit{Centro para el Desarrollo Tecnologico Industrial} (CDTI, Ministry of Science and Innovation) through the \emph{Red Cervera} Programme (AI4ES project). Javier Del Ser also acknowledges funding support from the Consolidated Research Group MATHMODE (IT1294-19) granted by the Department of Education of the Basque Government. Siham Tabik, Daniel Molina and Francisco Herrera would like to thank the Spanish Government for its funding support (SMART-DaSCI project, TIN2017-89517-P), as well as the BBVA Foundation through its \emph{Ayudas Fundaci\'on BBVA a Equipos de Investigaci\'on Cient\'ifica} 2018 call (DeepSCOP project).

\appendix
\section{Deep Learning Models} \label{ssec:DLmodels}
\renewcommand{\thefigure}{A.\arabic{figure}}
\setcounter{figure}{0}

The spectrum of Deep Learning models is certainly huge nowadays, with new learning variants for different tasks proposed continuously by the community. In what follows we provide a brief introduction of the DL models in which Evolutionary Algorithms and Swarm Intelligence methods have been mostly used for addressing the above problems. 

\subsection*{Deep Boltzman Machines and Deep Belief Nets}

Deep Boltzman Machines (DBMs) and Deep Belief Nets (DBNs) are considered as generative probabilistic models comprised by a stacked hierarchy of Restricted Boltzmann Machine (RBM) layers. Unlike classical Boltzmann Machines, RBMs have no intra-layer connection between nodes, and comprise two layers of fully-connected neurons (visible and hidden), in charge of learning the probability distribution of a set of binary-valued inputs. DBMs are constructed as a series of RBMs stacked on top of each other, whereas DBNs are hybrid models whose interactions are composed of indirect connections at the top layers (RBM) and downward directed belief nets between the lower ones. 
    \begin{figure}[ht!]
    	\centering
    	\begin{tabular}{ccccc}
    	\includegraphics[width=0.3\columnwidth]{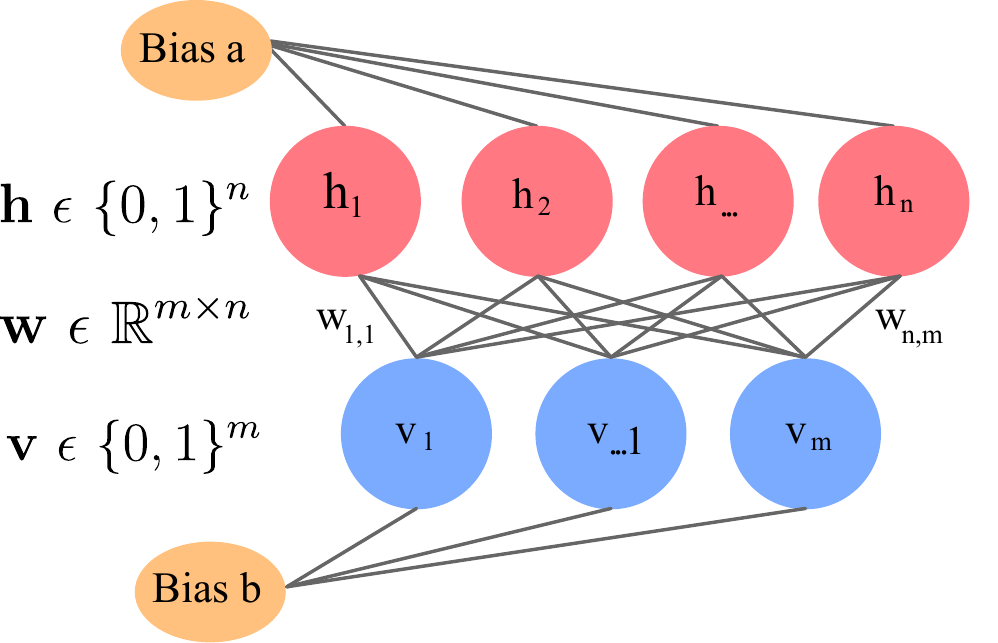} & &
    	\includegraphics[width=0.2\columnwidth]{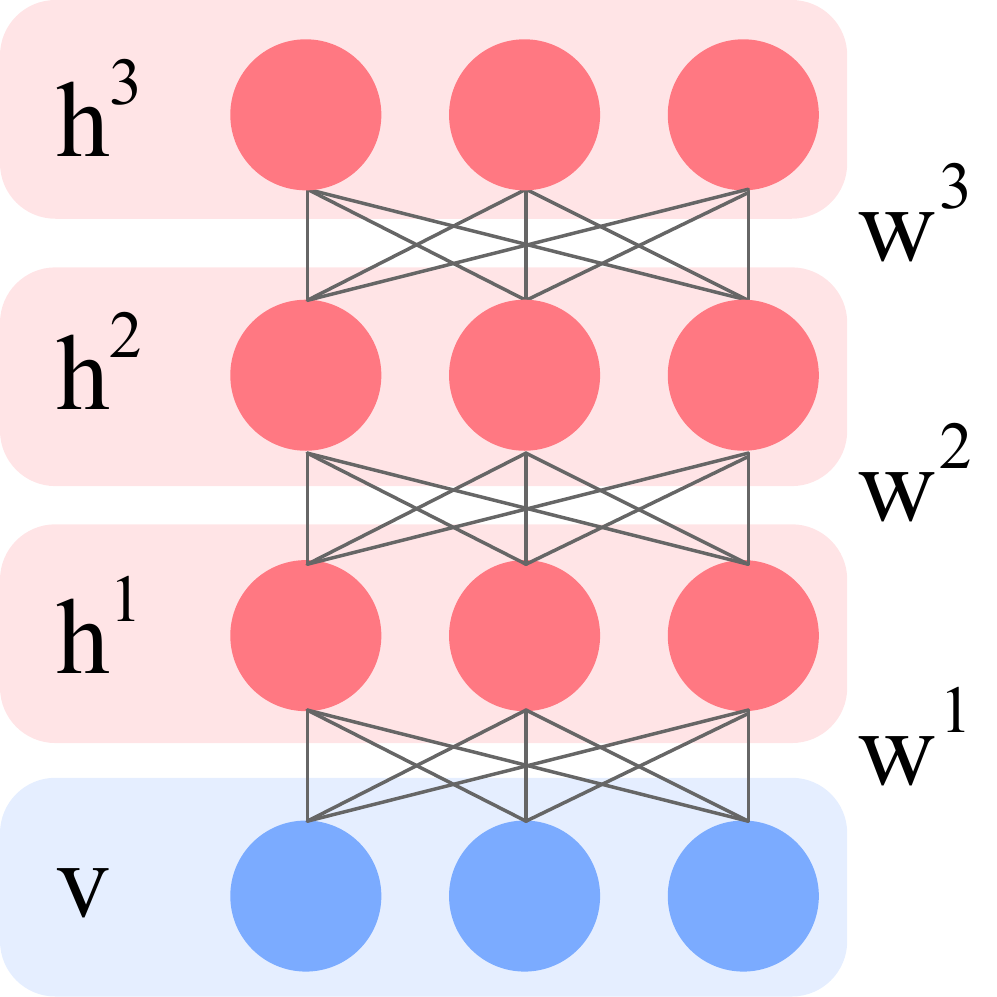} & &
    	\includegraphics[width=0.2\columnwidth]{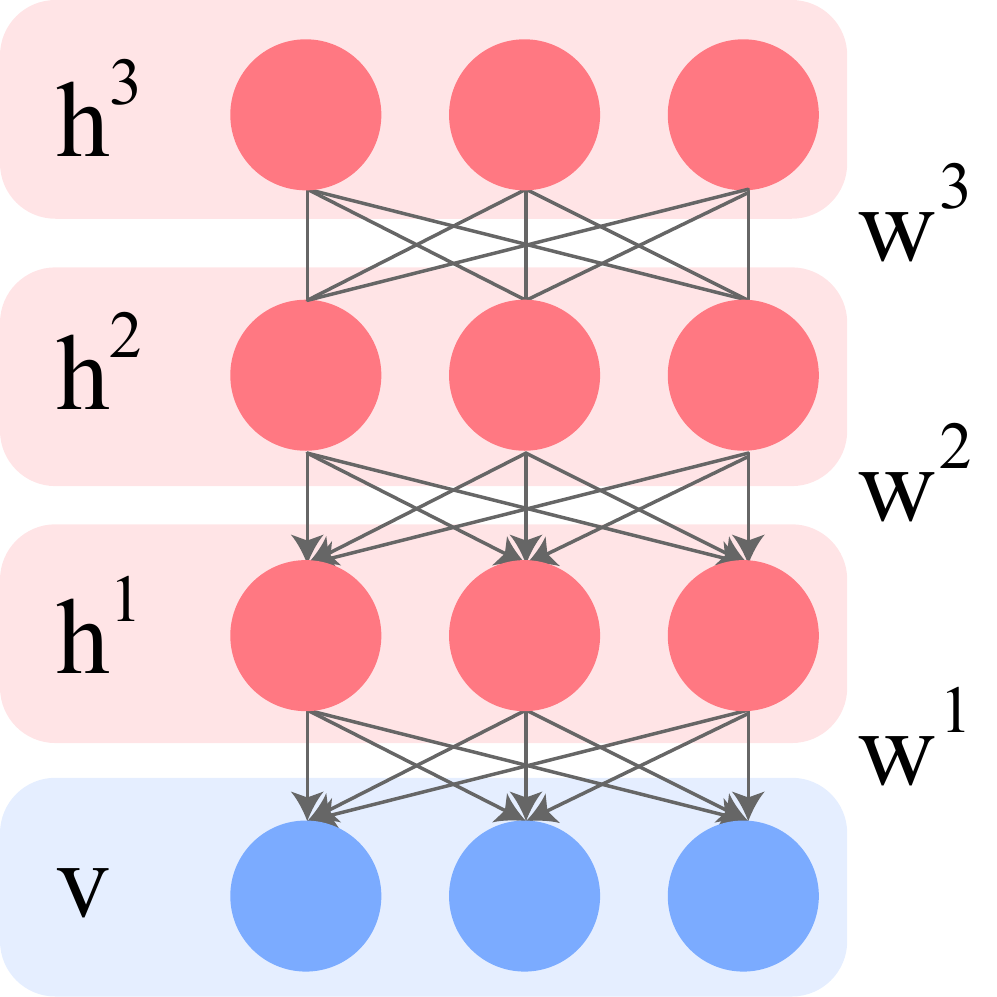}\\
    	(a) & & (b) & & (c)
    	\end{tabular}
    	\caption{Architectures of (a) a Restricted Boltzmann Machine; (b) a Deep Boltzman Machine; and (c) a Deep Belief Net.}
    	\label{fig:DBMDBN}
    \end{figure}
    
The learning process of DBM and DBN is conducted by a greedy layer-by-layer pre-training approach, followed by a discriminative fine-tuning process performed by a back-propagation algorithm according to labelled data provided in the training set (supervised learning). This procedure modifies the learned layerwise parameters to ensure the appropriate learning of the labels/desired outputs. 

Both DBMs and DBNs have restricted structural topologies. Therefore, the optimization tasks that have been central for these models relate to the proper selection of structural (number of hidden units) and training hyper-parameters (training epochs and learning rate, among others). Efficiently tuning these hyper-parameters has been a subject of intense scientific research over the history of these models, as we show in our literature study. 
    
\subsection*{Autoencoders}

Autoencoders (AEs) are composed by two different typically mirror frameworks, named encoder and decoder, and are deemed as the inspiration behind $G$ models. Briefly explained, the principal goal of AEs is to learn an encoded representation of their input data. In order to achieve that, the reconstruction error between the input data and the generated output is minimized by making the encoder create condensed low-dimensional data abstractions such that, when decoded, the original input is reconstructed with high fidelity. This process assist the model to learn critical and latent features present in the data. Thus, AEs are composed by four core parts:
\begin{itemize}[leftmargin=*]
\item \textit{Encoder}: in charge of projecting high-dimensional data down to a subspace of encoded representation.
\item \textit{Bottleneck}: the most compressed representation of the input data.
\item \textit{Decoder}: the model learns the process for the reconstruction of the data.
\item \textit{Reconstruction Loss}: responsible for assessing the adequacy of the solutions offered by the decoder, measuring how close the output is regarding the input data.
\end{itemize}
    \begin{figure}[ht!]
    	\centering
    	\includegraphics[width=0.6\columnwidth]{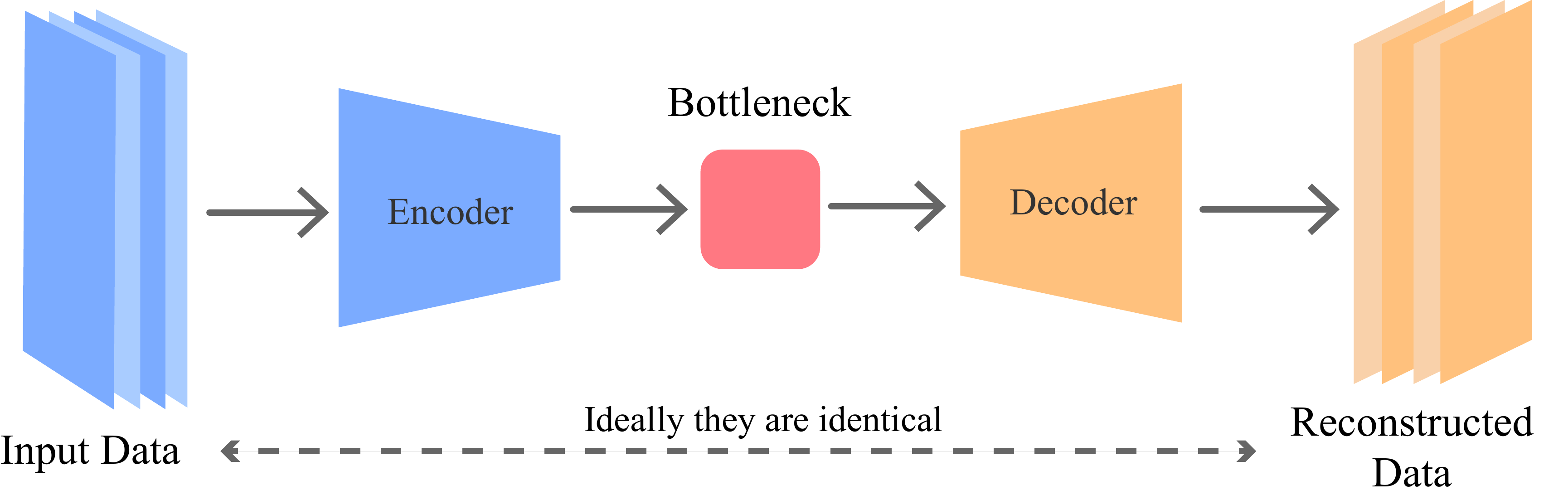}
    	\caption{Architecture of a basic Autoencoder.}
    	\label{fig:auto}
    \end{figure}

It is remarkable that a wide variety of AEs can be found along the literature, such as Convolutional, Variational or Denoising AEs, among many other approaches. Having each of them their specific characteristics and optimizable hyper-parameters. In all of them, the main structural hyper-parameter is the size of the bottleneck, also known as latent space size. Nevertheless, the fact that they have two separated networks to optimize, makes AEs attractive for topology optimization (Problem 1). Although the topological architectures of encoder and decoder parts are mirrored, they can be optimized on their own, as well as their connection patters. In terms of training, the AE is trained using gradient back-propagation. As such, a fully or partially substitution of the algorithm can be performed (Problems 3 and 4 as defined in Section \ref{sec:preliminaries}). 

\subsection*{Convolutional Neural Networks}

Commonly employed to to deal with pattern extraction from images and videos, Convolutional Neural Networks (CNNs) imitate the behavior of the visual cortex when processing images (\cite{enroth1966contrast,hochstein1976quantitative}) by laying out, in their simplest architectures, a hierarchical set of convolutional and pooling layers. Convolutional layers allow for an spatial analysis of their input data by applying a sliding dot product between the input tensor and a set of filters (also referred to as \emph{kernels}), which transforms the input tensor into a feature map. Pooling layers perform a spatial downscaling on its input tensor, making the model less sensitive to small variations and achieving better generalization properties of the overall model. 
\begin{figure}[ht!]
\centering
\includegraphics[width=0.8\columnwidth]{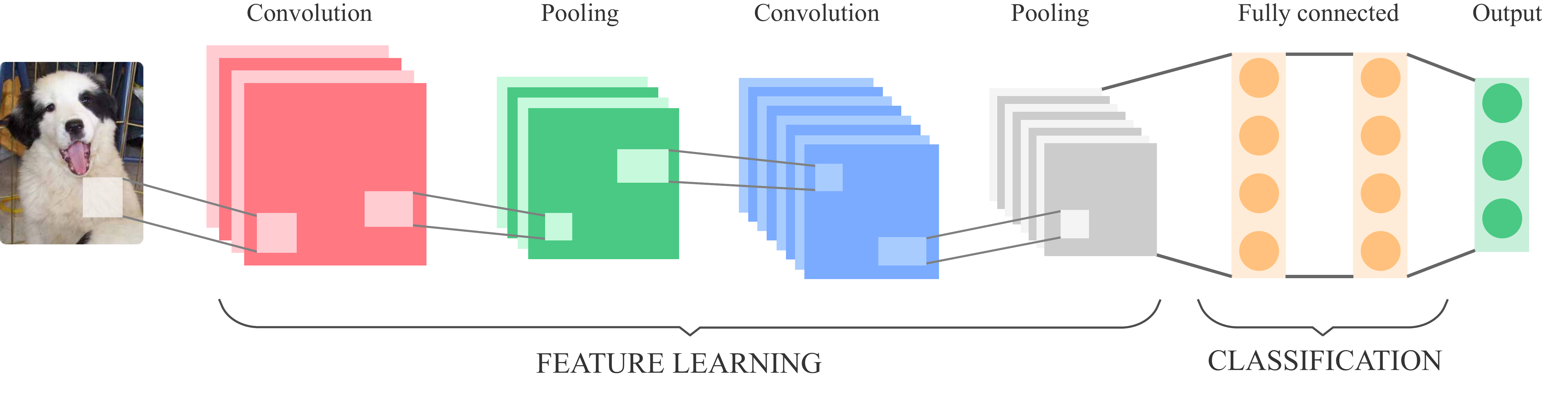}
\caption{Architecture of a Convolutional Neural Network.}
\label{fig:CNN}
\end{figure}
    
CNNs unleash a rich background for optimization due to the variety of layers composing their architecture. Indeed, there are multiple structural hyper-parameters related to those layers, such as \textit{kernel size}, \textit{padding size} or \textit{number of filters} in convolutional layers, or \textit{pool size} and \textit{stride number} in pooling layers, among others. CNNs are also suitable for topological optimization tasks, with the selection of layer types or the optimization of connection patterns among layers as the ones mostly addressed in the literature. The number of neurons/filters to allocate at each layer is considered as structural hyper-parameter optimization. Finally, the training optimization task related to CNNs is to substitute the conventionally used gradient back-propagation based solver by a meta-heuristic algorithm.
    
\subsection*{Recurrent Neural Networks}
    
Recurrent Neural Networks (RNNs) allow to learn from the relationship between items in sequence data, and are therefore typically used in speech and handwriting recognition, time series forecasting, natural language processing and video processing. RNNs are constructed as rolled neural networks with an internal state called \emph{memory}, which allows processing inputs of variable length, as well as previous predictions to be used as inputs for subsequent predictions steps over the input sequence. Designed as blocks of memory cells, they internally comprise a repeated neural network comprising two core elements: a hidden state and gates. Information flows through the cells, with gates regulating which portion of the input information should be thrown away or kept (\emph{forget gate}), retained in the cell's memory (\emph{input gate}), and revealed to the cell's output (\emph{output gate}). 
    \begin{figure}[ht!]
        \centering
        \includegraphics[width=0.3\columnwidth]{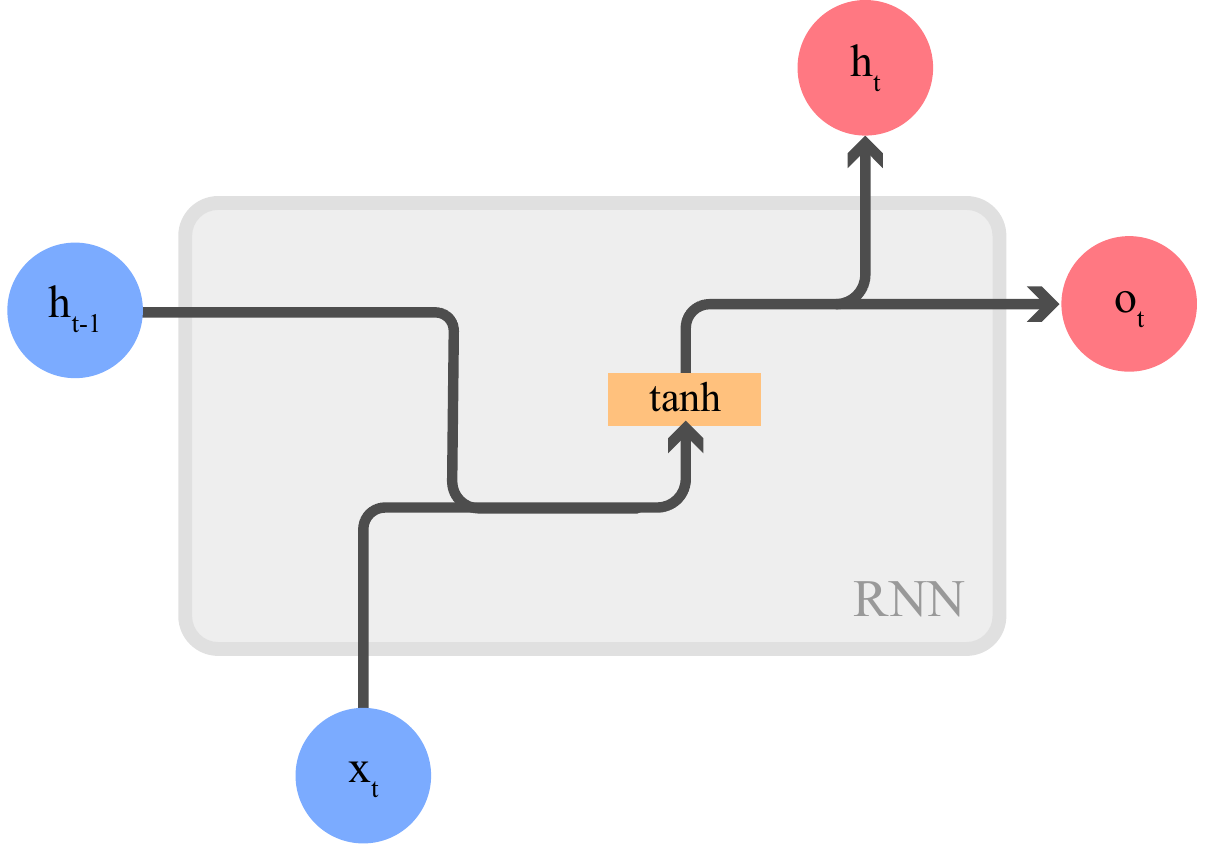}
        \includegraphics[width=0.3\columnwidth]{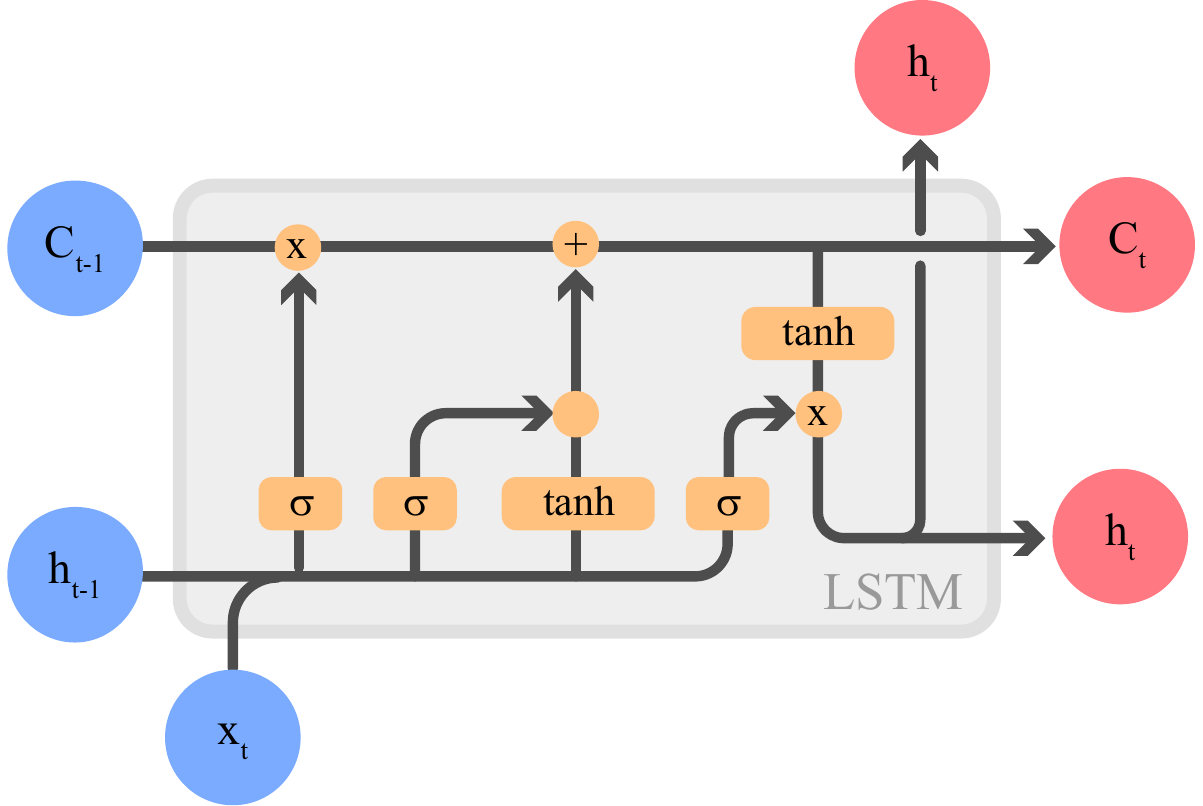}
        \includegraphics[width=0.3\columnwidth]{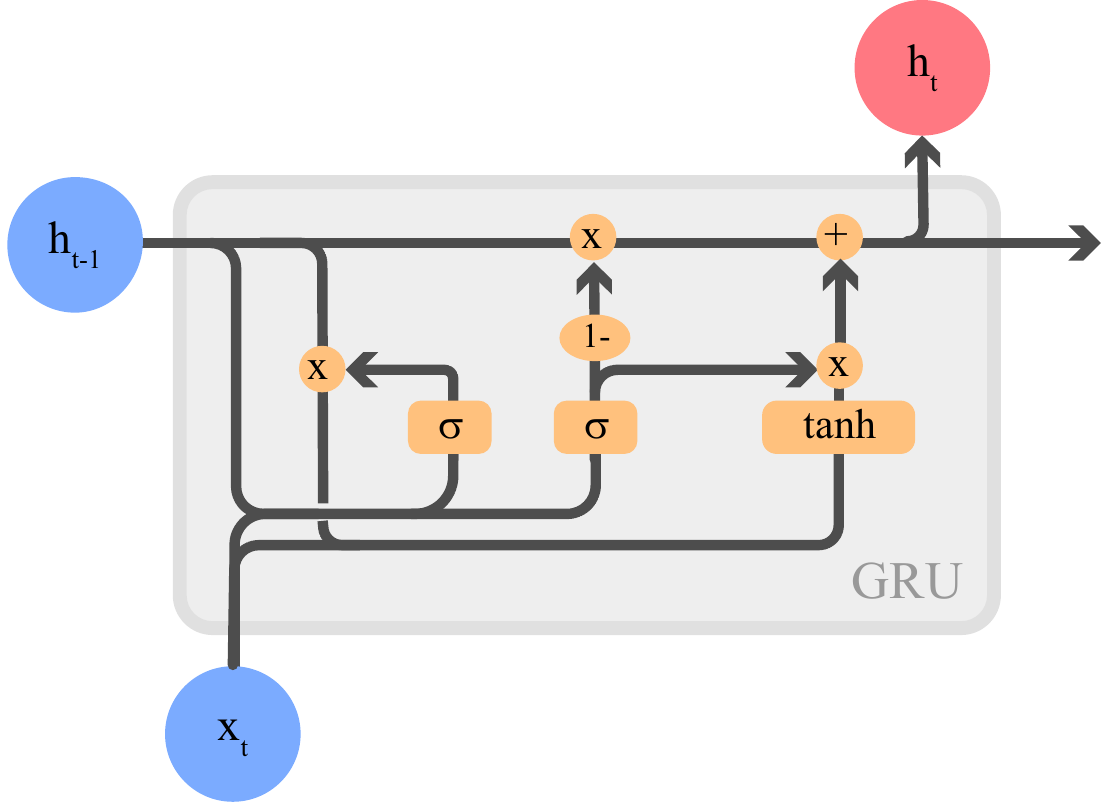}
        \caption{RNN, LSTM, GRU architectures.}
        \label{fig:rnn_architectures}
    \end{figure}

This is indeed the working mechanism underneath two of the most renowned RNN variants to date: Long Short-Term Memory units (LSTMs) and Gated Recurrent Units (GRUs). These two RNN models emerged as a workaround to the vanishing gradient problem faced by traditional RNNs. The most remarkable difference between simple RNNs and LSTMs/GRUs is that the latter incorporate a mechanism able to keep information along the units (temporally), letting the output at $t-1$ influence on the result at instant $t$. On the other hand, the main difference between GRUs and LSTMs is that LSTMs provides more flexibility in the control of the information flowing through the cell, by virtue of its internal cell state. On the contrary, GRU cells provide less control as they couple forget and input gates, but as a result have less parameters to be trained. Therefore, GRUs can be seen as a simplified version of LSTM cells. 

The performance of recurrent neural networks is closely related to the number of units taking part in the recursive connection. Because of their restricted architecture, the main task for this networks is the optimization of the number of hidden units (topology optimization, problem 1 in Section \ref{sec:preliminaries}). As noted in the literature analysis, dropout and recurrent dropout rates are also often optimized in recurrent architectures (structural hyper-parameter optimization, problem 2). 
    
\subsection*{Generative Adversarial Networks}

\begin{figure}[ht!]
        \centering
        \begin{tabular}{ccc}
        \includegraphics[width=0.52\columnwidth]{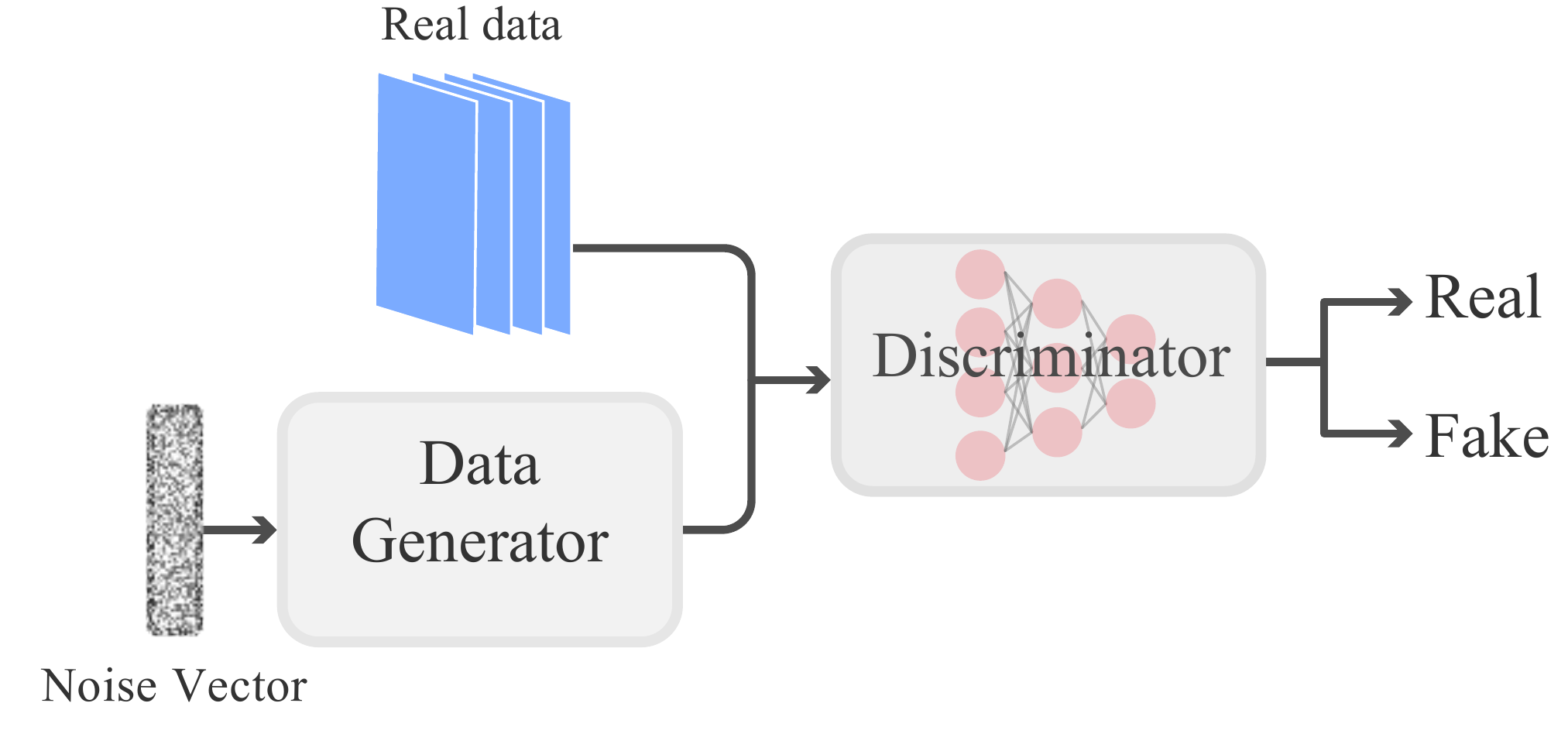} & &
        \includegraphics[width=0.42\columnwidth]{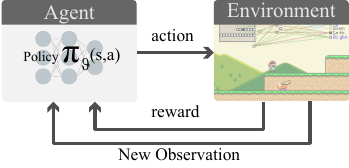} \\
        (a) & & (b)
        \end{tabular}
        \caption{General architecture of (a) a Generative Adversarial Network; and (b) a Deep Reinforcement Learning model.}
        \label{fig:GANDQL}
   \end{figure}

Generative Adversarial Networks (GANS) were introduced by Goodfellow et al. in \cite{goodfellow2014generative}. They are essentially two neural networks trying to beat each other, consequently progressing as per their opposite respective goals: a discriminator network, whose main objective is the classification of the input data and the evaluation of its authenticity; and a generator network, in charge of producing realistic synthetic data. The generator network attempts at ``fooling'' the discriminator network by generating new data instances increasingly closer to the genuine ones, whereas the discriminator aims at detecting whether the output of the discriminator is real or fake, progressively becoming more competent in this detection task. A schematic diagram of a typical GAN is depicted in Figure \ref{fig:GANDQL}.a.
   
Like CNNs, GAN architectures present a vast variety of optimization problems. Nevertheless, all of them are very similar to those of CNNs and RNNs described previously in this section. Depending on the nature of the data under consideration, both discriminator and generator may include convolutional and/or recurrent layers, thereby exposing the same need for optimizing the topology, hyper-parameters and/or trainable parameters of CNNs and RNNs. 
    
\subsection*{Deep Reinforcement Learning}

Deep Reinforcement Learning (DRL) is considered as an special case due to the amount of contributions in the field. It is not considered a model architecture, but rather a different learning paradigm in which the output of the Deep Learning model defines how an agent should interact with an environment. In general terms, the modules involved in a DRL approach are:
\begin{itemize}[leftmargin=*]
\item \textit{Environment}, namely, a scenario/asset/system modeled in a way that a human can interact with them through several variables. Upon the application of certain actions, the environment feeds the agent with observations, and evaluates the actions taken by the agent returning a value of reward.

\item \textit{Agent}, which embodies a policy that maps observations to actions. In the specific formulation of Deep Reinforcement Learning, the policy itself is intrinsically encoded in the parameters of a Deep Learning model, which is designed to take the action that maximizes the reward given a specific observation from the environment. 
\end{itemize}
    
The general architecture of this kind of approaches is illustrated in Figure \ref{fig:GANDQL}.b. Within the field of DRL many type of approaches are covered, from MLP to RNN, Dueling or Multiagent environments, which have been demonstrated to get good results when they are combined with meta-heuristic algorithms \cite{Stanley2005, Petroski2017, cardamone2009evolving}. Besides the discussed architectural and topological hyper-parameters that depend on the model that is adopted in the DQL approach (e.g. CNN, RNN or LSTM), the environment information is also susceptible to be manipulated to decrease the complexity, or to try to make the learning procedure computationally affordable.


\renewcommand{\thefigure}{B.\arabic{figure}}
\setcounter{figure}{0}

\section{Bio-inspired Optimization: Evolutionary Computation and Swarm Intelligence} \label{ssec:metah}

The need for search algorithms capable of efficiently dealing with the optimization problems arising from Deep Learning has stimulated an upsurge of literature proposing different solvers for this purpose. We now provide a brief overview of the optimization research area, with a focus on meta-heuristic algorithms that are inspired by biological sources of inspiration. For a more detailed overview of developments and prospects in this research area we refer to recent comprehensive reviews in \cite{del2019bio,dmolina:20}.

\begin{figure}[ht!]
\centering
\includegraphics[width=\columnwidth]{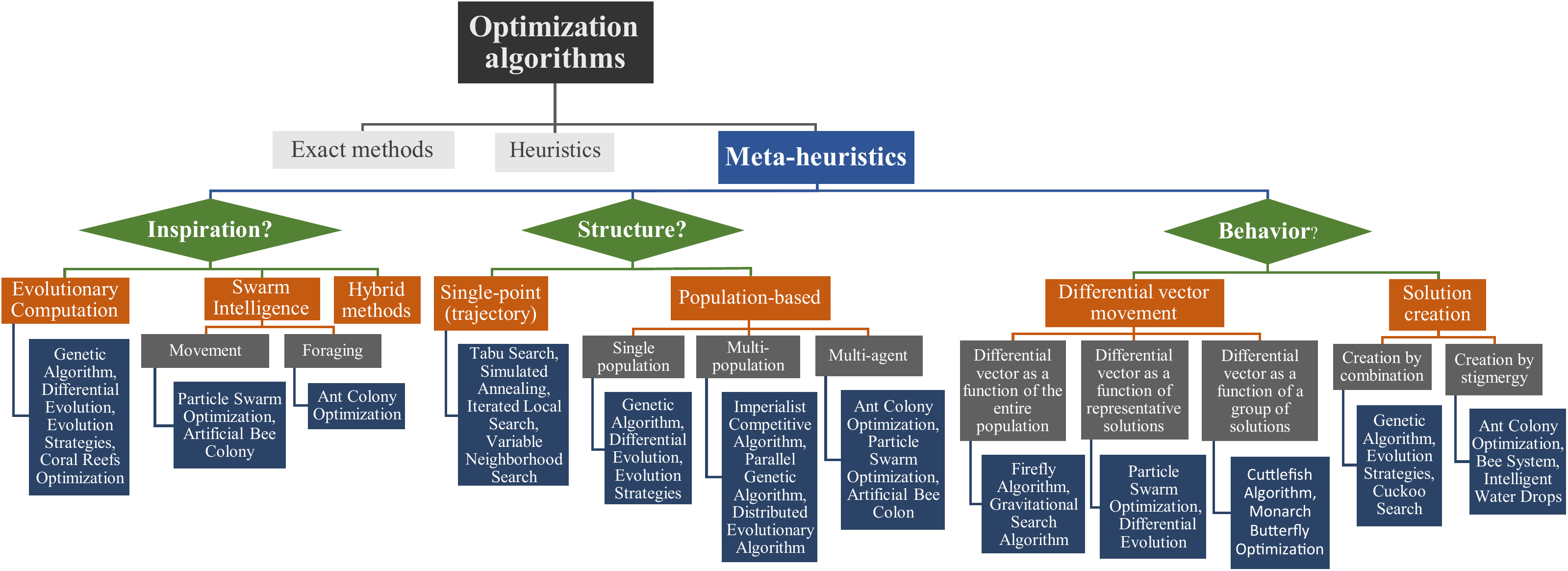}
\caption{Taxonomy of optimization algorithms, with a focus on meta-heuristic optimization algorithms as per the different criteria under which they can be classified. Some examples of algorithms are also given. For a further insight into nature and bio-inspired optimization approaches we encourage the reader to the extensive review in \cite{molina2020comprehensive}.}
\label{fig:taxonomy_optimization_algorithms}
\end{figure}

As shown in the taxonomy of Figure \ref{fig:taxonomy_optimization_algorithms}, optimization methods can be first grouped in three categories: exact methods, heuristics and meta-heuristics. Exact methods are those that always solve an optimization problem to optimality, either by exploring the entire space of solutions or by taking advantage of specific characteristics of the problem at hand (e.g. linearity, convexity). On the other hand, a heuristic search algorithm addresses a given optimization problem by resorting to knowledge related to the domain where the problem is formulated. By exploiting this domain-specific information in its search algorithm, a heuristic explores the space of feasible solutions efficiently, intensifying the search around the most promising areas as per the objective(s) under consideration. Finally, the third category corresponds to meta-heuristics, which lie at the core of this study. 

Briefly explained, a meta-heuristic optimization algorithm solves a problem using only general information and knowledge common to a wide variety of problems with similar characteristics \cite{boussaid2013survey}. Meta-heuristic algorithms explore the solution space by progressively learning how candidate solutions should be modified towards optimality, with the aim of reaching increasingly promising results disregarding the characteristics of the problem being tackled. Given their self-learning nature and their abstraction from the problem itself, meta-heuristic approaches are well-suited to deal with real-world problems featuring complex search spaces and even non-analytically defined objectives/constraints. This is in fact the reason why meta-heuristics have taken a prominent role when addressing the optimization problems underneath Deep Learning.

Deeper into the taxonomy of Figure \ref{fig:taxonomy_optimization_algorithms}, meta-heuristics can be further divided into different groups depending on several criteria. To begin with, we can distinguish between 1) single-point (also referred to as trajectory-based) meta-heuristic methods, which rely on the progressive improvement of a single solution to the problem by exploring its neighborhood under a set of movement operators (as in e.g. Tabu Search, TS \cite{glover1998tabu} or Simulated Annealing, SA \cite{kirkpatrick1983optimization}; and 2) population-based techniques, which maintain a set of possible solutions of the problem that interact with each other towards producing new solutions of increased quality (e.g. Genetic Algorithm, GA \cite{genetic1,genetic2}, Ant Colony Optimization, ACO \cite{dorigo2010ant} or Particle Swarm Optimization, PSO \cite{PSO}). This last category can be broken down in 2.1) multi-population techniques, where the population is divided into different that evolve separately, exchanging information periodically (for instance, the Imperialist Competitive Algorithm \cite{atashpaz2007imperialist}); 2.2) multi-agent methods, whose population is composed by multiple diverse agents with different roles that interact with each other towards optimality (e.g. Artificial Bee Colony, ABC \cite{ABC1}); and 2.3) single-population approaches, such as the aforementioned GA. At the same time, meta-heuristics can also be divided as per its search behavior, yielding A) differential vector movement based methods, which rely on the computation of a differential vector to move from a reference solution towards a new candidate; and B) solution creation based methods, which generate new solutions to explore the search space instead of moving existing ones. Other criteria can be adopted for organizing the enormous corpus of literature related to optimization meta-heuristics, such as the support of the optimization variables of problems that can be tackled by the meta-heuristic at hand (discrete/continuous/mixed), the scope of the search (global/local), or the stochastic/deterministic nature of their search operators, among others. 

Among these criteria, the inspiration underneath the search algorithm itself has sprung a vast area of research widely known as bio-inspired optimization \cite{yang2013swarm}. Over the last decades, a manifold of behavioral patterns observed in biological systems have been emulated to yield intelligent algorithms capable of mimicking the learning and adaptation capabilities of such biological systems to address complex computational problems. Therefore, a bio-inspired meta-heuristic algorithm can be categorized as such if its main search strategy gets partially or fully inspired by biological phenomena, such as the evolution of species, the echolocation of bats or the foraging behavior of ant colonies. A plethora of inspiring metaphors can be found nowadays in contributions dealing with new bio-inspired optimization algorithms, not without an ongoing controversy on the value of the metaphor itself for the novelty and scientific soundness of the reported methods \cite{del2019bio}.

Leaving such disputes aside, a research trend that has so far endured over the years is the hybridization of bio-inspired algorithms with problem-specific local search methods. The main reason behind this practice is to exploit the advantages of bio-inspired solvers, and to overcome their disadvantages when dealing with problems for which ad-hoc heuristics can be developed and inserted into the overall search process. Arguably, Memetic Algorithms \cite{moscato1989evolution} capitalize on this principle, with many application domains having so far harnessed this synergy between global and local search algorithms \cite{neri2012memetic}. As highlighted in the prospective part of this survey, the incorporation of domain-based knowledge in the design of bio-inspired optimization algorithms can play a central role in the future when it comes to the intersection between Deep Learning and bio-inspired optimization.

\bibliographystyle{elsarticle-num}
\bibliography{bibliography}

\end{document}